\documentclass{article}

\usepackage[final]{neurips_2019}

\usepackage[utf8]{inputenc} 
\usepackage[T1]{fontenc}    
\usepackage{hyperref}       
\usepackage{url}            
\usepackage{booktabs}       
\usepackage{amsfonts}       
\usepackage{nicefrac}       
\usepackage{microtype}      
\usepackage{natbib}
\usepackage{subcaption}
\usepackage{graphicx}
\usepackage{mathtools}
\usepackage{color}
\usepackage{makecell}

\usepackage{tikz}
\usepackage{pgf-pie-hacked}

\iffalse 
    \newcommand\todo[1]{}

    \newcommand{\ruben}[1]{}
    \newcommand{\arka}[1]{}
    \newcommand{\harini}[1]{}
    \newcommand{\honglak}[1]{}
    \newcommand{\dumitru}[1]{}
    \newcommand{\quoc}[1]{}
\else 
    \newcommand{\todo}[1]{{\textcolor{red}{[[TODO: {#1}]]}}}

    \newcommand{\ruben}[1]{\textcolor{magenta}{[Ruben: {#1}]}}
    \newcommand{\arka}[1]{\textcolor{green}{[Arkanath: {#1}]}}
    \newcommand{\harini}[1]{\textcolor{blue}{[Harini: {#1}]}}
    \newcommand{\honglak}[1]{\textcolor{orange}{[Honglak: {#1}]}}
    \newcommand{\dumitru}[1]{\textcolor{cyan}{[Dumitru: {#1}]}}
    \newcommand{\quoc}[1]{\textcolor{brown}{[Quoc: {#1}]}}
\fi

\iftrue 
    \newcommand{\cutabstractup}{\vspace*{-0.1in}}
    \newcommand{\cutabstractdown}{\vspace*{-0.1in}}

    \newcommand{\cutsectionup}{\vspace*{-0.08in}}
    \newcommand{\cutsectiondown}{\vspace*{-0.07in}}

    \newcommand{\cutsubsectionup}{\vspace*{-0.08in}}
    \newcommand{\cutsubsectiondown}{\vspace*{-0.07in}}

    \newcommand{\cutparagraphup}{\vspace*{-0.09in}}

\else 
    \newcommand{\cutsectionup}{}
    \newcommand{\cutsectiondown}{}

    \newcommand{\cutsubsectionup}{}
    \newcommand{\cutsubsectiondown}{}

    \newcommand{\cutparagraphup}{}

\fi

\usepackage[export]{adjustbox}
\usepackage{makecell}
\usepackage{siunitx}

\title{High Fidelity Video Prediction with \\Large Stochastic Recurrent Neural Networks}

\author{%
  Ruben Villegas\textsuperscript{\textnormal{1,4}} \quad
  Arkanath Pathak\textsuperscript{\textnormal{3}} \quad
  Harini Kannan\textsuperscript{\textnormal{2}} \\
  \textbf{Dumitru Erhan}\textsuperscript{\textnormal{2}} \quad
  \textbf{Quoc V. Le}\textsuperscript{\textnormal{2}} 
  \quad
  \textbf{Honglak Lee}\textsuperscript{\textnormal{2}} 
  \\
  \textsuperscript{\textnormal{1}} University of Michigan \\
  \textsuperscript{\textnormal{2}} Google Research\\
  \textsuperscript{\textnormal{3}} Google\\
  \textsuperscript{\textnormal{4}} Adobe Research \\
}

\begin{document}

\newcommand\blfootnote[1]{%
  \begingroup
  \renewcommand\thefootnote{}\footnote{#1}%
  \addtocounter{footnote}{-1}%
  \endgroup
}

\maketitle
\thispagestyle{empty}

\vspace{-.1in}
\begin{abstract}
\cutabstractup
Predicting future video frames is extremely challenging, as there are many factors of variation that make up the dynamics of how frames change through time. Previously proposed solutions require complex inductive biases inside network architectures with highly specialized computation, including segmentation masks, optical flow, and foreground and background separation. 
In this work, we question if such handcrafted architectures are necessary and instead propose a different approach: finding minimal inductive bias for video prediction while maximizing network capacity.
We investigate this question by performing the first large-scale empirical study and demonstrate state-of-the-art performance by learning large models on three different datasets: one for modeling object interactions, one for modeling human motion, and one for modeling car driving\footnote{This work was done while the first author was an intern at Google}.
\cutabstractdown
\end{abstract}

\cutsectionup
\section{Introduction}
\cutsectiondown

From throwing a ball to driving a car, humans are very good at being able to interact with objects in the world and anticipate the results of their actions.
Being able to teach agents to do the same has enormous possibilities for training intelligent agents capable of generalizing to many tasks.
Model-based reinforcement learning is one such technique that seeks to do this -- by first learning a model of the world, and then by planning with the learned model. There has been some recent success with training agents in this manner by first using video prediction to model the world.
Particularly, video prediction models combined with simple planning algorithms \citep{danijar} or policy-based learning \citep{lukasz} for model-based reinforcement learning have been shown to perform equally or better than model-free methods with far less interactions with the environment.
Additionally, \citet{frederik} showed that video prediction methods are also useful for robotic control, especially with regards to specifying unstructured goal positions.

However, training an agent to accurately predict what will happen next is still an open problem. Video prediction, the task of generating future frames given context frames, is notoriously hard. There are many spatio-temporal factors of variation present in videos that make this problem very difficult for neural networks to model. Many methods have been proposed to tackle this problem \citep{Oh15,FinnGL16,Vondrick16,villegas17mcnet,prednet,mocogan,dualmo,emily17,Wichers18,mohammad,emily,alexlee,wonmin,Yan18,Kumar19}. Most of these works propose some type of separation of information streams (e.g., motion/pose and content streams), specialized computations (e.g., warping, optical flow, foreground/background masks, predictive coding, etc), additional high-level information (e.g., landmarks, semantic segmentation masks, etc) or are simply shown to work in relatively simpler environments (e.g., Atari, synthetic shapes, centered human faces and bodies, etc).

 Simply making neural networks larger has been shown to improve performance in many areas such as image classification \citep{real18,zoph18,huang18}, image generation \citep{brock18}, and language understanding \citep{devlin18,radford19}, amongst others. Particularly, \citet{brock18} recently showed that increasing the capacity of GANs \citep{NIPS2014_5423} results in dramatic improvements for image generation.
 
 In his blog post "The Bitter Lesson", Rich Sutton comments on these types of developments by arguing that the most significant breakthroughs in machine learning have come from increasing the compute provided to simple models, rather than from specialized, handcrafted architectures \citep{sutton19}. For example, he explains that the early specialized algorithms of computer vision (edge detection, SIFT features, etc.) gave way to larger but simpler convolutional neural networks. In this work, we seek to answer a similar question: do we really need specialized architectures for video prediction? Or is it sufficient to maximize network capacity on models with minimal inductive bias?

In this work, we perform the first large-scale empirical study of the effects of minimal inductive bias and maximal capacity on video prediction. We show that without the need of optical flow, segmentation masks, adversarial losses, landmarks, or any other forms of inductive bias, it is possible to generate high quality video by simply increasing the scale of computation. Overall, our experiments demonstrate that: (1) large models with minimal inductive bias tend to improve the performance both qualitatively and quantitatively, (2) recurrent models outperform non-recurrent models, and (3) stochastic models perform better than non-stochastic models, especially in the presence of uncertainty (e.g., videos with unknown action or control). %

\cutsectionup
\section{Related Work}
\cutsectiondown

The task of predicting multiple frames into the future has been studied for a few years now. Initially, many early methods tried to simply predict future frames in small videos or patches from large videos \citep{Michalsky14,Ranzato14,Srivastava15}. This type of video prediction caused rectangular-shaped artifacts when attempting to fuse the predicted patches, since each predicted patch was blind to its surroundings. Then, action-conditioned video prediction models were built with the aim of being used for model-based reinforcement learning \citep{Oh15,FinnGL16}. Later, video prediction models started becoming more complex and better at predicting future frames. 
\citet{prednet} proposed a neural network based on predictive coding. \citet{villegas17mcnet} proposed to separate motion and content streams in video input. \cite{villegas17hierchvid} proposed to predict future video as landmarks in the future and then use these landmarks to generate frames. \cite{emily17} proposed to have a pose and content encoders as separate information streams. However, all of these methods focused on predicting a single future. Unfortunately, real-world video is highly stochastic -- that is, there are multiple possible futures given a single past.

Many methods focusing on the stochastic nature of real-world videos have been recently proposed. \citet{mohammad} build on the optical flow method proposed by \citet{FinnGL16} by introducing a variational approach to video prediction where the entire future is encoded into a posterior distribution that is used to sample latent variables. \cite{alexlee} also build on optical flow and propose an adversarial version of stochastic video prediction where two discriminator networks are used to enable sharper frame prediction. \cite{emily} also propose a similar variational approach. In their method, the latent variables are sampled from a prior distribution of the future during inference time, and only frames up to the current time step are used to model the future posterior distribution.  \cite{Kumar19} propose a method based on normalizing flows where the exact log-likelihood can be computed for training.

In this work, we investigate whether we can achieve high quality video predictions without the use of the previously mentioned techniques (optical flows, adversarial objectives, etc.) by just maximizing the capacity of a standard neural network. To the best of our knowledge, this work is the first to perform a thorough investigation on the effect of capacity increases for video prediction.

\cutsectionup
\section{Scaling up video prediction} \label{sec:method}
\cutsectiondown

In this section, we present our method for scaling up video prediction networks. We first consider the Stochastic Video Generation (SVG) architecture presented in \citet{emily}, a stochastic video prediction model that is entirely made up of standard neural network layers without any special computations (e. g. optical flow). SVG is competitive with other state-of-the-art stochastic video prediction models (SAVP, SV2P) \citep{alexlee}; however, unlike SAVP and SV2P, it does not use optical flow, adversarial losses, etc. As such, SVG was a fitting starting point to our investigation.

To build our baseline model, we start with the stochastic component that models the inherent uncertainty in future predictions from  \citet{emily}. We also use shallower encoder-decoders that only have convolutional layers to enable more detailed image reconstruction \citep{Dosovitskiy16}. A slightly shallower encoder-decoder architecture results in less information lost in the latent state, as the resulting convolutional map from the bottlenecked layers is larger. Then, in contrast to \citet{emily}, we use a convolutional LSTM architecture, instead of a fully-connected LSTM, to fit the shallow encoders-decoders.
Finally, the last difference is that we optimize the $\ell_1$ loss with respect to the ground-truth frame for all models like in the SAVP model, instead of using $\ell_2$ like in SVG.  \citet{alexlee} showed that $\ell_1$ encouraged sharper frame prediction over $\ell_2$.

We optimize our baseline architecture by maximizing the following variational lowerbound:
\begin{align}
\sum_{t=1}^T \mathbb{E}_{q_\phi(\mathbf{z}_{\leq T}|\mathbf{x}_{\leq T})} \log p_\theta(\mathbf{x}_t|\mathbf{z}_{\leq t}, \mathbf{x}_{< t}) \ - \nonumber \beta D_{KL}\left(q_\phi(\mathbf{z}_t|\mathbf{x}_{\leq t}) || p_\psi(\mathbf{z}_t|\mathbf{x}_{< t})\right),
\label{eq:objective}
\end{align}
where $\mathbf{x}_{t}$ is the frame at time step $t$, $q_\phi(\mathbf{z}_{\leq T}|\mathbf{x}_{\leq T})$ the approximate posterior distribution, $p_\psi(\mathbf{z}_t|\mathbf{x}_{< t})$ is the prior distribution, $p_\theta(\mathbf{x}_t|\mathbf{z}_{\leq t}, \mathbf{x}_{< t})$ is the generative distribution, and $\beta$ regulates the strength of the KL term in the lowerbound. During training time, the frame prediction process at time step $t$ is as follows:
\begin{align}
\mu_{\phi}(t), \sigma_{\phi}(t) &= \text{LSTM}_{\phi}(\mathbf{h}_t; M) &&\text{where} \quad \mathbf{h}_t = f^{\text{enc}}\left(\mathbf{x}_t; K\right), \nonumber \\
\mathbf{z}_t &\sim \mathcal{N}\left(\mu_{\phi}(t), \sigma_{\phi}(t)\right),\nonumber\\
\mathbf{g}_t &= \text{LSTM}_{\theta}\left(\mathbf{h}_{t-1}, \mathbf{z}_t; M\right) &&\text{where} \quad \mathbf{h}_{t-1} = f^{\text{enc}}\left(\mathbf{x}_{t-1}; K\right), \nonumber \\
\mathbf{x}_t &= f^{\text{dec}}\left(\mathbf{g}_t; K\right), \nonumber
\end{align}
where $f^{\text{enc}}$ is an image encoder and $f^{\text{dec}}$ is an image decoder neural network. $\text{LSTM}_{\phi}$ and $\text{LSTM}_{\theta}$ are LSTMs modeling the posterior and generative distributions, respectively. $\mu_{\phi}(t)$ and $\sigma_{\phi}(t)$ are the parameters of the posterior distribution modeling the Gaussian latent code $\mathbf{z}_t$. Finally, $\mathbf{x}_t$ is the predicted frame at time step $t$.

To increase the capacity of our baseline model, we use hyperparameters $K$ and $M$, which denote the factors by which the number of neurons in each layer of the encoder, decoder and LSTMs are increased. For example, if the number of neurons in LSTM is $d$, then we scale up by $d \times M$. The same applies to the encoder and decoder networks but using $K$ as the factor. In our experiments we increase both $K$ and $M$ together until we reach the device limits. Due to the LSTM having more parameters, we stop increasing the capacity of the LSTM at $M=3$ but continue to increase $K$ up to $5$.
At test time, the same process is followed, however, the posterior distribution is replaced by the Gaussian parameters computed by the prior distribution:
\begin{align}
\mu_{\psi}(t), \sigma_{\psi}(t) &= \text{LSTM}_{\psi}(\mathbf{h}_{t-1}; M) &&\text{where} \quad \mathbf{h}_{t-1} = f^{\text{enc}}\left(\mathbf{x}_{t-1}; K\right), \nonumber
\end{align}

Next, we perform ablative studies on our baseline architecture to better quantify exactly how much each individual component affects the quality of video prediction as capacity increases. First, we remove the stochastic component, leaving behind a fully deterministic architecture with just a CNN-based encoder-decoder and a convolutional LSTM. For this version, we simply disable the prior and posterior networks as described above. Finally, we remove the LSTM component, leaving behind only the encoder-decoder CNN architectures. For this version, we simply use $f^{\text{enc}}$ and $f^{\text{dec}}$ as the full video prediction network. However, we let $f^{\text{enc}}$ observe the same number of initial history as the recurrent counterparts.

Details of the devices we use to scale up computation can be found in the supplementary material.

\cutsectionup
\section{Experiments}
\cutsectiondown

In this section, we evaluate our method on three different datasets, each with different challenges. 

\textbf{Object interactions.}.
We use the action-conditioned towel pick dataset from \citet{frederik} to evaluate how our models perform with standard object interactions. This dataset contains a robot arm that is interacting with towel objects. Even though this dataset uses action-conditioning, stochastic video prediction is still required for this task. This is because the motion of the objects is not fully determined by the actions (the movements of the robot arm), but also includes factors such as friction and the objects' current state. For this dataset, we resize the original resolution of 48x64 to 64x64. For evaluation, we use the first 256 videos in the test set as defined by \citet{frederik}.

\textbf{Structured motion.} We use the Human 3.6M dataset \citep{human36m} to measure the ability of our models to predict structured motion. This dataset is comprised of humans performing actions inside a room (walking around, sitting on a chair, etc.). Human motion is highly structured (i.e., many degrees of freedom), and so, it is difficult to model. We use the train/test split from \citet{villegas17hierchvid}. For this dataset, we resize the original resolution of 1000x1000 to 64x64.

\textbf{Partial observability.} We use the KITTI driving dataset \citep{kitti} to measure how our models perform in conditions of partial observability. This dataset contains driving scenes taken from a front camera view of a car driving in the city, residential neighborhoods, and on the road. The front view camera of the vehicle causes partial observability of the vehicle environment, which requires a model to generate seen and unseen areas when predicting future frames. We use the train/test split from \citet{prednet} in our experiments. We extract 30 frame clips and skip every 5 frames from the test set so that the test videos do not significantly overlap, which gives us 148 test clips in the end. For this dataset, we resize the original resolution of 128x160 to 64x64.

\cutsubsectionup
\subsection{Evaluation metrics} \label{sec:eval_metrics}
\cutsubsectiondown

\begin{table}[t] 
\centering
\setlength{\tabcolsep}{3pt}
\begin{tabular}{l||cc||cc||cc}
\Xhline{4\arrayrulewidth}
\multicolumn{1}{c||}{} & \multicolumn{2}{c||}{CNN models} & \multicolumn{2}{c||}{LSTM models} & \multicolumn{2}{c}{SVG' models} \\
\Xhline{4\arrayrulewidth}
Dataset & \makecell{Biggest \\ (M=3, K=5)} & \makecell{Baseline \\ (M=1, K=1)} & \makecell{Biggest \\ (M=3, K=5)} & \makecell{Baseline \\ (M=1, K=1)}& \makecell{Biggest \\ (M=3, K=5)} & \makecell{Baseline \\ (M=1, K=1)} \\
\Xhline{4\arrayrulewidth}
Towel Pick & 199.81 & 281.07 & 100.04 & 206.49 & \textbf{93.71} & 189.91 \\
\hline
Human 3.6M & 1321.23 & 1077.55 & 458.77 & 614.21 & \textbf{429.88} & 682.08 \\
\hline
KITTI & 2414.64 & 2906.71 & \textbf{1159.25} & 2502.69 & 1217.25 & 2264.91 \\
\Xhline{4\arrayrulewidth}
\end{tabular}
\vspace{7pt}
\caption{\textbf{Fr\'echet Video Distance evaluation (lower is better)}. We compare the biggest model we were able to train against the baseline models (M=1, K=1). Note that all models (SVG', CNN, and LSTM). The biggest recurrent models are significantly better than their small counterpart. Please refer to our supplementary material for plots showing how gradually increasing model capacity results in better performance.}
\vspace{-0.2in}
\label{table:fvd_together}
\end{table}

We perform a rigorous evaluation using five different metrics: 
Peak Signal-to-Noise Ratio (PSNR), Structural Similarity (SSIM), VGG Cosine Similarity, Fr\'echet Video Distance (FVD) \citep{fvd}, and human evaluation from Amazon Mechanical Turk (AMT) workers.
We perform these evaluations on all models described in Section \ref{sec:method}: our baseline (denoted as SVG'), the recurrent deterministic model (denoted as LSTM), and the encoder-decoder CNN model (denoted as CNN). In addition, we present a study comparing the video prediction performance as a result of using skip-connections from every layer of the encoder to every layer of the decoder versus not using skip connections at all (Supplementary \ref{supp:skipconnections}), and the effects of the number of context frames (Supplementary \ref{supp:contextframes}).
\cutsubsectionup
\subsubsection{Frame-wise evaluation}
\cutsubsectiondown
We use three different metrics to perform frame-wise evaluation: PSNR, SSIM, and VGG cosine similarity. 
PSNR and SSIM perform a pixel-wise comparison between the predicted frames and generated frames, effectively measuring if the exact pixels have been generated.
VGG Cosine Similarity has been used in prior work \citep{alexlee} to compare frames in a perceptual level. VGGnet \citep{vggnet} is used to extract features from the predicted and ground-truth frames, and cosine similarity is performed at feature-level. Similar to \citet{Kumar19,mohammad,alexlee}, we sample 100 future trajectories per video and pick the highest scoring trajectory as the main score.

\cutsubsectionup
\subsubsection{Dynamics-based evaluation}
\cutsubsectiondown
We use two different metrics to measure the overall realism of the generated videos: FVD and human evaluations. FVD, a recently proposed metric for video dynamics accuracy, uses a 3D CNN trained for video classification to extract a single feature vector from a video. Analogous to the well-known FID \citep{fid}, it compares the distribution of features extracted from ground-truth videos and generated videos. Intuitively, this metric compares the quality of the overall predicted video dynamics with that of the ground-truth videos rather than a per-frame comparison. For FVD, we also sample 100 future trajectories per video, but in contrast, all 100 trajectories are used in this evaluation metric (i.e., not just the max, as we did for VGG cosine similarity).

We also use Amazon Mechanical Turk (AMT) workers to perform human evaluations. The workers are presented with two videos (baseline and largest models) and asked to either select the more realistic video or mark that they look about the same. We choose the videos for both models by selecting the highest scoring videos in terms of the VGG cosine similarity with respect to the ground truth. 
We use 10 unique workers per video and choose the selection with the most votes as the final answer.
Finally, we also show qualitative evaluations on pairs of videos, also selected by using the highest VGG cosine similarity scores for both the baseline and the largest model.
We run the human perception based evaluation on the best two architectures we scale up.

\begin{table}[t] 
\centering
\setlength{\tabcolsep}{3pt}
\begin{tabular}{l||ccc||ccc}
\Xhline{4\arrayrulewidth}
\multicolumn{1}{c||}{} & \multicolumn{3}{c||}{LSTM models} & \multicolumn{3}{c}{SVG' models} \\
\Xhline{4\arrayrulewidth}
Dataset & \makecell{Biggest \\ (M=3, K=5)} & \makecell{Baseline \\ (M=1, K=1)} & \makecell{About \\ the same}  & \makecell{Biggest  \\ (M=3, K=5)} & \makecell{Baseline  \\ (M=1, K=1)} & \makecell{About \\ the same} \\
\Xhline{4\arrayrulewidth}
Towel Pick & \textbf{90.2}\% & 9.0\% & 0.8\% & \textbf{68.8}\% & 25.8\% & 5.5\% \\
\hline
Human 3.6M & \textbf{98.7}\% & 1.3\% & 0.0\% & \textbf{95.8}\% & 3.4\% & 0.8\% \\
\hline
KITTI & \textbf{99.3}\% & 0.7\% & 0.0\% & \textbf{99.3}\% & 0.7\% & 0.0\% \\
\Xhline{4\arrayrulewidth}
\end{tabular}
\vspace{4pt}
\caption{\textbf{Amazon Mechanical Turk human worker preference}. We compared the biggest and baseline models from LSTM and SVG'. The bigger models are more frequently preferred by humans. We present a full comparison for all large models in Supplementary \ref{supp:all_compare}.}
\vspace{-0.2in}
\label{table:mturk_together}
\end{table}

\cutsubsectionup
\subsection{Robot arm} \label{sec:robots}
\cutsubsectiondown
For this dataset, we perform action-conditioned video prediction. We modify the baseline and large models to take in actions as additional input to the video prediction model. Action conditioning does not take away the inherent stochastic nature of video prediction due to the dynamics of the environment. During training time, the models are conditioned on 2 input frames and predict 10 frames into the future. During test time, the models predict 18 frames into the future.

\cutparagraphup
\paragraph{Dynamics-based evaluation.}
We first evaluate the action-conditioned video prediction models using FVD to measure the realism in the dynamics. In Table \ref{table:fvd_together} (top row), we present the results of scaling up the three models described in Section \ref{sec:method}. Firstly, we see that our baseline architecture improves dramatically at the largest capacity we were able to train. Secondly, for our ablative experiments, we notice that larger capacity improves the performance of the vanilla CNN architecture. Interestingly, by increasing the capacity of the CNN architecture, it approaches the performance of the baseline SVG' architecture. However, as capacity increases, the lack of recurrence heavily affects the performance of the vanilla CNN architecture in comparison with the models that do have an LSTM (Supplementary  \ref{supp:robot_arm}).
Both the LSTM model and SVG' perform similarly well, with SVG' model performing slightly better. This makes sense as the deterministic LSTM model is more likely to produce videos closer to the ground truth; however, the stochastic component is still quite important as a good video prediction model must be both realistic and capable of handling multiple possible futures. 
Finally, we use human evaluations through Amazon Mechanical Turk to compare our biggest models with the corresponding baselines. We asked workers to focus on how realistic the interaction between the robot arm and objects looks. As shown in Table \ref{table:mturk_together}, the largest SVG' is preferred 68.8\% of the time vs 25.8\% of the time for the baseline (right), and the largest LSTM model is preferred 90.2\% of the time vs 9.0\% of the time for the baseline (left). 

\begin{figure}[htp!]
    \centering
    \hspace{-8pt}
	\includegraphics[width=.35\linewidth]{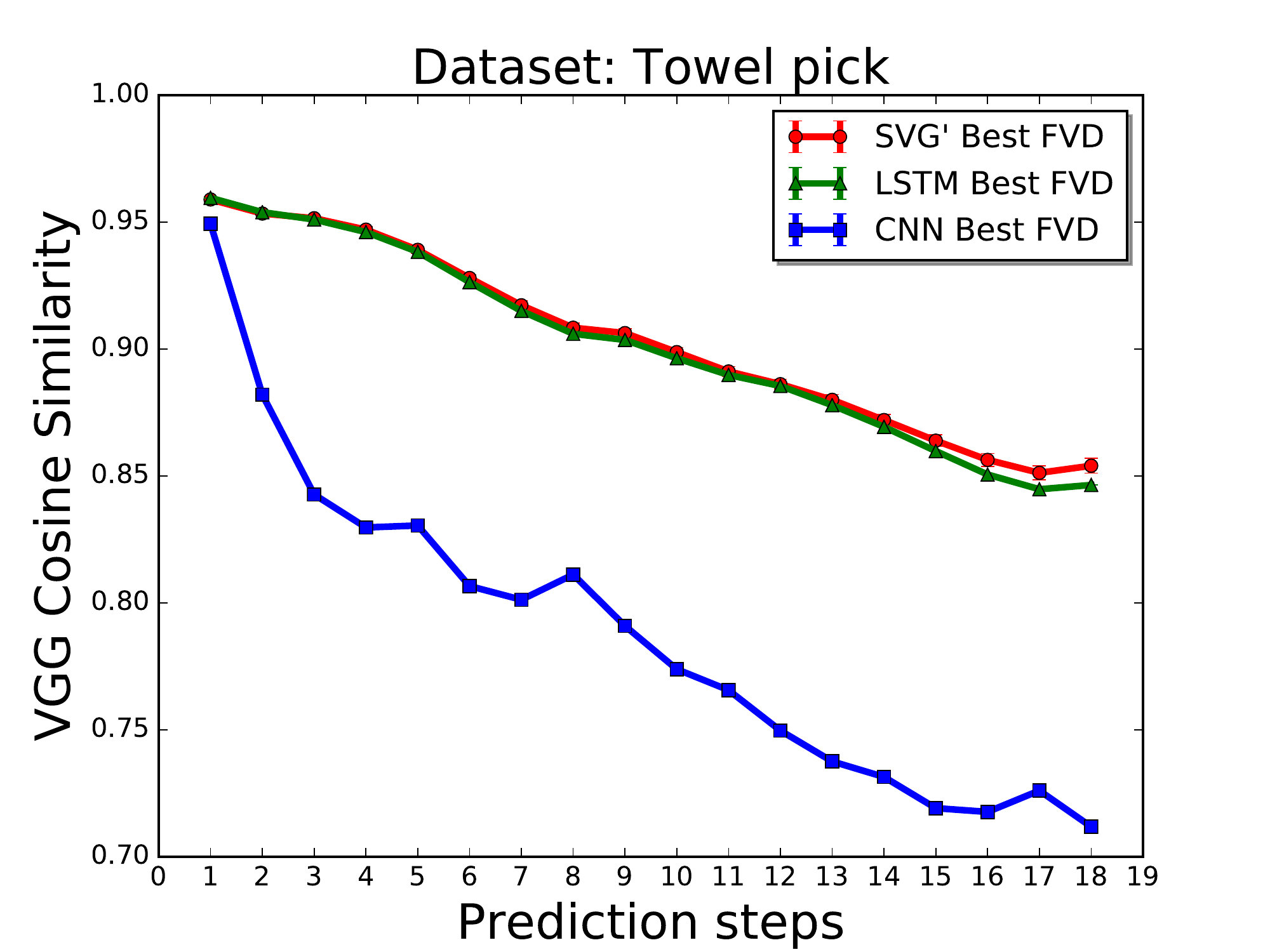} \hspace{-14pt}
	\includegraphics[width=.35\linewidth]{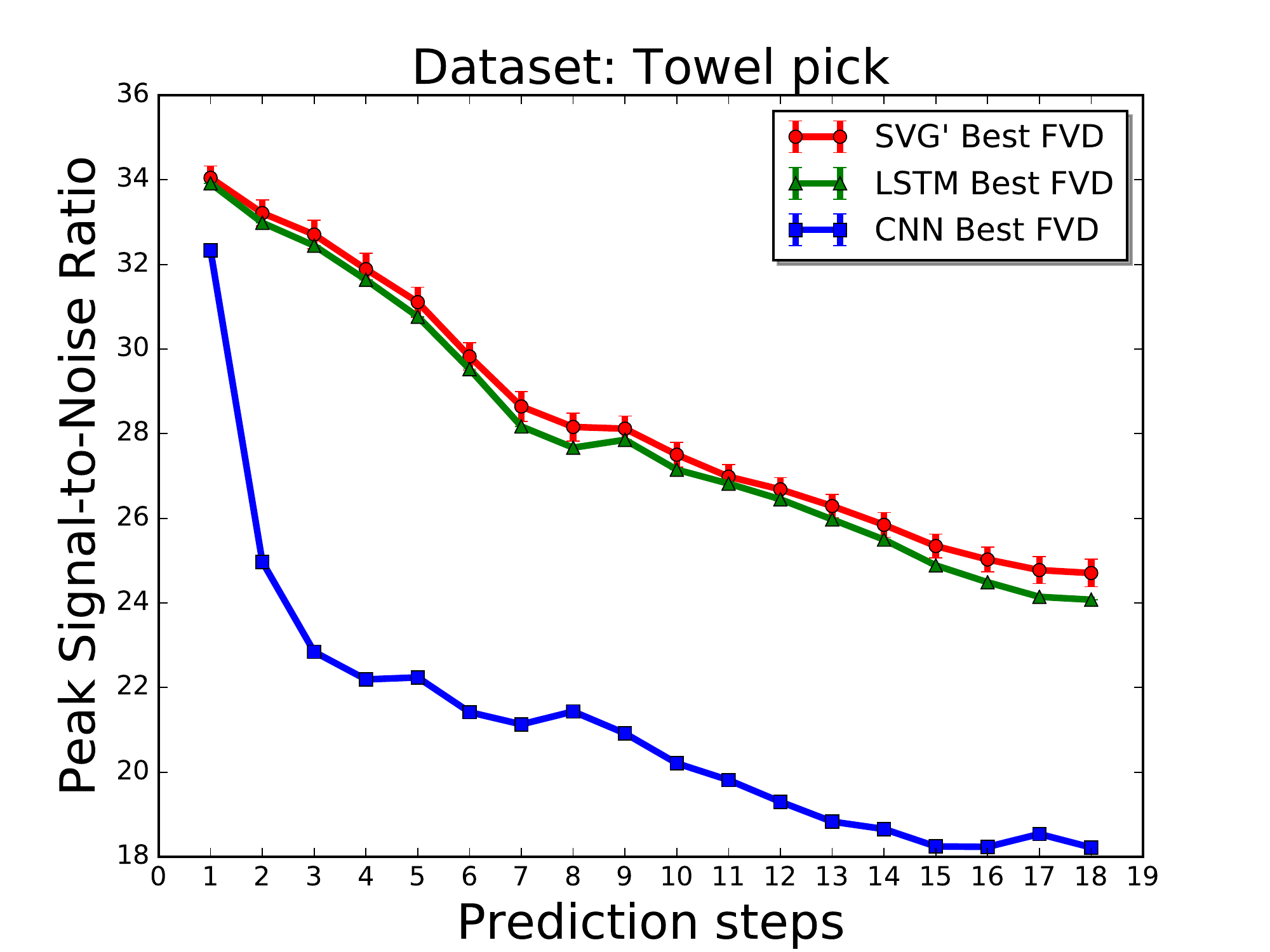} \hspace{-14pt}
	\includegraphics[width=.35\linewidth]{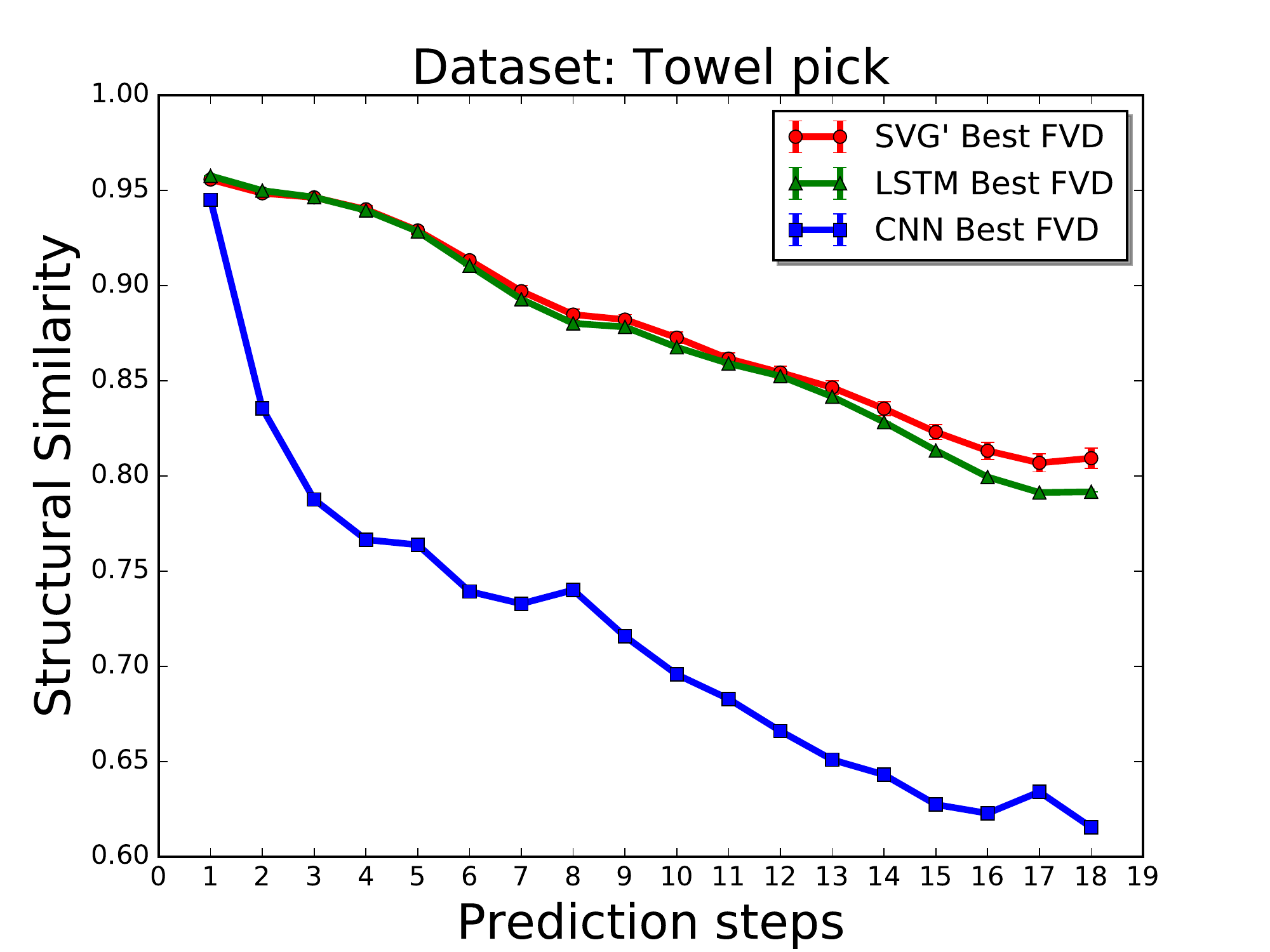}
	\caption{Towel pick per-frame evaluation (higher is better). We compare the best performing models in terms of FVD. For model capacity comparisons, please refer to Supplementary \ref{supp:robot_arm}.}
\label{fig:frame_towel}
\end{figure}

\cutparagraphup
\paragraph{Frame-wise evaluation.}
Next, we use FVD to select the best models from CNN, LSTM, and SVG', and perform frame-wise evaluation on each of these three models. 
Since models that copy background pixels perfectly can perform well on these frame-wise evaluation metrics, in the supplementary material we also discuss a comparison against a simple baseline where the last observed frame is copied through time. 
From Figure \ref{fig:frame_towel}, we can see that the CNN model performs much worse than the models that have recurrent connections. This is a clear indication that recurrence is necessary to predict future frames, and capacity cannot make up for it. Both LSTM and SVG perform similarly well, however, towards the end, SVG slightly outperforms LSTM. The full evaluation on all capacities for SVG', LSTM, and CNN is presented in the supplementary material.

\begin{figure}[thbp]
    \hspace*{-.5cm}
    \centering
	\begin{subfigure}{0.1\linewidth}
        \raggedleft
        \rotatebox{90}{
        \hspace{-.5cm}
        \parbox{1.6cm}{\centering Smallest model \\ (Baseline)}
        \parbox{1.6cm}{\centering Biggest model \\ (Ours)}
        \parbox{1.6cm}{\centering Ground-truth}
        }
    \end{subfigure}
    \begin{subfigure}{0.12\linewidth}
        \caption*{t=5}
        \vspace{-7pt}
  		\includegraphics[width=1\linewidth]{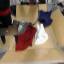}
  		\includegraphics[width=1\linewidth]{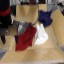}
  		\includegraphics[width=1\linewidth]{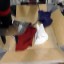}
	\end{subfigure} 
    \begin{subfigure}{0.12\linewidth}
        \caption*{t=6}
        \vspace{-7pt}  		\includegraphics[width=1\linewidth]{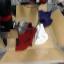}
  		\includegraphics[width=1\linewidth]{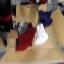}
  		\includegraphics[width=1\linewidth]{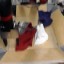}
	\end{subfigure}
    \begin{subfigure}{0.12\linewidth}
        \caption*{t=9}
        \vspace{-7pt}
  		\includegraphics[width=1\linewidth]{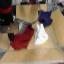}
  		\includegraphics[width=1\linewidth]{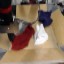}
  		\includegraphics[width=1\linewidth]{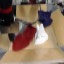}
	\end{subfigure} 
    \begin{subfigure}{0.12\linewidth}
        \caption*{t=12}
        \vspace{-7pt}
  		\includegraphics[width=1\linewidth]{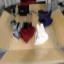}
  		\includegraphics[width=1\linewidth]{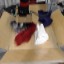}
  		\includegraphics[width=1\linewidth]{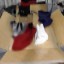}
	\end{subfigure}
	\begin{subfigure}{0.12\linewidth}
        \caption*{t=15}
        \vspace{-7pt}
  		\includegraphics[width=1\linewidth]{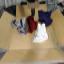}
  		\includegraphics[width=1\linewidth]{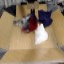}
  		\includegraphics[width=1\linewidth]{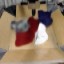}
	\end{subfigure}
	\begin{subfigure}{0.12\linewidth}
        \caption*{t=18}
        \vspace{-7pt}
  		\includegraphics[width=1\linewidth]{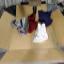}
  		\includegraphics[width=1\linewidth]{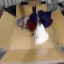}
  		\includegraphics[width=1\linewidth]{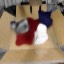}
	\end{subfigure}
	\begin{subfigure}{0.12\linewidth}
        \caption*{t=20}
        \vspace{-7pt}
  		\includegraphics[width=1\linewidth]{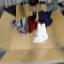}
  		\includegraphics[width=1\linewidth]{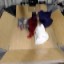}
  		\includegraphics[width=1\linewidth]{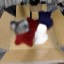}
	\end{subfigure}


    \caption{Robot towel pick qualitative evaluation. Our highest capacity model (middle row) produces better modeling of the robot arm dynamics, as well, as object interactions. The baseline model (bottom row) fails at modeling the objects (object blurriness), and also, the robot arm dynamics are not well modeled (gripper is open when the it should be close at t=18). 
    For best viewing and more results, please visit our website \url{https://cutt.ly/QGuCex}.
    }
\label{fig:qualitative_towel}
\vspace{-0.1in}
\end{figure}

\cutparagraphup
\paragraph{Qualitative evaluation.} In Figure \ref{fig:qualitative_towel}, we show example videos from the smallest SVG' model, the largest SVG' model, and the ground truth. The predictions from the small baseline model are blurrier compared to the largest model, while the edges of objects from the larger model's predictions stay continuously sharp throughout the entire video. This is clear evidence that increasing the model capacity enables more accurate modeling of the pick up dynamics.
For more videos, please visit our website \url{https://cutt.ly/QGuCex}

\cutsubsectionup
\subsection{Human activities} \label{sec:humans}
\cutsubsectiondown
For this dataset, we perform action-free video prediction. We use a single model to predict all action sequences in the Human 3.6M dataset. During training time, the models are conditioned on 5 input frames and predict 10 frames into the future. At test time, the models predict 25 frames.

\cutparagraphup \paragraph{Dynamics-based evaluation.}
We evaluate the predicted human motion with FVD (Table \ref{table:fvd_together}, middle row). The performance of the CNN model is poor in this dataset, and increasing the capacity of the CNN does not lead to any increase in performance. We hypothesize that this is because the lack of action conditioning and the many degrees of freedom in human motion makes it very difficult to model with a simple encoder-decoder CNN. However, after adding recurrence, both LSTM and SVG' perform significantly better, and both models' performance become better as their capacity is increased (Supplementary  \ref{supp:humans}). Similar to Section \ref{sec:robots}, we see that SVG' performs better than LSTM. This is again likely due to the ability to sample multiple futures, leading to a higher probability of matching the ground truth future. Secondly, in our human evaluations for SVG', 95.8\% of the AMT workers agree that the bigger model has more realistic videos in comparison to the smaller model, and for LSTM, 98.7\% of the workers agree that the LSTM largest model is more realistic. Our results, especially the strong agreement from our human evaluations, show that high capacity models are better equipped to handle the complex structured dynamics in human videos. 

\begin{figure}[htp!]
    \centering
    \hspace{-8pt}
	\includegraphics[width=.35\linewidth]{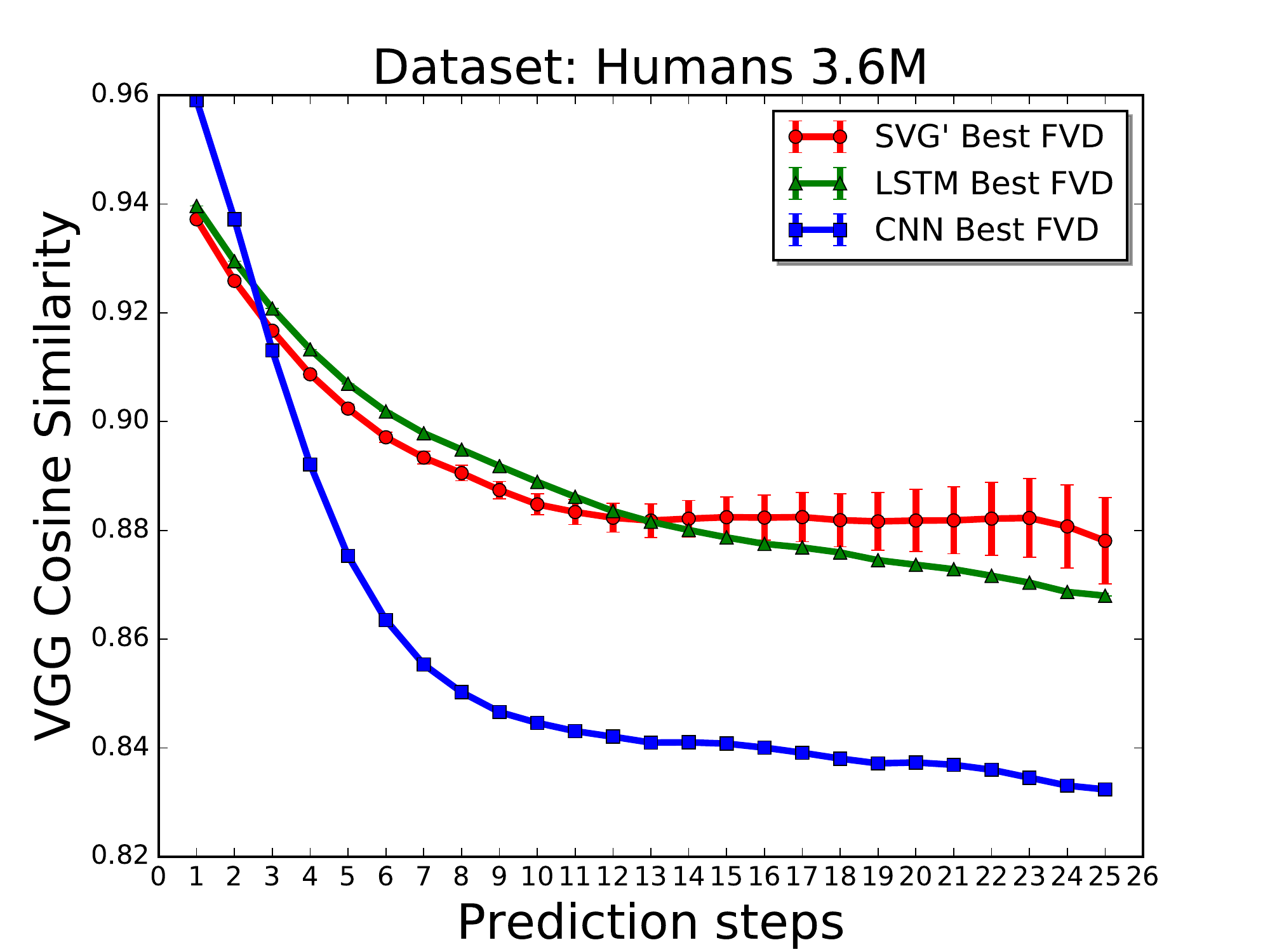} \hspace{-14pt}
	\includegraphics[width=.35\linewidth]{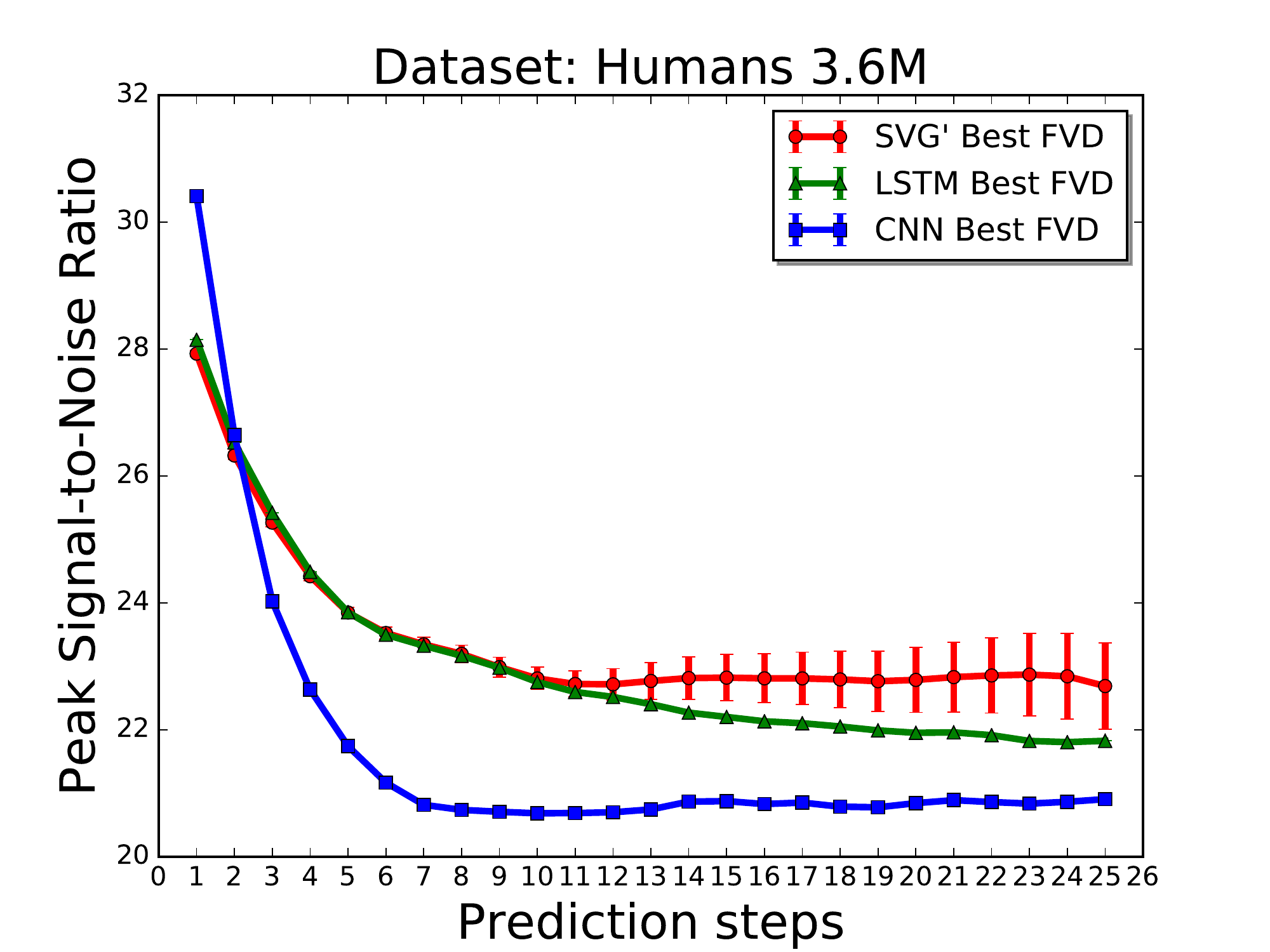} \hspace{-14pt}
	\includegraphics[width=.35\linewidth]{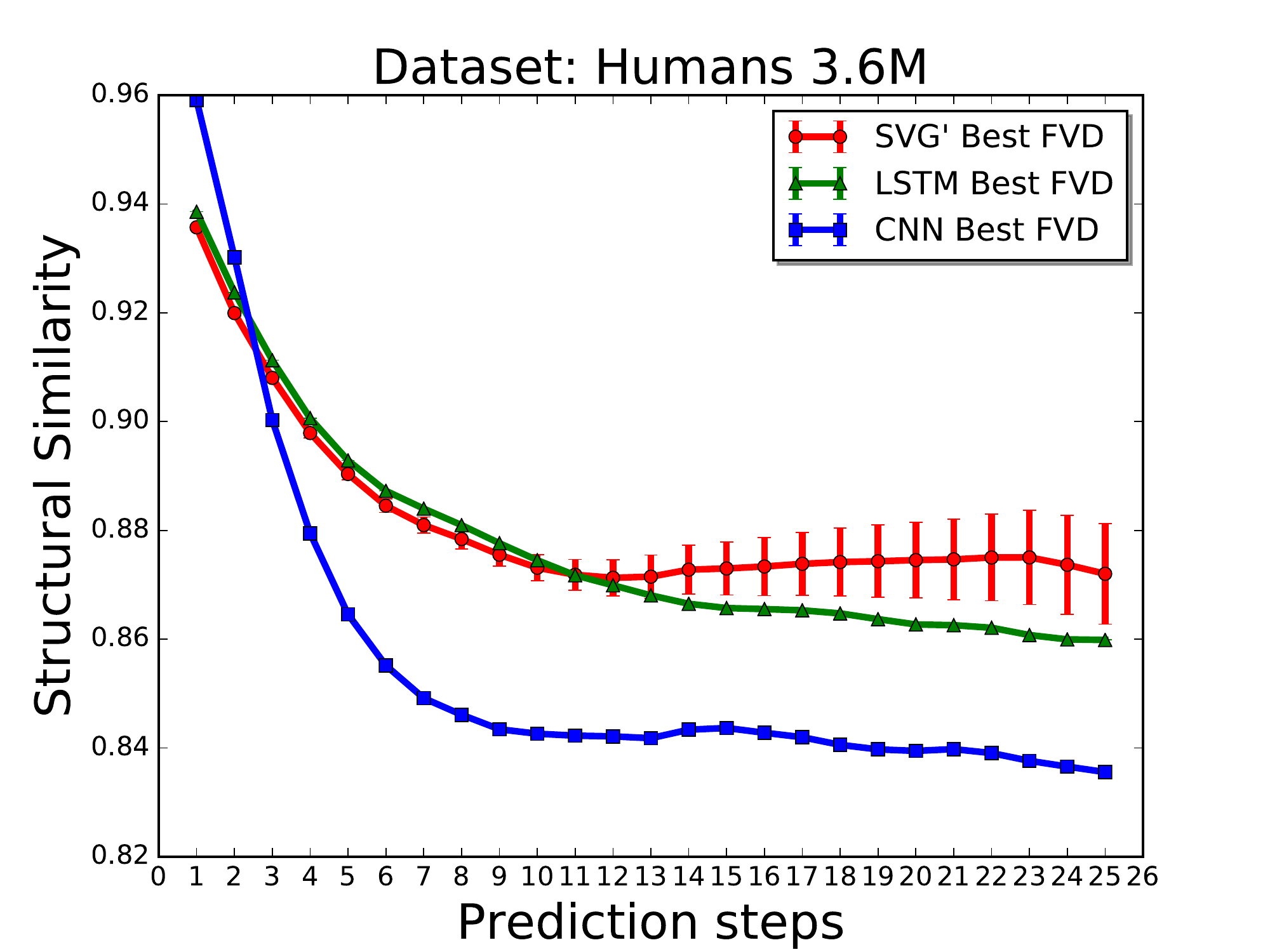}
	\caption{Human 3.6M per-frame evaluation (higher is better). We compare the best performing models in terms of FVD. For model capacity comparisons, please refer to Supplementary \ref{supp:humans}.}
\label{fig:frame_humans}
\end{figure}

\begin{figure}[t]
    \hspace*{-.5cm}
    \centering
	\begin{subfigure}{0.1\linewidth}
        \raggedleft
        \rotatebox{90}{
        \hspace{-.5cm}
        \parbox{1.6cm}{\centering Smallest model \\ (Baseline)}
        \parbox{1.6cm}{\centering Biggest model \\ (Ours)}
        \parbox{1.6cm}{\centering Ground-truth}
        }
    \end{subfigure}
    \begin{subfigure}{0.12\linewidth}
        \caption*{t=8}
        \vspace{-7pt}
  		\includegraphics[width=1\linewidth]{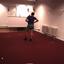}
  		\includegraphics[width=1\linewidth]{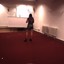}
  		\includegraphics[width=1\linewidth]{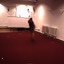}
	\end{subfigure} 
    \begin{subfigure}{0.12\linewidth}
        \caption*{t=11}
        \vspace{-7pt}  		\includegraphics[width=1\linewidth]{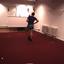}
  		\includegraphics[width=1\linewidth]{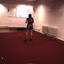}
  		\includegraphics[width=1\linewidth]{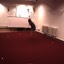}
	\end{subfigure}
    \begin{subfigure}{0.12\linewidth}
        \caption*{t=14}
        \vspace{-7pt}
  		\includegraphics[width=1\linewidth]{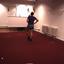}
  		\includegraphics[width=1\linewidth]{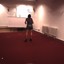}
  		\includegraphics[width=1\linewidth]{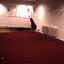}
	\end{subfigure} 
    \begin{subfigure}{0.12\linewidth}
        \caption*{t=17}
        \vspace{-7pt}
  		\includegraphics[width=1\linewidth]{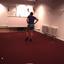}
  		\includegraphics[width=1\linewidth]{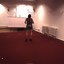}
  		\includegraphics[width=1\linewidth]{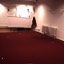}
	\end{subfigure}
	\begin{subfigure}{0.12\linewidth}
        \caption*{t=20}
        \vspace{-7pt}
  		\includegraphics[width=1\linewidth]{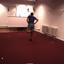}
  		\includegraphics[width=1\linewidth]{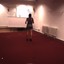}
  		\includegraphics[width=1\linewidth]{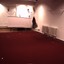}
	\end{subfigure}
	\begin{subfigure}{0.12\linewidth}
        \caption*{t=23}
        \vspace{-7pt}
  		\includegraphics[width=1\linewidth]{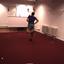}
  		\includegraphics[width=1\linewidth]{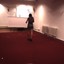}
  		\includegraphics[width=1\linewidth]{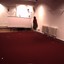}
	\end{subfigure}
	\begin{subfigure}{0.12\linewidth}
        \caption*{t=26}
        \vspace{-7pt}
  		\includegraphics[width=1\linewidth]{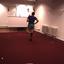}
  		\includegraphics[width=1\linewidth]{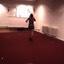}
  		\includegraphics[width=1\linewidth]{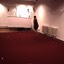}
	\end{subfigure}

    \caption{Human 3.6M qualitative evaluation. Our highest capacity model (middle) produces better modeling of the human dynamics. The baseline model (bottom) is able to keep the human dynamics to some degree but in often cases the human shape is unrecognizable or constantly vanishing and reappearing. For more videos, please visit our website \url{https://cutt.ly/QGuCex}.}
\label{fig:qualitative_humans}
\end{figure}

\cutparagraphup 
\paragraph{Frame-wise evaluation.} Similar to the previous per-frame evaluation, we select the best performing models in terms of FVD and perform a frame-wise evaluation.
In Figure \ref{fig:frame_humans}, we can see that the CNN based model performs poorly against the LSTM and SVG' baselines. The recurrent connections in LSTM and SVG' are necessary to be able to identify the human structure and the action being performed in the input frames. In contrast to Section \ref{sec:robots}, there are no action inputs to guide the video prediction which significantly affects the CNN baseline. The LSTM and SVG' networks perform similarly at the beginning of the video while SVG' outperforms LSTM in the last time steps. This is a result of SVG' being able to model multiple futures from which we pick the best future for evaluation as described in Section \ref{sec:eval_metrics}.
We present the full evaluation on all capacities for SVG’, LSTM, and CNN in the supplementary material.

\cutparagraphup
\paragraph{Qualitative evaluation.} Figure \ref{fig:qualitative_humans} shows a comparison between the smallest and largest stochastic models. In the video generated by the smallest model, the shape of the human is not well-defined at all, while the largest model is able to clearly depict the arms and the legs of the human. Moreover, our large model is able to successfully predict the human's movement throughout all of the frames into the future. The predicted motion is close to the ground-truth motion providing evidence that being able to model more factors of variation with larger capacity models can enable accurate motion identification and prediction.
For more videos, please visit our website \url{https://cutt.ly/QGuCex}.

\cutsubsectionup
\subsection{Car driving}
\cutsubsectiondown
For this dataset, we also perform action-free video prediction. During training time, the models are conditioned on 5 input frames and predict 10 frames into the future. At test time, the models predict 25 frames into the future. This video type is the most difficult to predict since it requires the model to be able to hallucinate unseen parts in the video given the observed parts.

\cutparagraphup
\paragraph{Dynamics-based evaluation.} We see very similar results to the previous dataset when measuring the realism of the videos. For both LSTM and SVG', we see a large improvement in FVD when comparing the baseline model to the largest model we were able to train (Table \ref{table:fvd_together}, bottom row). However, we see a similarly poor performance for the CNN architecture as in Section \ref{sec:humans}, where capacity does not help.
One interesting thing to note is that the largest LSTM model performs better than the largest SVG' model. This is likely related to the architecture design and the data itself. The movements of cars driving is mostly predictable, and so, the deterministic architecture becomes highly competitive as we increase the model capacity (Supplementary \ref{supp:driving}).
However, our original premise that increasing model's capacity improves network performance still holds.
Finally, for human evaluations, we see in Table \ref{table:mturk_together} that the largest capacity SVG' model is preferred by human raters 99.3\% of the time (right), and the largest capacity LSTM model (left) is also preferred by human raters 99.3\% time (left).

\cutparagraphup
\paragraph{Frame-wise evaluation} Now, when we evaluate based on frame-wise accuracy, we see similar but not exactly the same behavior as the experiments in Section \ref{sec:humans}. The CNN architecture performs poorly as expected, however, LSTM and SVG' perform similarly well.

\begin{figure}[t]
    \centering
    \hspace{-8pt}
	\includegraphics[width=.35\linewidth]{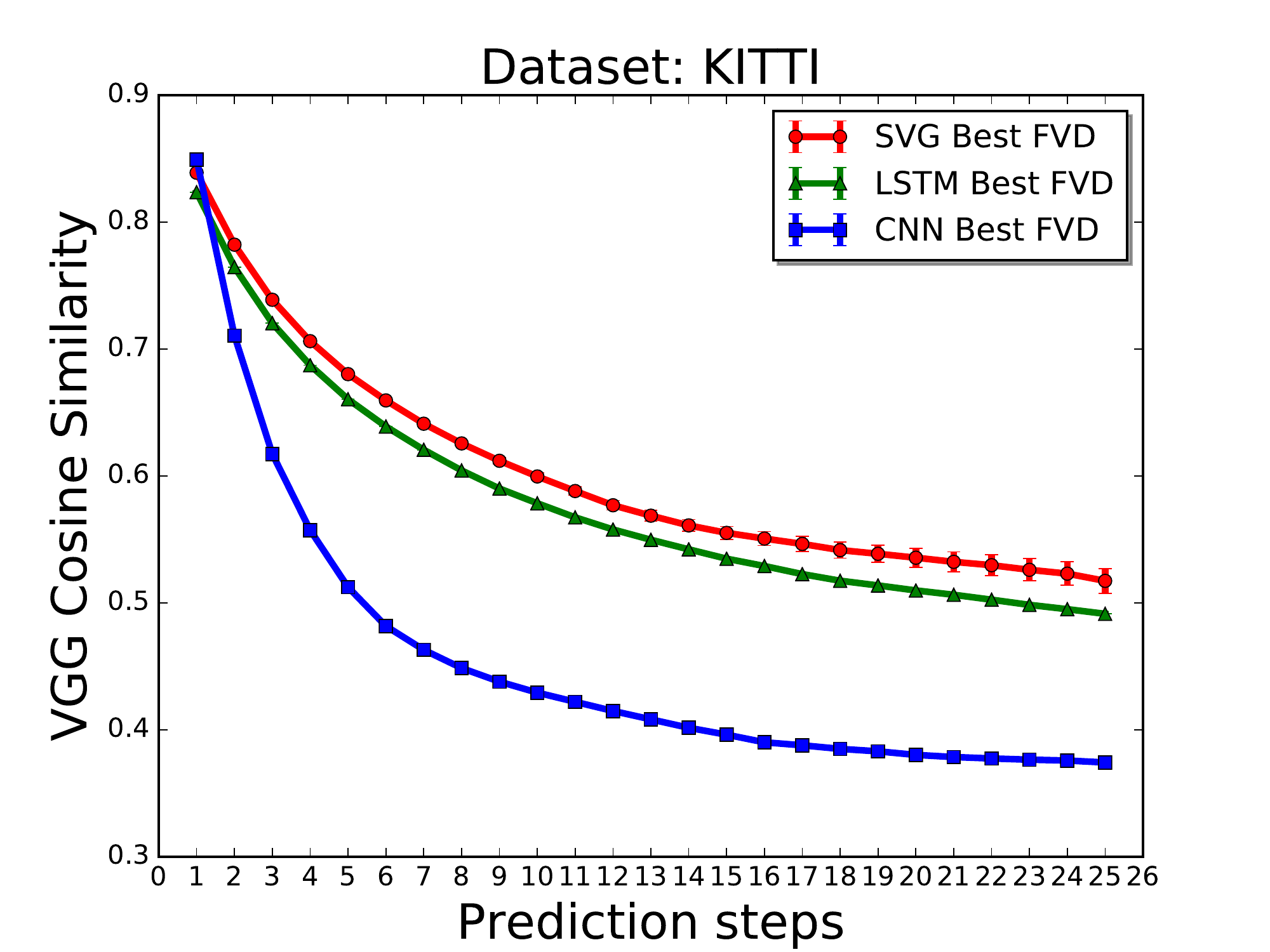} \hspace{-14pt}
	\includegraphics[width=.35\linewidth]{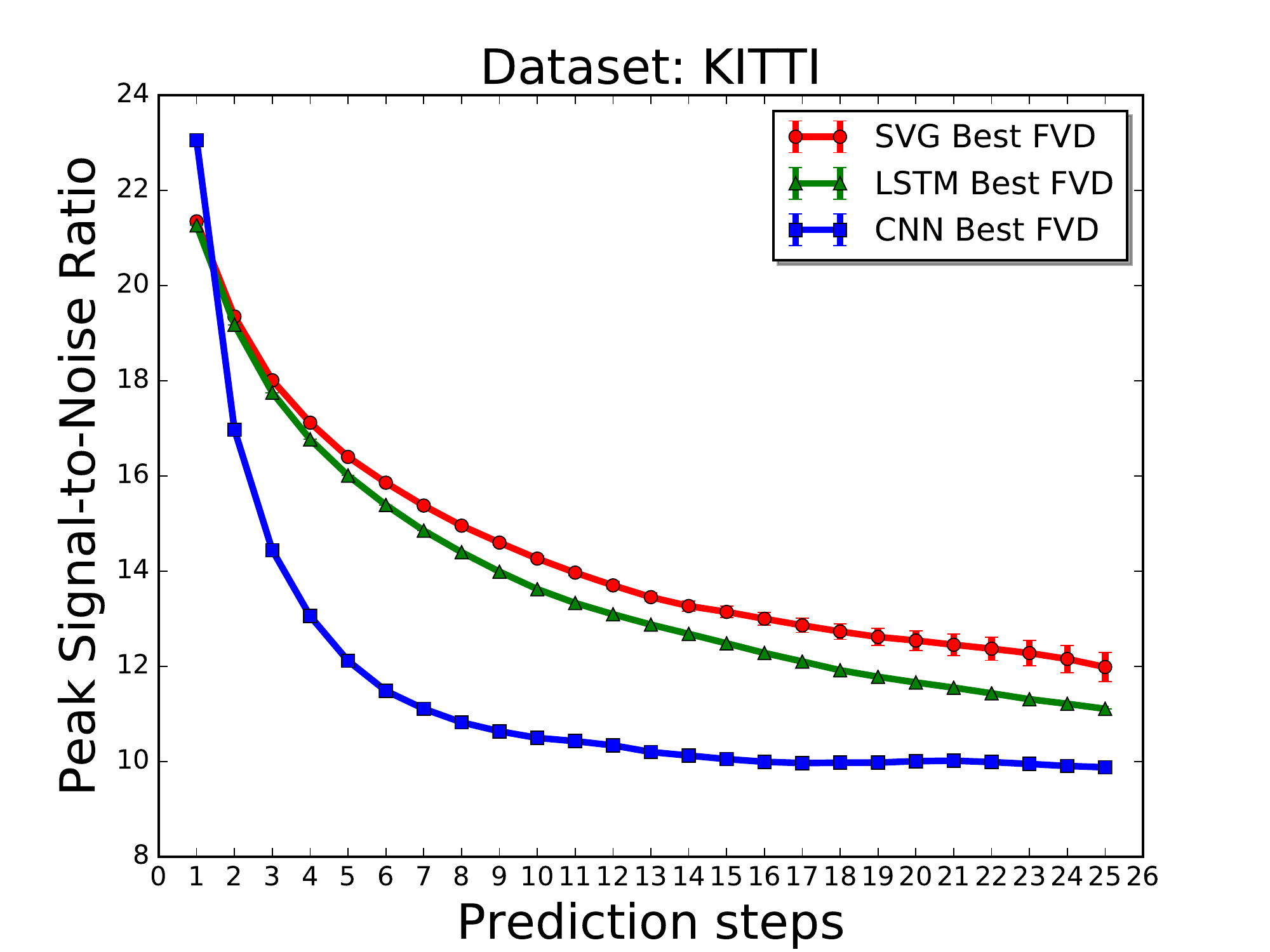} \hspace{-14pt}
	\includegraphics[width=.35\linewidth]{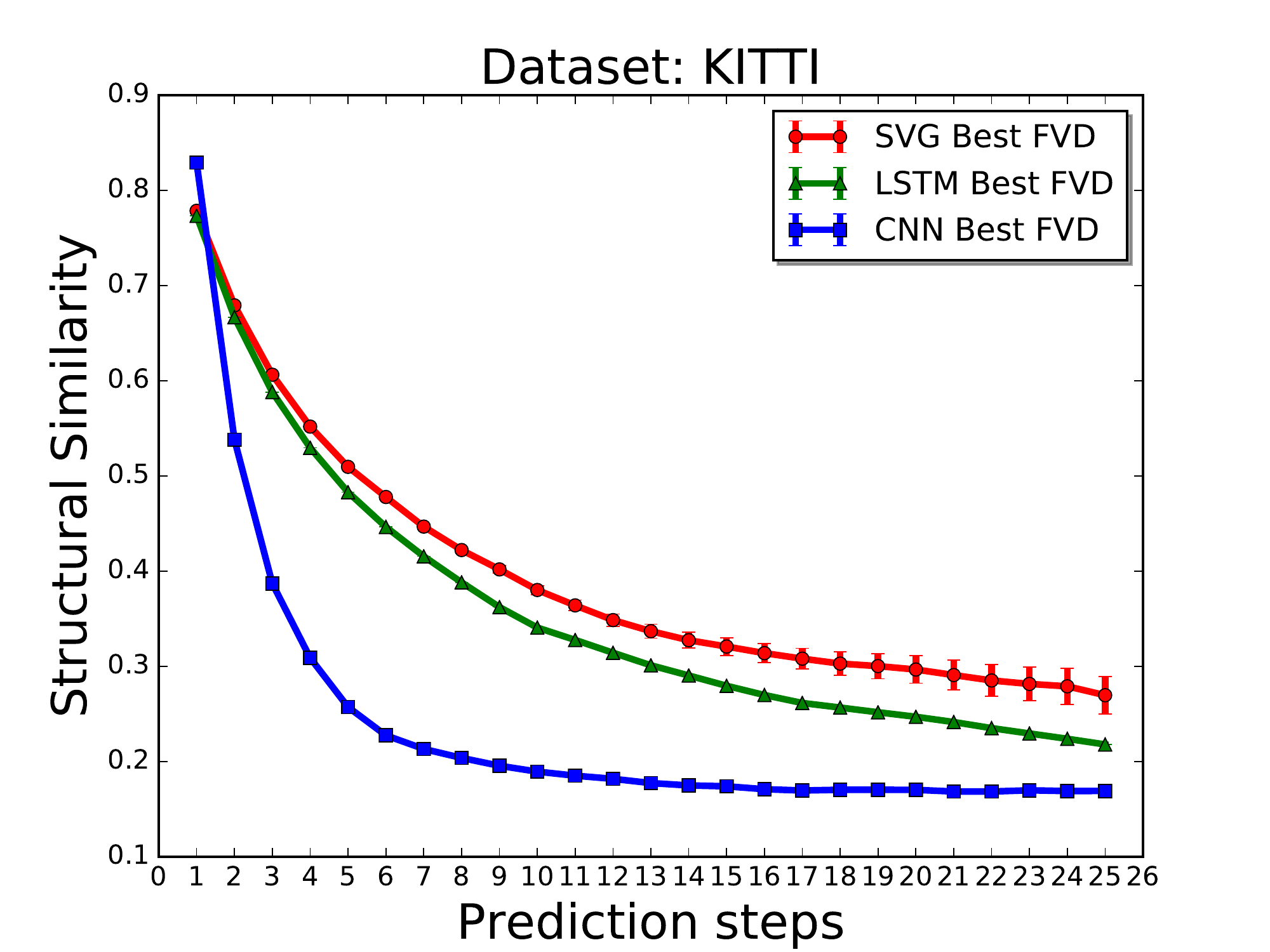}
	\caption{KITTI driving per-frame evaluation (higher is better). For model capacity comparisons, please refer to Supplementary \ref{supp:driving}.}
\label{fig:frame_kitti}
\end{figure}

\begin{figure}[t]
    \hspace*{-.5cm}
    \hspace{-14pt}
    \centering
	\begin{subfigure}{0.1\linewidth}
        \raggedleft
        \rotatebox{90}{
        \hspace{-.3cm}
        \parbox{1.6cm}{\centering Smallest model \\ (Baseline)}
        \parbox{1.6cm}{\centering Biggest model \\ (Ours)}
        \parbox{1.6cm}{\centering Ground-truth}
        }
    \end{subfigure}
    \begin{subfigure}{0.12\linewidth}
        \vspace{5pt}
  		\includegraphics[width=1\linewidth]{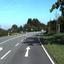}
  		\includegraphics[width=1\linewidth]{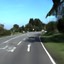}
  		\includegraphics[width=1\linewidth]{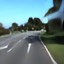}
	\end{subfigure} 
    \begin{subfigure}{0.12\linewidth}
        \vspace{5pt}  		\includegraphics[width=1\linewidth]{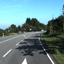}
  		\includegraphics[width=1\linewidth]{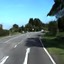}
  		\includegraphics[width=1\linewidth]{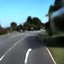}
	\end{subfigure}
    \begin{subfigure}{0.12\linewidth}
        \vspace{5pt}
  		\includegraphics[width=1\linewidth]{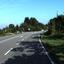}
  		\includegraphics[width=1\linewidth]{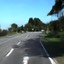}
  		\includegraphics[width=1\linewidth]{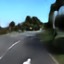}
	\end{subfigure} 
    \begin{subfigure}{0.12\linewidth}
        \vspace{5pt}
  		\includegraphics[width=1\linewidth]{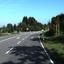}
  		\includegraphics[width=1\linewidth]{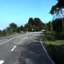}
  		\includegraphics[width=1\linewidth]{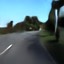}
	\end{subfigure}
	\begin{subfigure}{0.12\linewidth}
        \vspace{5pt}
  		\includegraphics[width=1\linewidth]{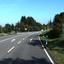}
  		\includegraphics[width=1\linewidth]{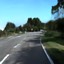}
  		\includegraphics[width=1\linewidth]{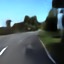}
	\end{subfigure}
	\begin{subfigure}{0.12\linewidth}
        \vspace{5pt}
  		\includegraphics[width=1\linewidth]{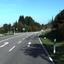}
  		\includegraphics[width=1\linewidth]{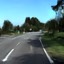}
  		\includegraphics[width=1\linewidth]{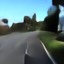}
	\end{subfigure}
	\begin{subfigure}{0.12\linewidth}
        \vspace{5pt}
  		\includegraphics[width=1\linewidth]{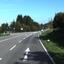}
  		\includegraphics[width=1\linewidth]{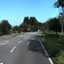}
  		\includegraphics[width=1\linewidth]{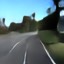}
	\end{subfigure}

    \caption{KITTI driving qualitative evaluation. Our highest capacity model (middle) is able to maintain the observed dynamics of driving forward and is able to generate unseen street lines and the moving background. The baseline (bottom) loses the street lines and the background becomes blurry. For best viewing and more results, please visit our website \url{https://cutt.ly/QGuCex}.}
\label{fig:qualitative_driving}
\end{figure}

\cutparagraphup
\paragraph{Qualitative evaluation.} 
In Figure \ref{fig:qualitative_driving}, we show a comparison between the largest stochastic model and its baseline. The baseline model starts becoming blurry as the predictions move forward in the future, and important features like the lane markings disappear. However, our biggest capacity model makes very sharp predictions that look realistic in comparison to the ground-truth.

\section{Higher resolution videos}
Finally, we experiment with higher resolution videos. We train SVG' on the Human 3.6M and KITTI driving datasets. These two datasets contain much larger resolution images compared to the Towel pick dataset, enabling us to sub-sample frames to twice the resolution of previous experiments (128x128). We follow the same protocol for the number of input and predicted time steps during training (5 inputs and 10 predictions), and the same protocol for testing (5 inputs and 25 predictions). In contrast to the networks used in the previous experiments, we add three more convolutional layers plus pooling to subsample the input to the same convolutional encoder output resolution as in previous experiments.

\begin{figure}[thbp!]

    \hspace*{-.5cm}
    \centering
	\begin{subfigure}{0.1\linewidth}
        \raggedleft
        \rotatebox{90}{
        \hspace{-.3cm}
        \parbox{1.6cm}{\centering Smallest model \\ (Baseline)}
        \parbox{1.6cm}{\centering Biggest model \\ (Ours)}
        \parbox{1.6cm}{\centering Groundtruth}
        }
    \end{subfigure}
    \begin{subfigure}{0.12\linewidth}
        \caption*{t=8}
        \vspace{-7pt}
  		\includegraphics[width=1\linewidth]{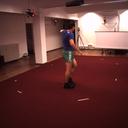}
  		\includegraphics[width=1\linewidth]{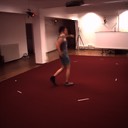}
  		\includegraphics[width=1\linewidth]{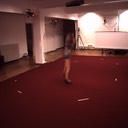}
	\end{subfigure} 
    \begin{subfigure}{0.12\linewidth}
        \caption*{t=11}
        \vspace{-7pt} 		\includegraphics[width=1\linewidth]{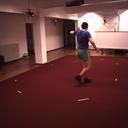}
  		\includegraphics[width=1\linewidth]{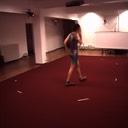}
  		\includegraphics[width=1\linewidth]{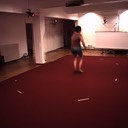}
	\end{subfigure}
    \begin{subfigure}{0.12\linewidth}
        \caption*{t=14}
        \vspace{-7pt}
  		\includegraphics[width=1\linewidth]{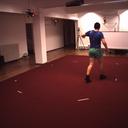}
  		\includegraphics[width=1\linewidth]{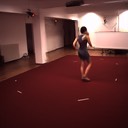}
  		\includegraphics[width=1\linewidth]{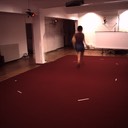}
	\end{subfigure} 
    \begin{subfigure}{0.12\linewidth}
        \caption*{t=17}
        \vspace{-7pt}
  		\includegraphics[width=1\linewidth]{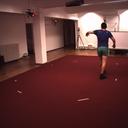}
  		\includegraphics[width=1\linewidth]{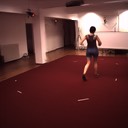}
  		\includegraphics[width=1\linewidth]{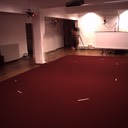}
	\end{subfigure}
	\begin{subfigure}{0.12\linewidth}
        \caption*{t=20}
        \vspace{-7pt}
  		\includegraphics[width=1\linewidth]{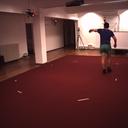}
  		\includegraphics[width=1\linewidth]{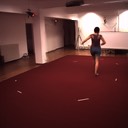}
  		\includegraphics[width=1\linewidth]{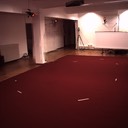}
	\end{subfigure}
	\begin{subfigure}{0.12\linewidth}
        \caption*{t=23}
        \vspace{-7pt}
  		\includegraphics[width=1\linewidth]{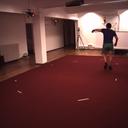}
  		\includegraphics[width=1\linewidth]{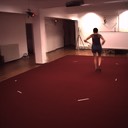}
  		\includegraphics[width=1\linewidth]{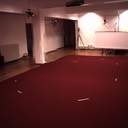}
	\end{subfigure}
	\begin{subfigure}{0.12\linewidth}
        \caption*{t=26}
        \vspace{-7pt}
  		\includegraphics[width=1\linewidth]{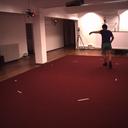}
  		\includegraphics[width=1\linewidth]{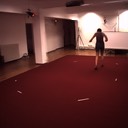}
  		\includegraphics[width=1\linewidth]{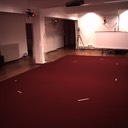}
	\end{subfigure}

    \hspace{-14pt}
    \centering
	\begin{subfigure}{0.1\linewidth}
        \raggedleft
        \rotatebox{90}{
        \hspace{-.5cm}
        \parbox{1.6cm}{\centering Smallest model \\ (Baseline)}
        \parbox{1.6cm}{\centering Biggest model \\ (Ours)}
        \parbox{1.6cm}{\centering Groundtruth}
        }
    \end{subfigure}
    \begin{subfigure}{0.12\linewidth}
        \vspace{5pt}
  		\includegraphics[width=1\linewidth]{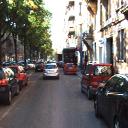}
  		\includegraphics[width=1\linewidth]{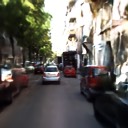}
  		\includegraphics[width=1\linewidth]{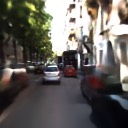}
	\end{subfigure} 
    \begin{subfigure}{0.12\linewidth}
        \vspace{5pt}  		\includegraphics[width=1\linewidth]{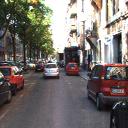}
  		\includegraphics[width=1\linewidth]{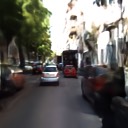}
  		\includegraphics[width=1\linewidth]{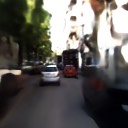}
	\end{subfigure}
    \begin{subfigure}{0.12\linewidth}
        \vspace{5pt}
  		\includegraphics[width=1\linewidth]{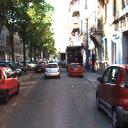}
  		\includegraphics[width=1\linewidth]{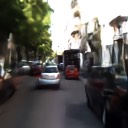}
  		\includegraphics[width=1\linewidth]{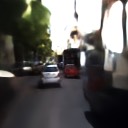}
	\end{subfigure} 
    \begin{subfigure}{0.12\linewidth}
        \vspace{5pt}
  		\includegraphics[width=1\linewidth]{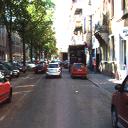}
  		\includegraphics[width=1\linewidth]{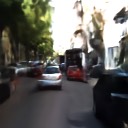}
  		\includegraphics[width=1\linewidth]{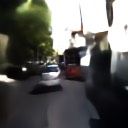}
	\end{subfigure}
	\begin{subfigure}{0.12\linewidth}
        \vspace{5pt}
  		\includegraphics[width=1\linewidth]{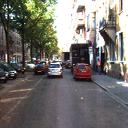}
  		\includegraphics[width=1\linewidth]{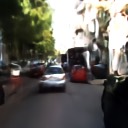}
  		\includegraphics[width=1\linewidth]{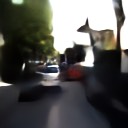}
	\end{subfigure}
	\begin{subfigure}{0.12\linewidth}
        \vspace{5pt}
  		\includegraphics[width=1\linewidth]{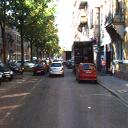}
  		\includegraphics[width=1\linewidth]{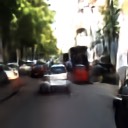}
  		\includegraphics[width=1\linewidth]{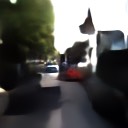}
	\end{subfigure}
	\begin{subfigure}{0.12\linewidth}
        \vspace{5pt}
  		\includegraphics[width=1\linewidth]{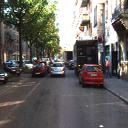}
  		\includegraphics[width=1\linewidth]{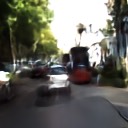}
  		\includegraphics[width=1\linewidth]{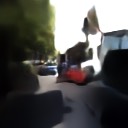}
	\end{subfigure}

    \caption{Human 3.6M and KITTI driving qualitative evaluation on high resolution videos (frame size of 128x128) with comparison between smallest model and largest model we were able to train (M=3, K=3). For best viewing and more results, please visit our website \url{https://cutt.ly/QGuCex}.}
\label{fig:qualitative_highres}
\end{figure}

In Figure \ref{fig:qualitative_highres} we show qualitative results comparing the smallest (baseline) and biggest (Ours) networks. The biggest network we were able to train had a configuration of M=3 and K=3. Higher resolution videos contain more details about the pixel dynamics observed in the frames. This enables the models to have a denser signal, and so, the generated videos become more difficult to distinguish from real videos. Therefore, this result suggests that besides training better and bigger models, we should also more towards larger resolutions. For more examples of videos, please visit our website: \url{https://cutt.ly/QGuCex}.

\cutsectionup
\section{Conclusion}
\cutsectiondown
In conclusion, we provide a full empirical study on the effect of finding minimal inductive bias and increasing model capacity for video generation. 
We perform a rigorous evaluation with five different metrics to analyze which types of inductive bias are important for generating accurate video dynamics, when combined with large model capacity.
Our experiments confirm the importance of recurrent connections and modeling stochasticity in the presence of uncertainty (e.g., videos with unknown action or control).
We also find that maximizing the capacity of such models improves the quality of video prediction.
%
We hope our work encourages the field to push along similar directions in the future -- i.e., to see how far we can get by finding the right combination of minimal inductive bias and maximal model capacity for achieving high quality video prediction.

\clearpage
\bibliographystyle{plainnat}
\bibliography{main}

\clearpage
\appendix
\section{Supplementary material}
\subsection{Video results}
We have provided video comparisons of the baseline and largest model for the best two models (LSTM and SVG') in this website: \url{https://cutt.ly/QGuCex}.
\subsection{Per-frame evaluation comparison as model capacity increases}
In this section, we present a per-frame evaluation for capacities in each of the models we experiment in our paper. 

\subsubsection{Robot arm.} \label{supp:robot_arm}
The plots show a slight improvement as the number of parameters increase for the CNN architecture. However, for the LSTM and SVG' architectures the improvement is more noticeable. We hypothesize that this is due to the model being able to better handle the robot arm interaction with the objects by having a large capacity.
\begin{figure}[htp!]
    \centering
    \hspace{-8pt}
	\includegraphics[width=.35\linewidth]{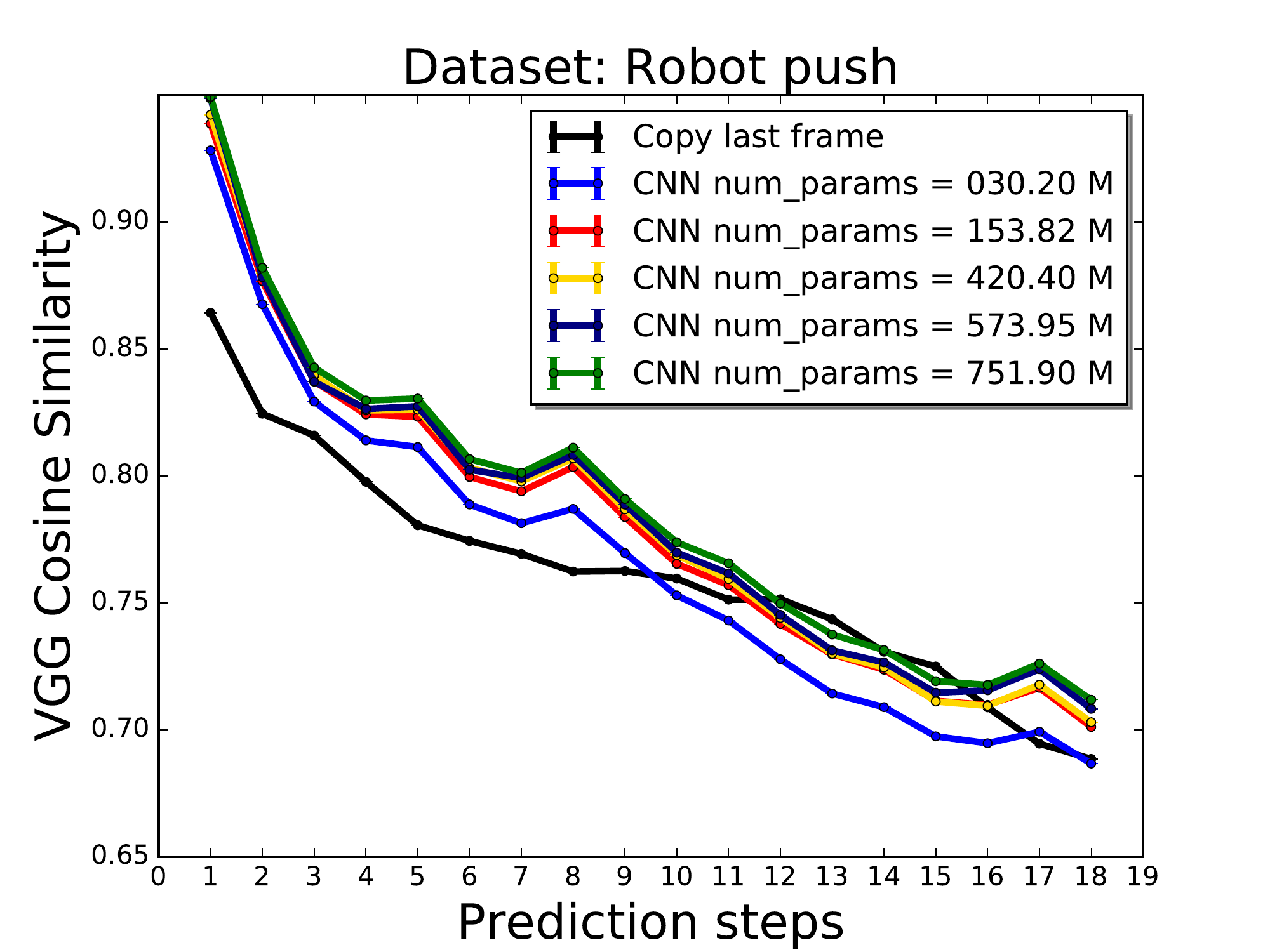} \hspace{-14pt}
	\includegraphics[width=.35\linewidth]{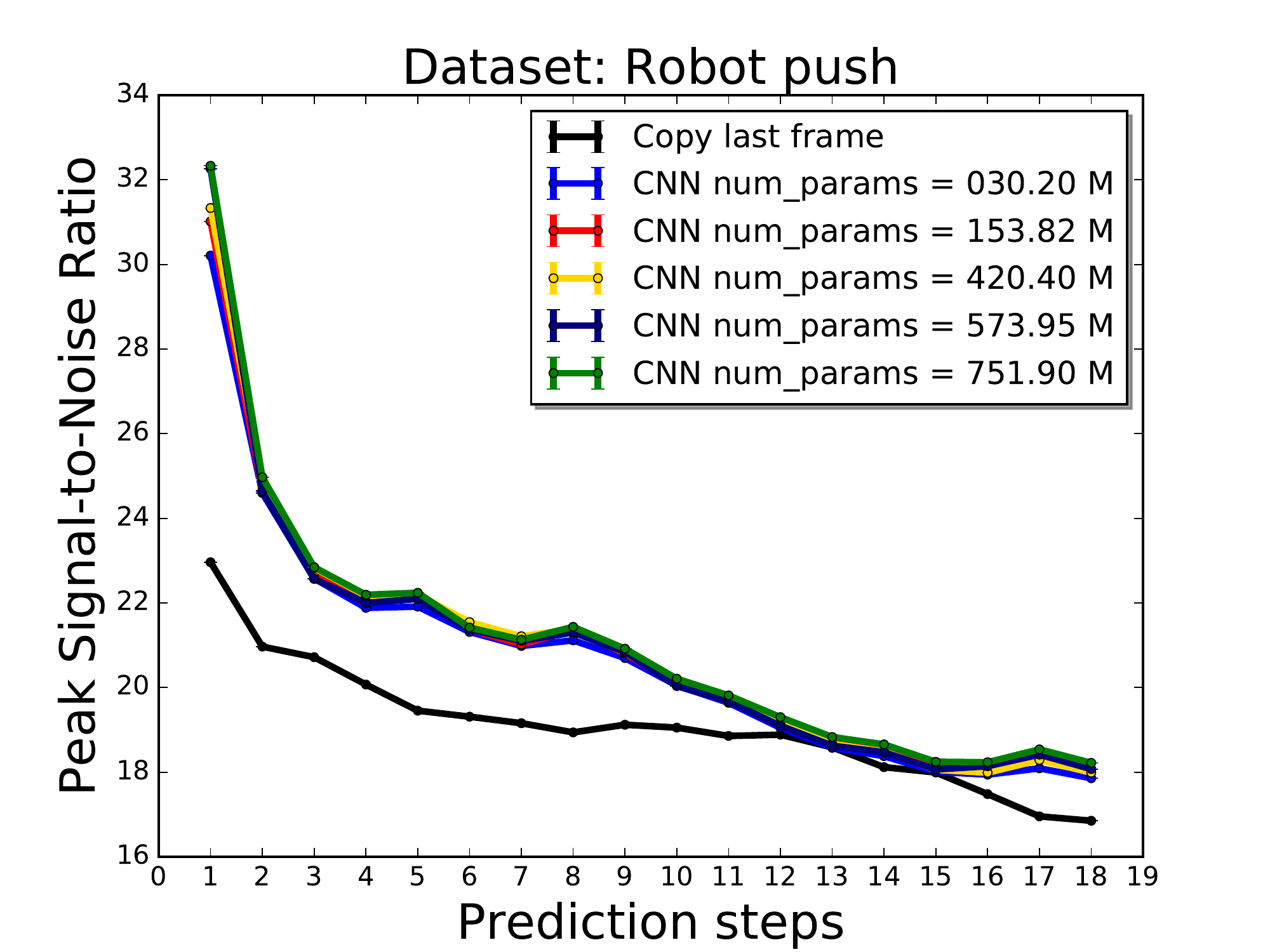} \hspace{-14pt}
	\includegraphics[width=.35\linewidth]{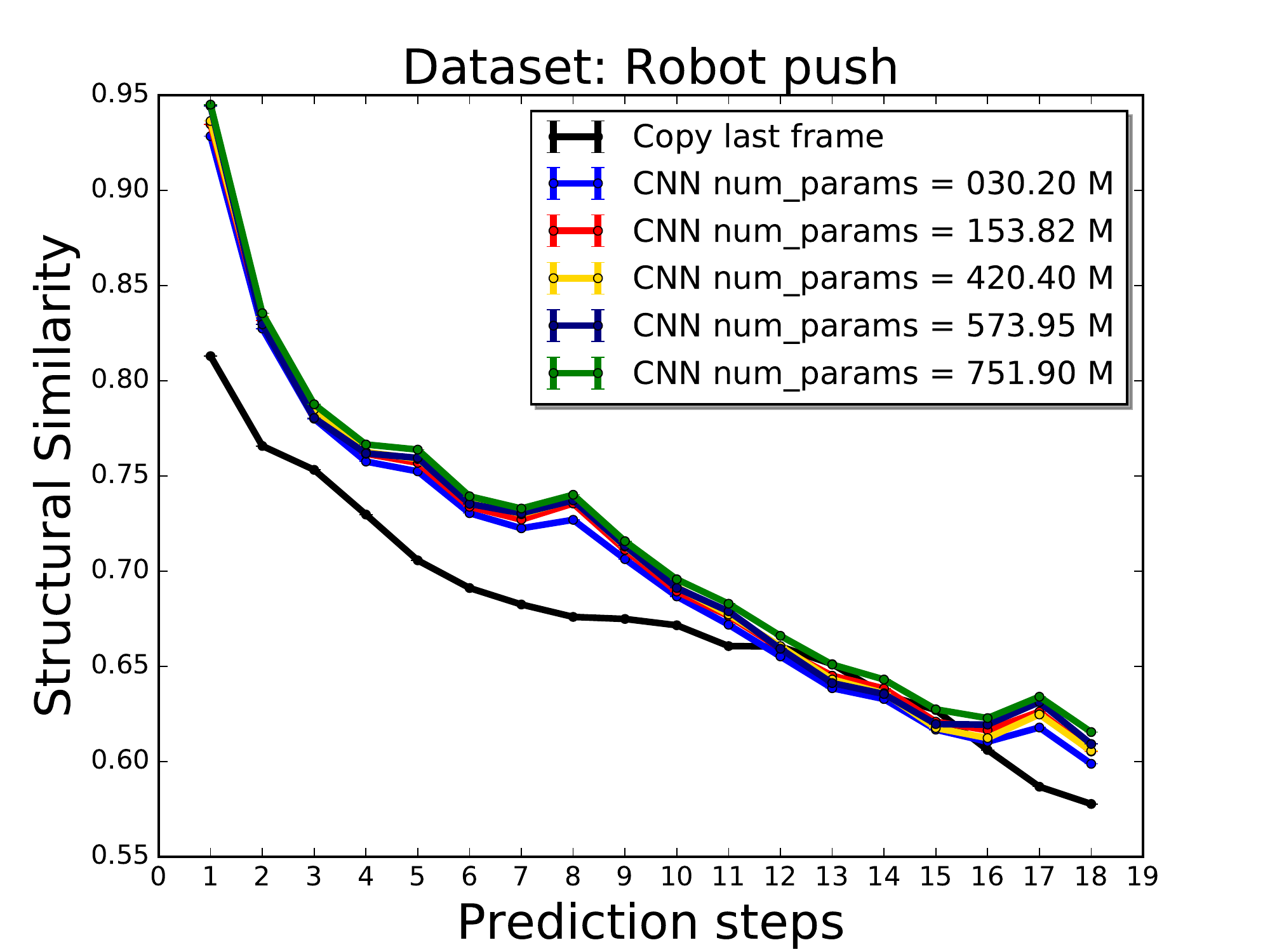}
\label{fig:frame_towel_cnn}
\end{figure}

\begin{figure}[htp!]
    \centering
    \hspace{-8pt}
	\includegraphics[width=.35\linewidth]{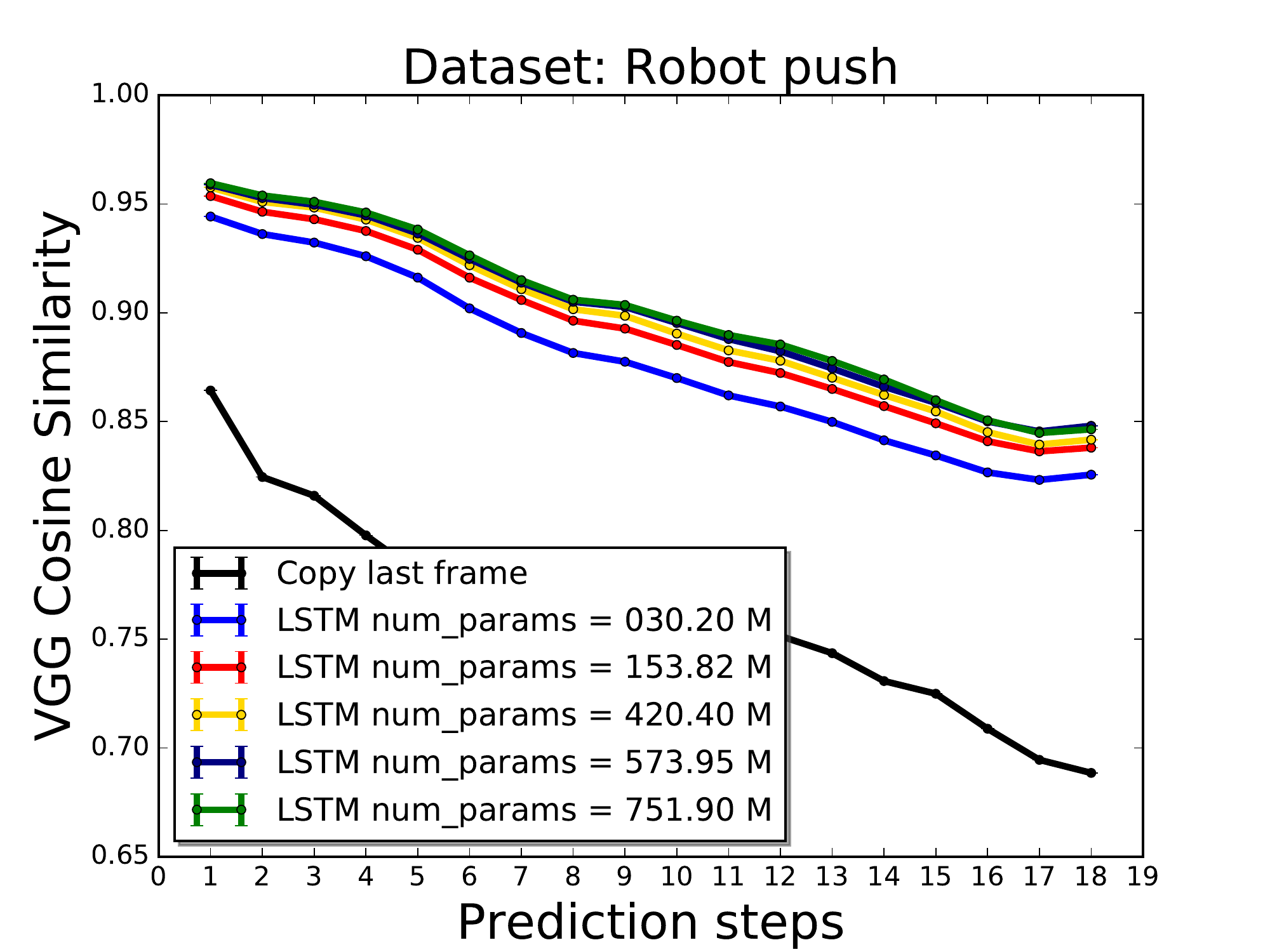} \hspace{-14pt}
	\includegraphics[width=.35\linewidth]{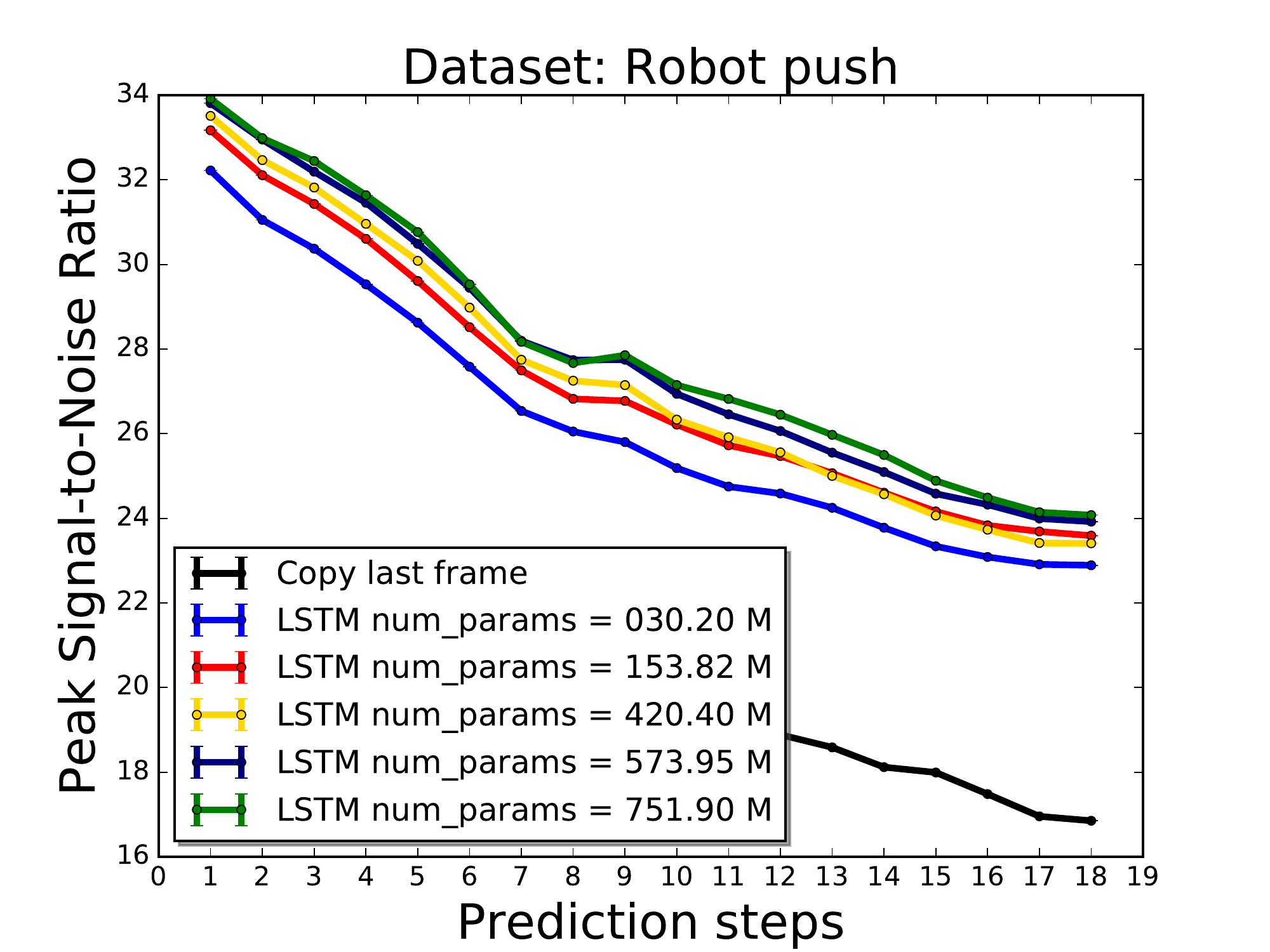} \hspace{-14pt}
	\includegraphics[width=.35\linewidth]{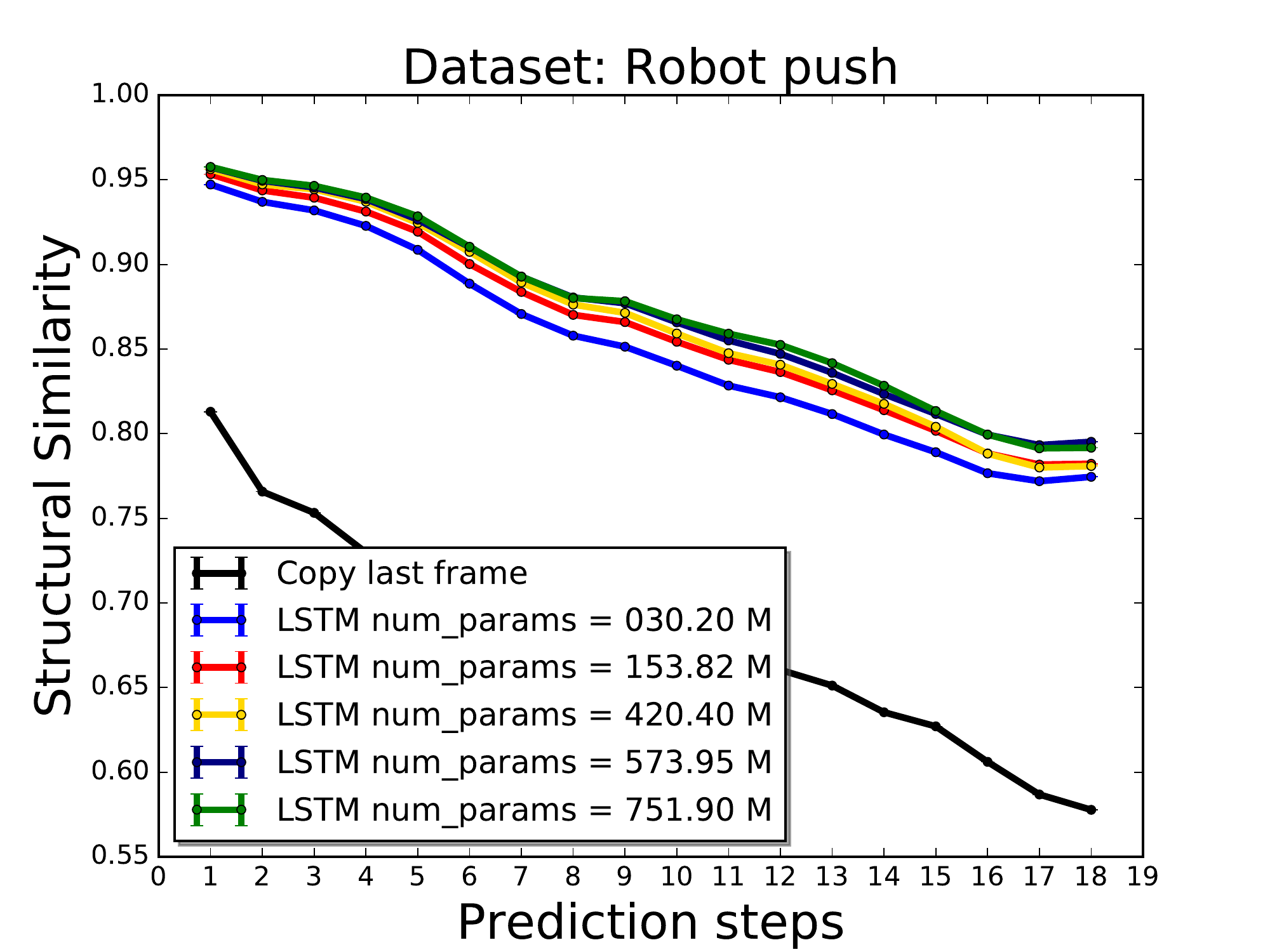}
\label{fig:frame_towel_lstm}
\end{figure}

\begin{figure}[htp!]
    \centering
    \hspace{-8pt}
	\includegraphics[width=.35\linewidth]{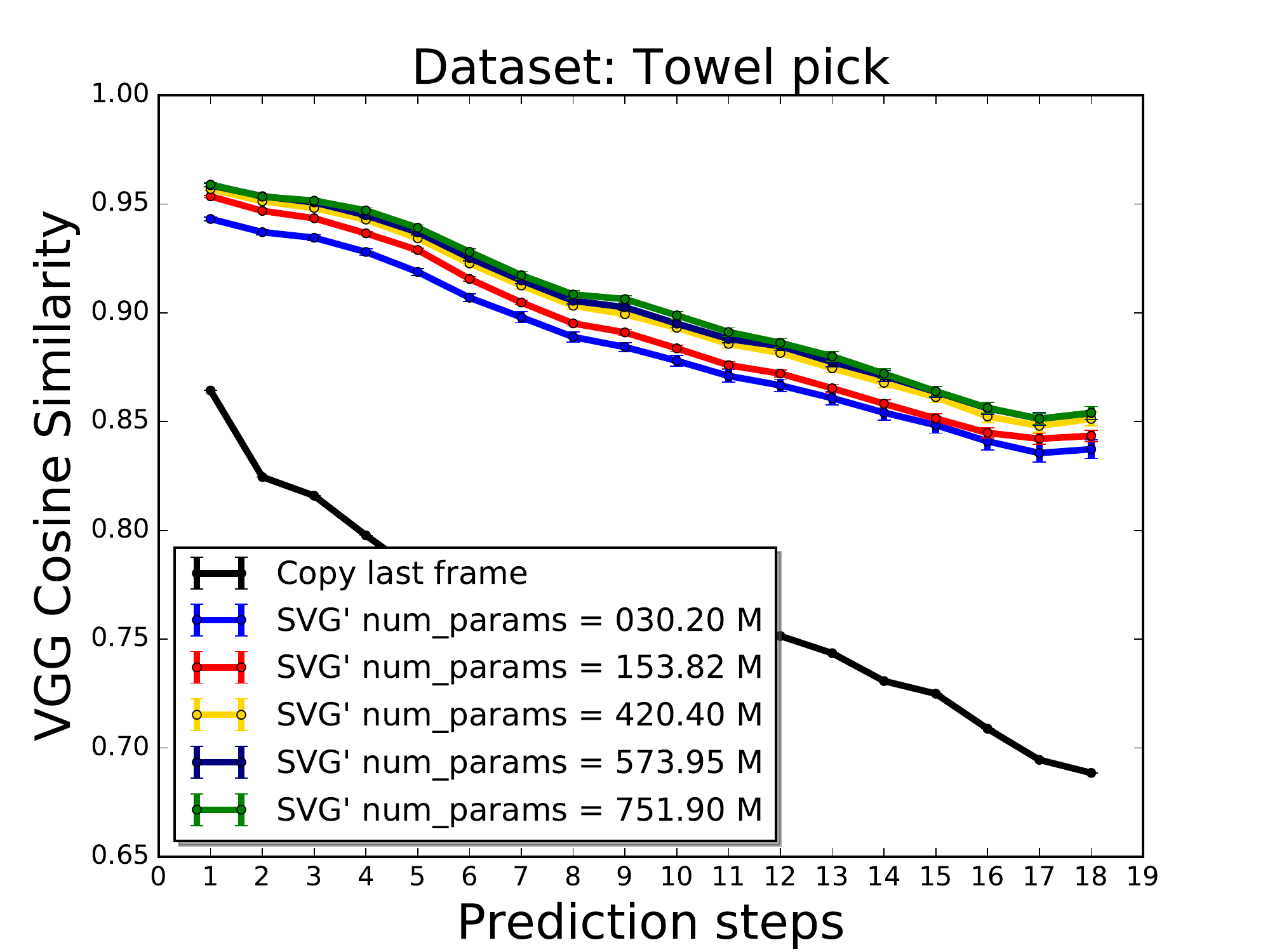} \hspace{-14pt}
	\includegraphics[width=.35\linewidth]{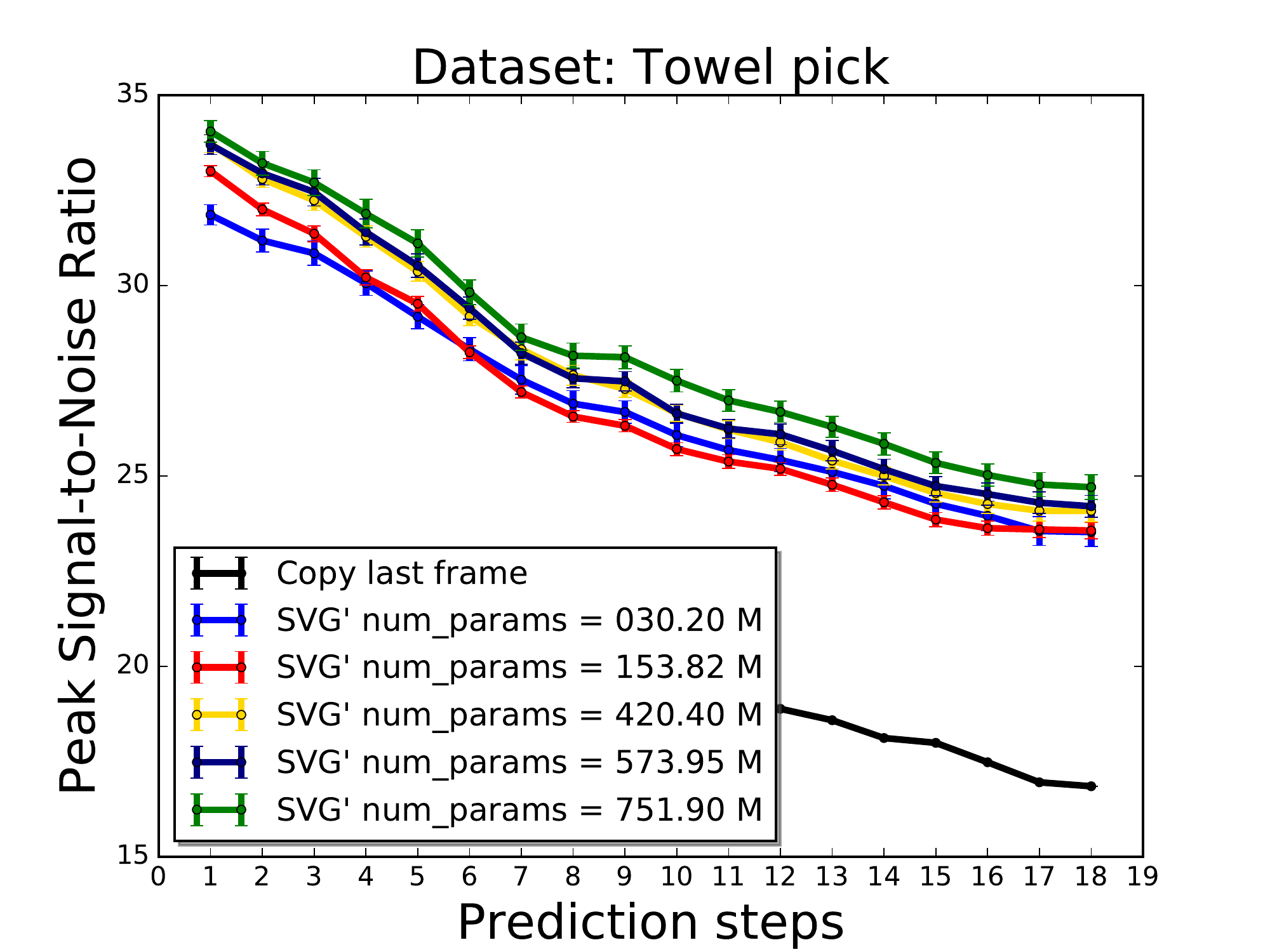} \hspace{-14pt}
	\includegraphics[width=.35\linewidth]{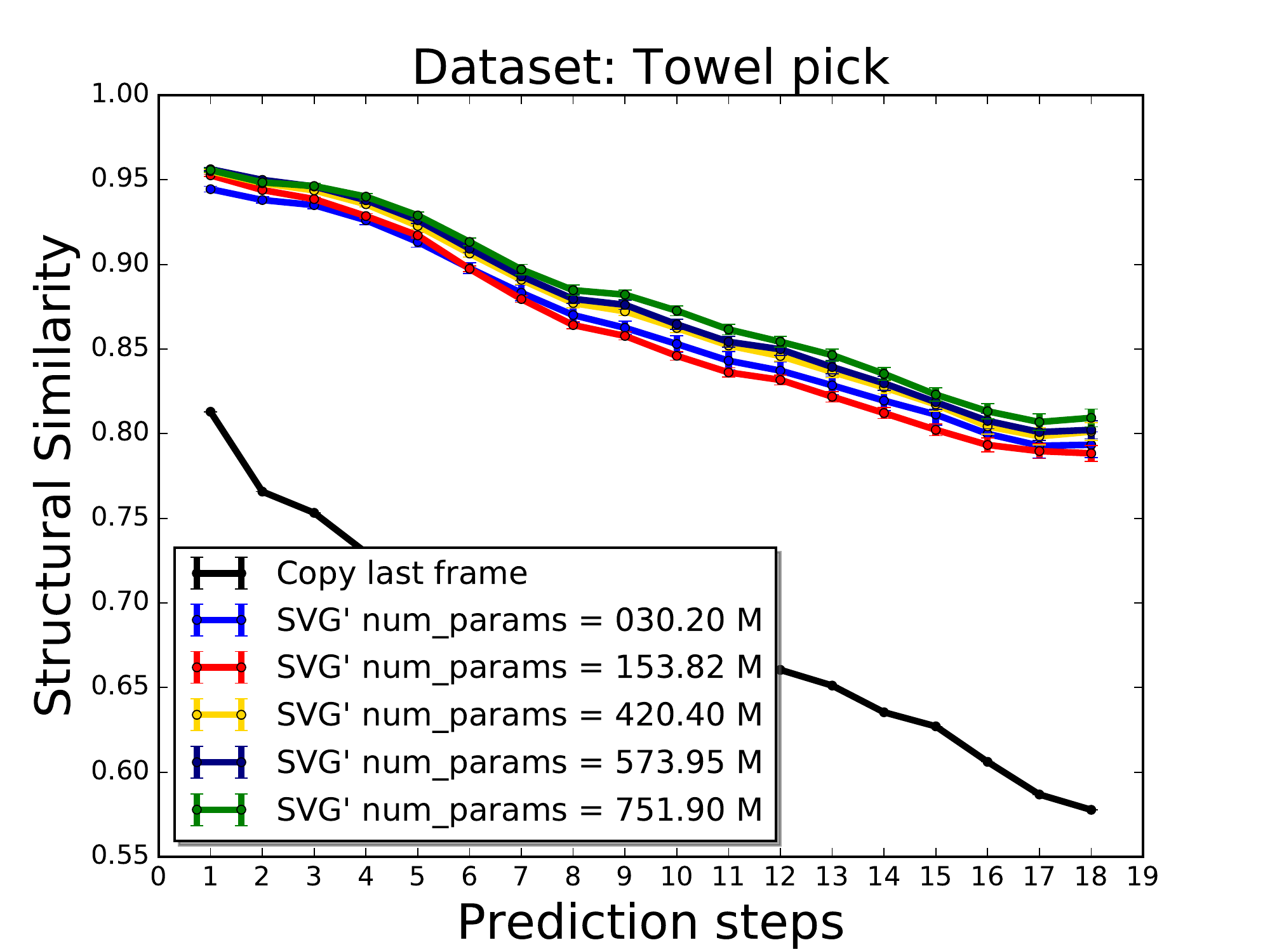}
	\caption{Towel pick per-frame evaluation (higher is better). As capacity increases, the per frame evaluation metrics become better. The increase is due to better modeling of interactions. The objects become sharper, and robot arm dynamics become better as the model capacity increases.}
\label{fig:frame_towel_svg}
\end{figure}

\clearpage
\subsubsection{Human activities.} \label{supp:humans}
The Human 3.6M dataset is mostly made of static background and the moving human occupies a relatively very small area of the frame. Therefore, models that are not capable of perfectly predicting the background become affected by this. To show our point, we include a baseline where we simply copy the last observed frame through time. This baseline significantly outperforms all models. Therefore, from these results we can conclude that per-frame evaluations are not reliable when a large portion of a video does not move.

\begin{figure}[htp!]
    \centering
    \hspace{-8pt}
	\includegraphics[width=.35\linewidth]{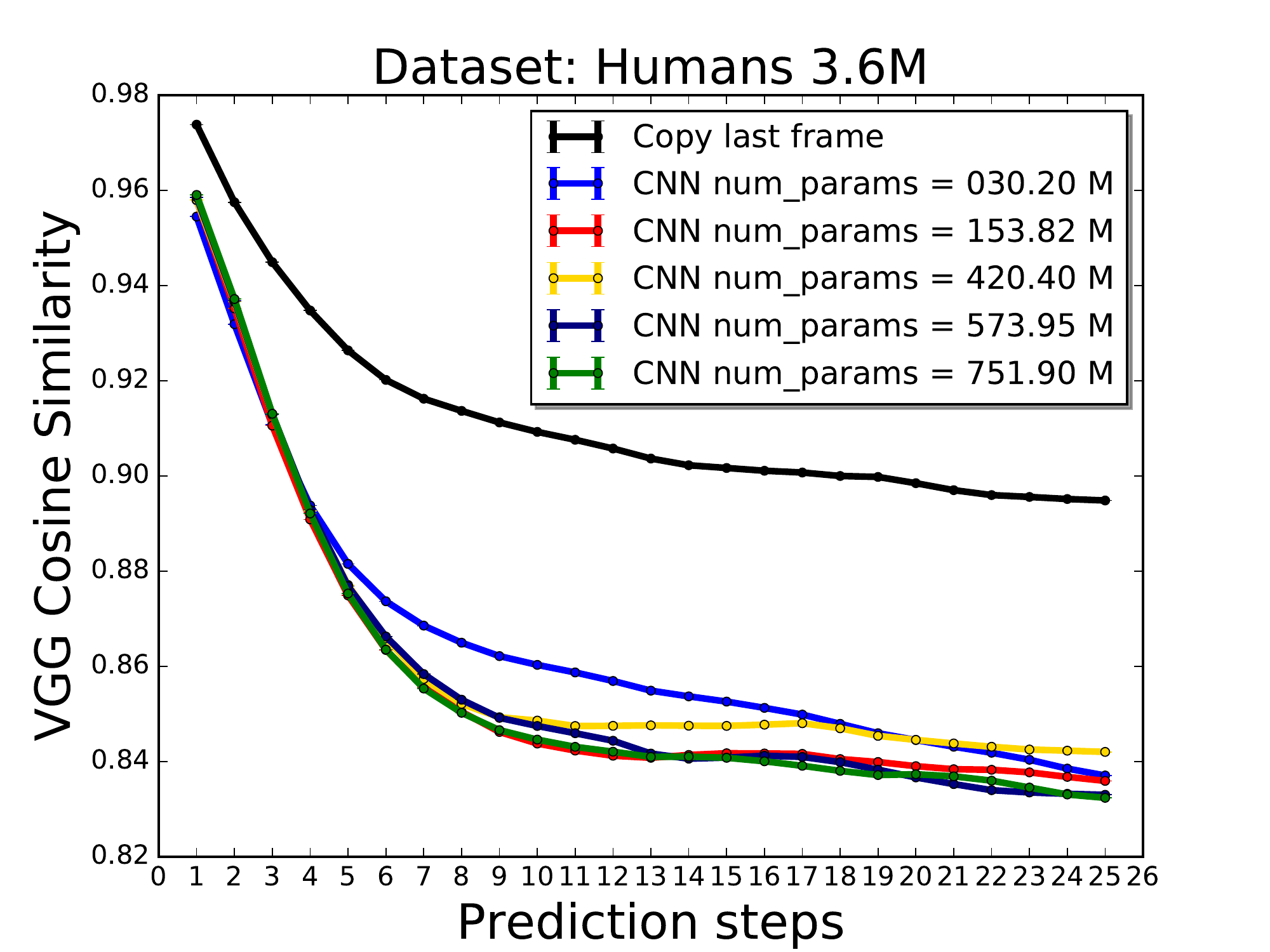} \hspace{-14pt}
	\includegraphics[width=.35\linewidth]{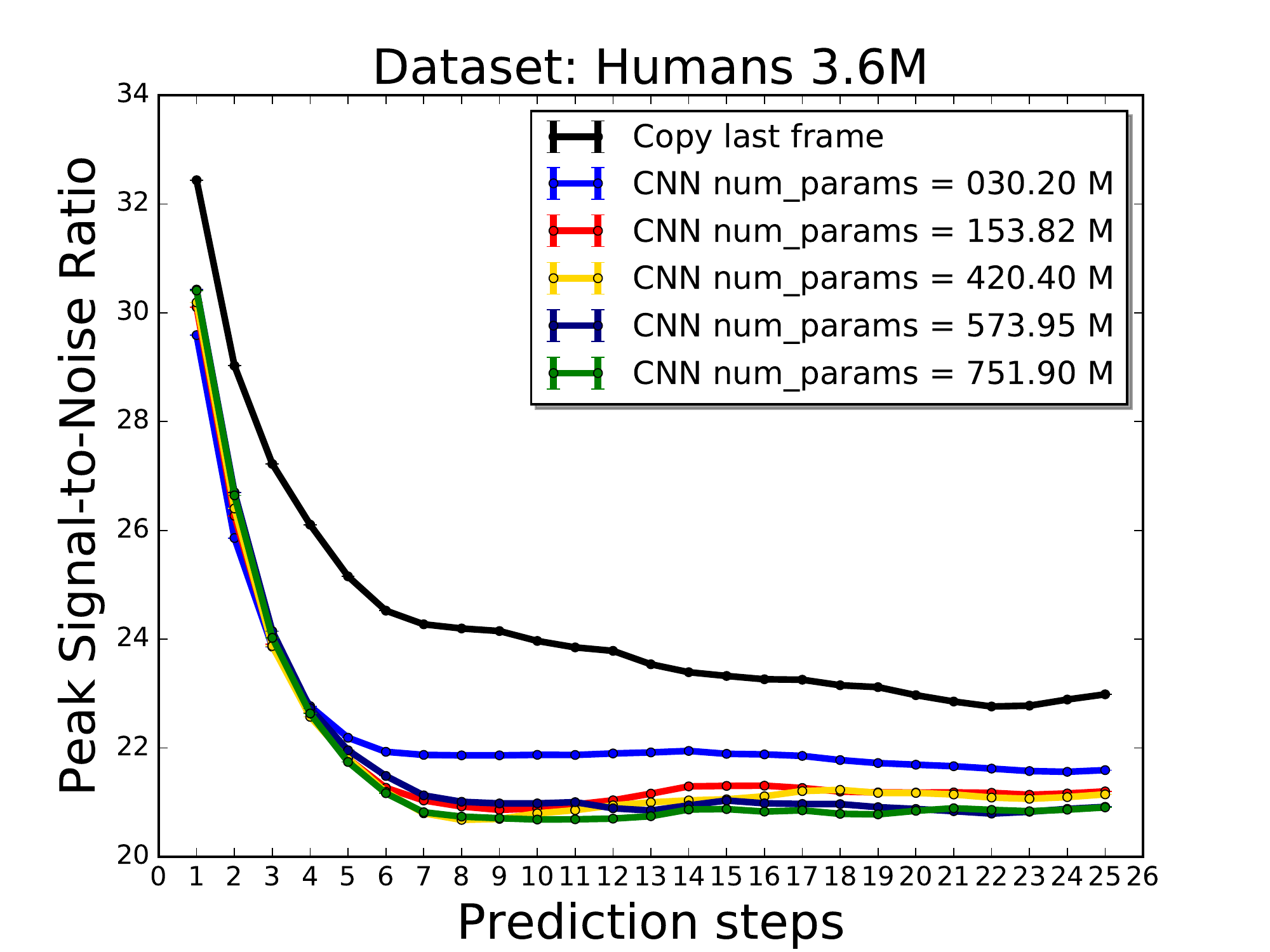} \hspace{-14pt}
	\includegraphics[width=.35\linewidth]{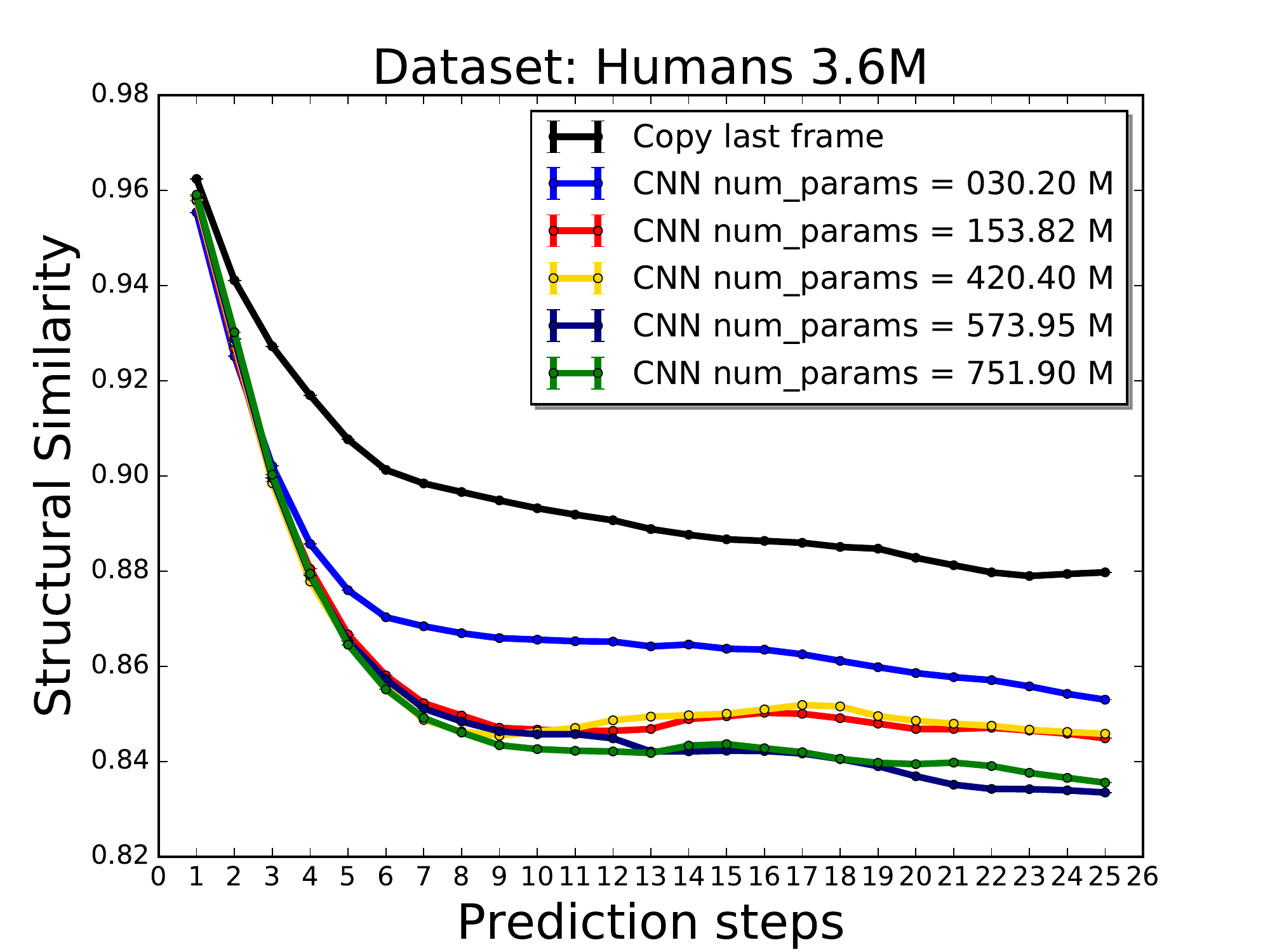}
\label{fig:frame_humans_cnn}
\end{figure}

\begin{figure}[htp!]
    \centering
    \hspace{-8pt}
	\includegraphics[width=.35\linewidth]{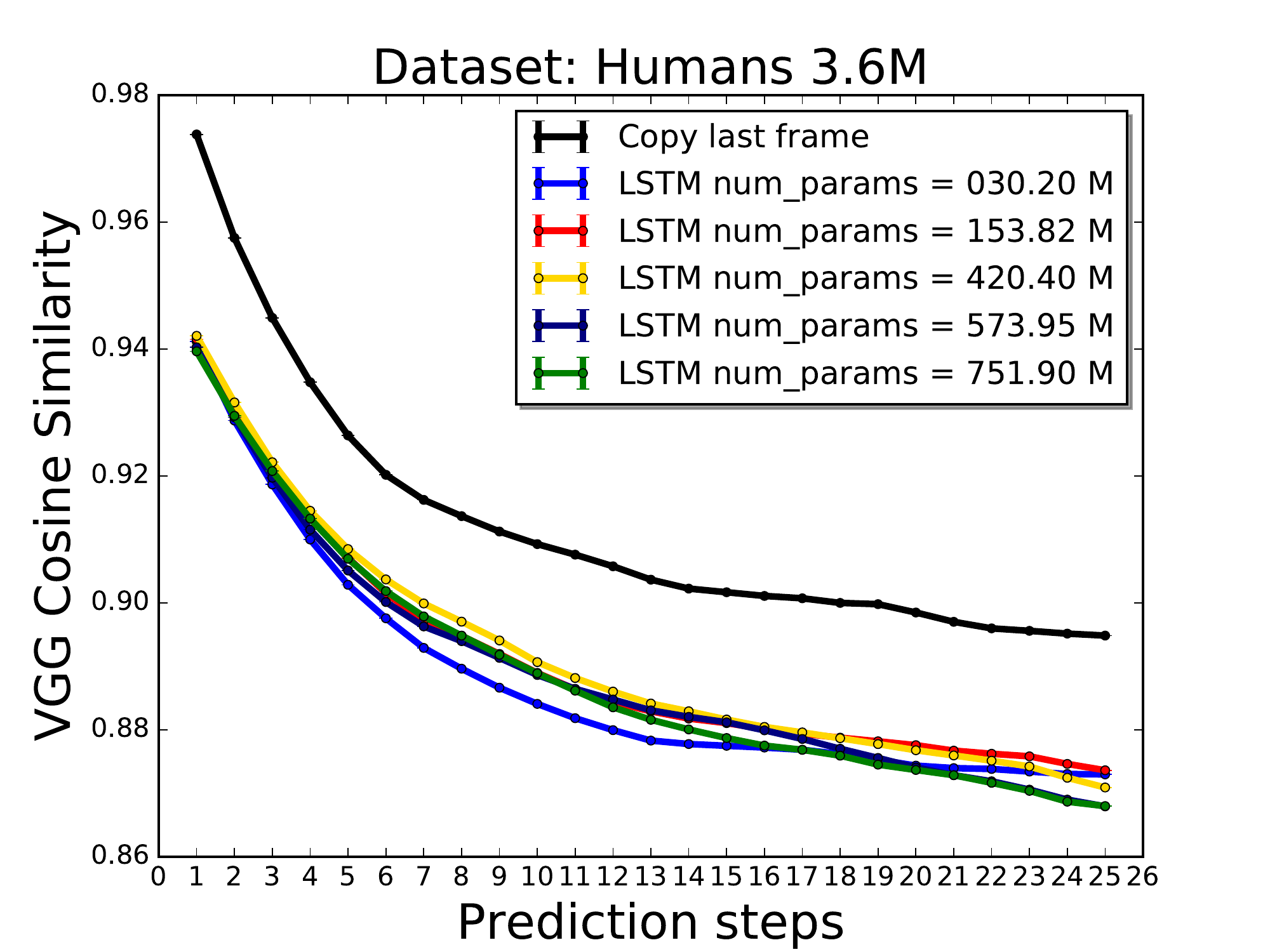} \hspace{-14pt}
	\includegraphics[width=.35\linewidth]{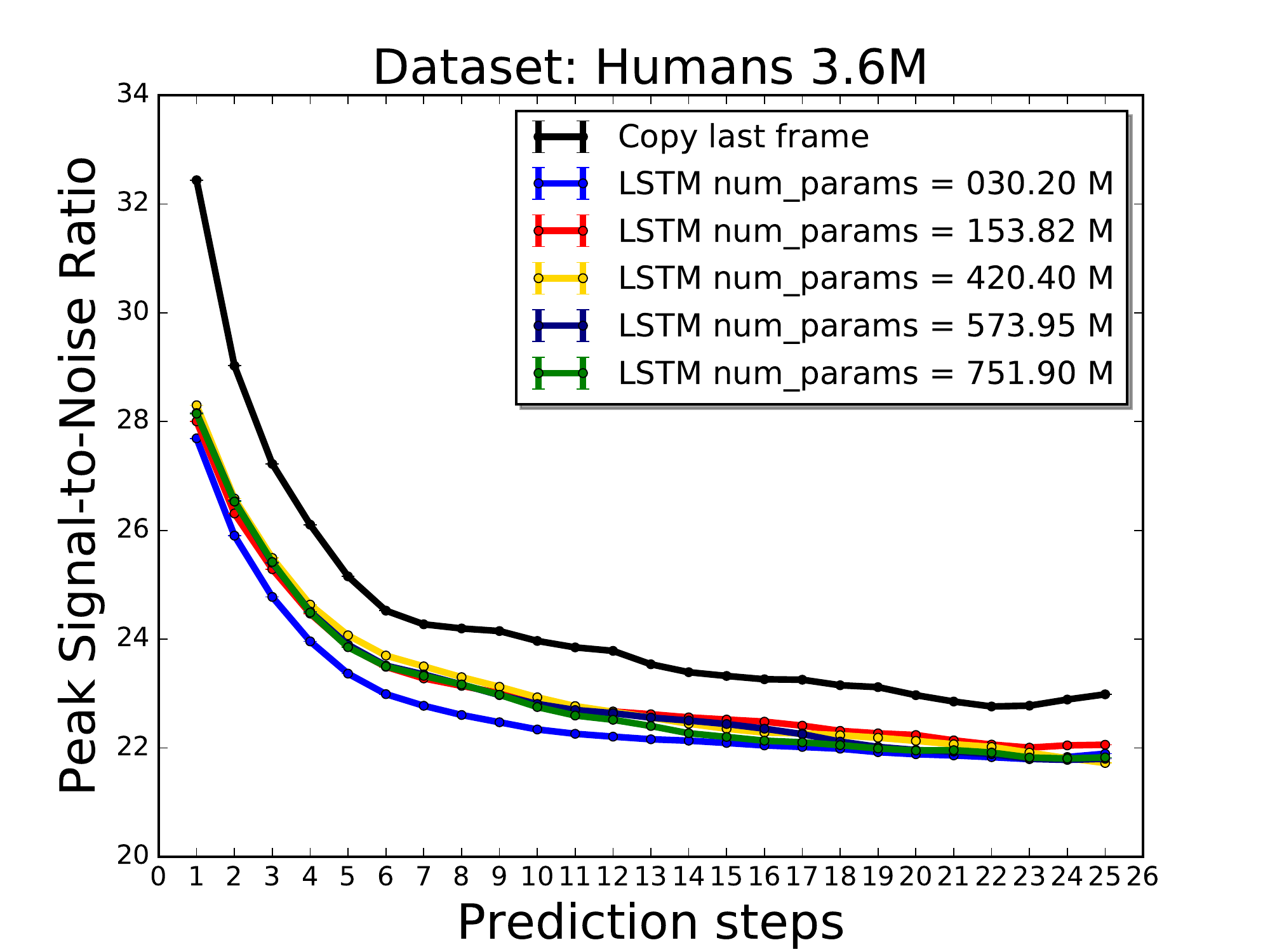} \hspace{-14pt}
	\includegraphics[width=.35\linewidth]{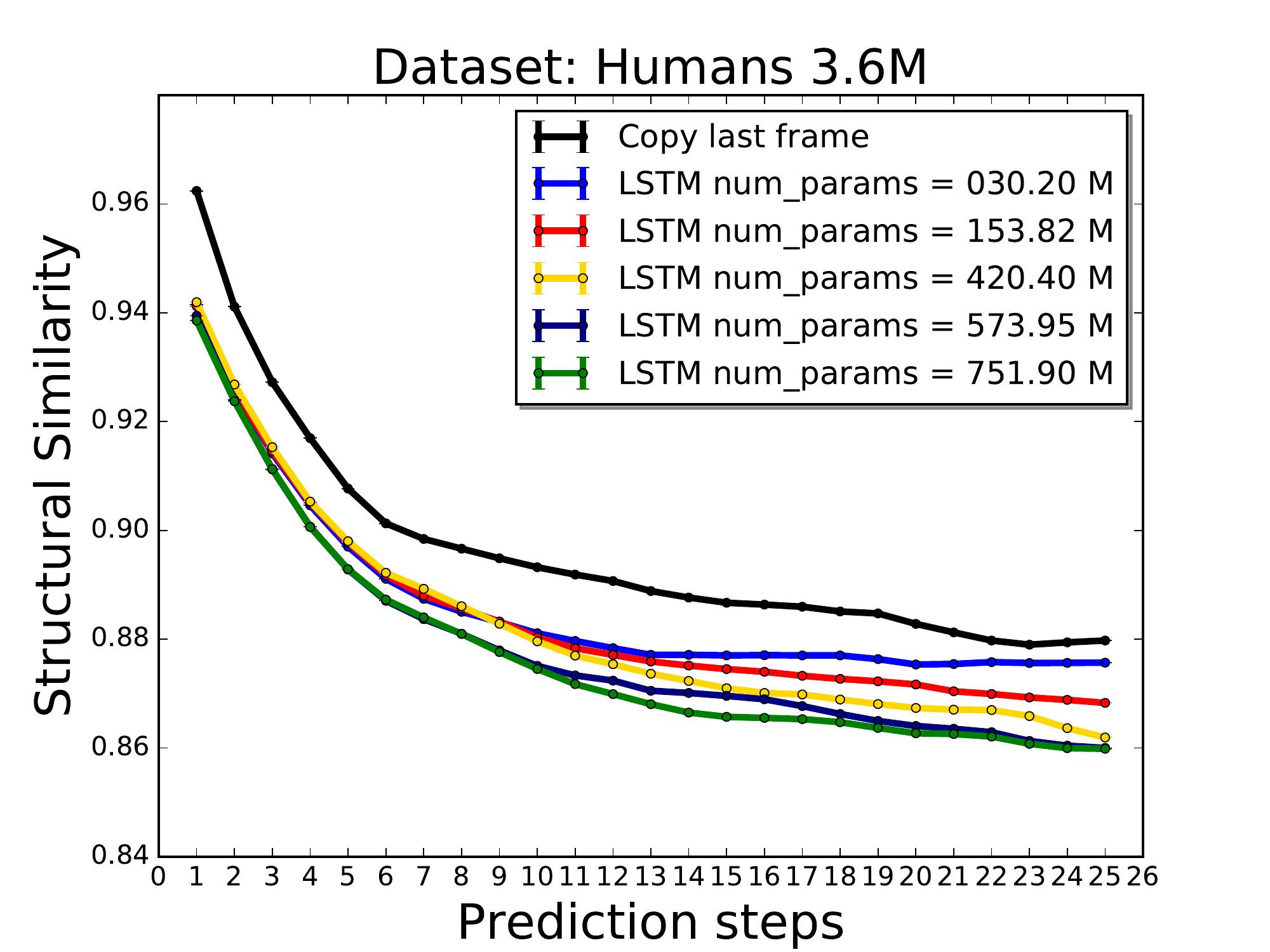}
\label{fig:frame_humans_lstm}
\end{figure}

\begin{figure}[htp!]
    \centering
    \hspace{-8pt}
	\includegraphics[width=.35\linewidth]{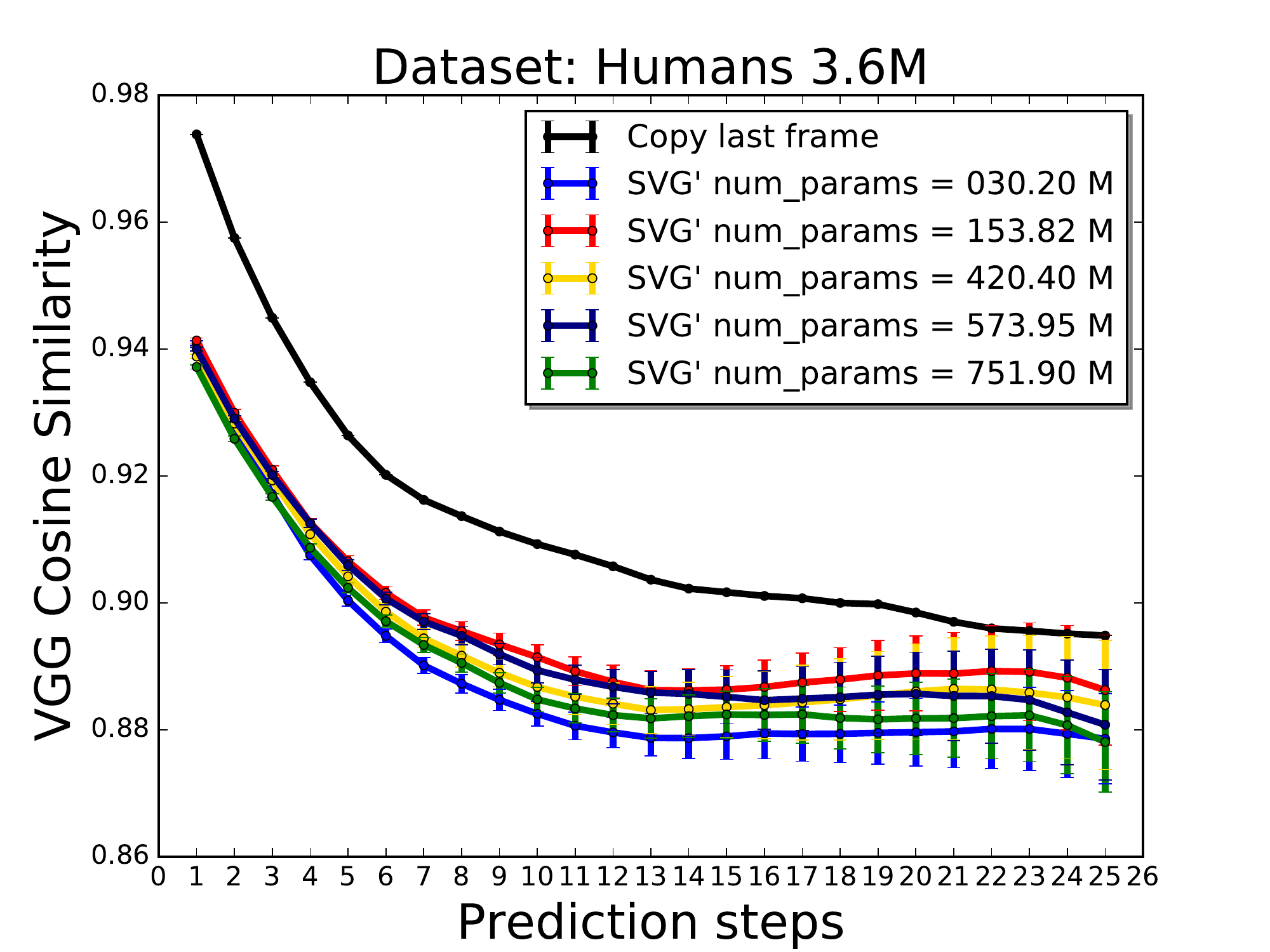} \hspace{-14pt}
	\includegraphics[width=.35\linewidth]{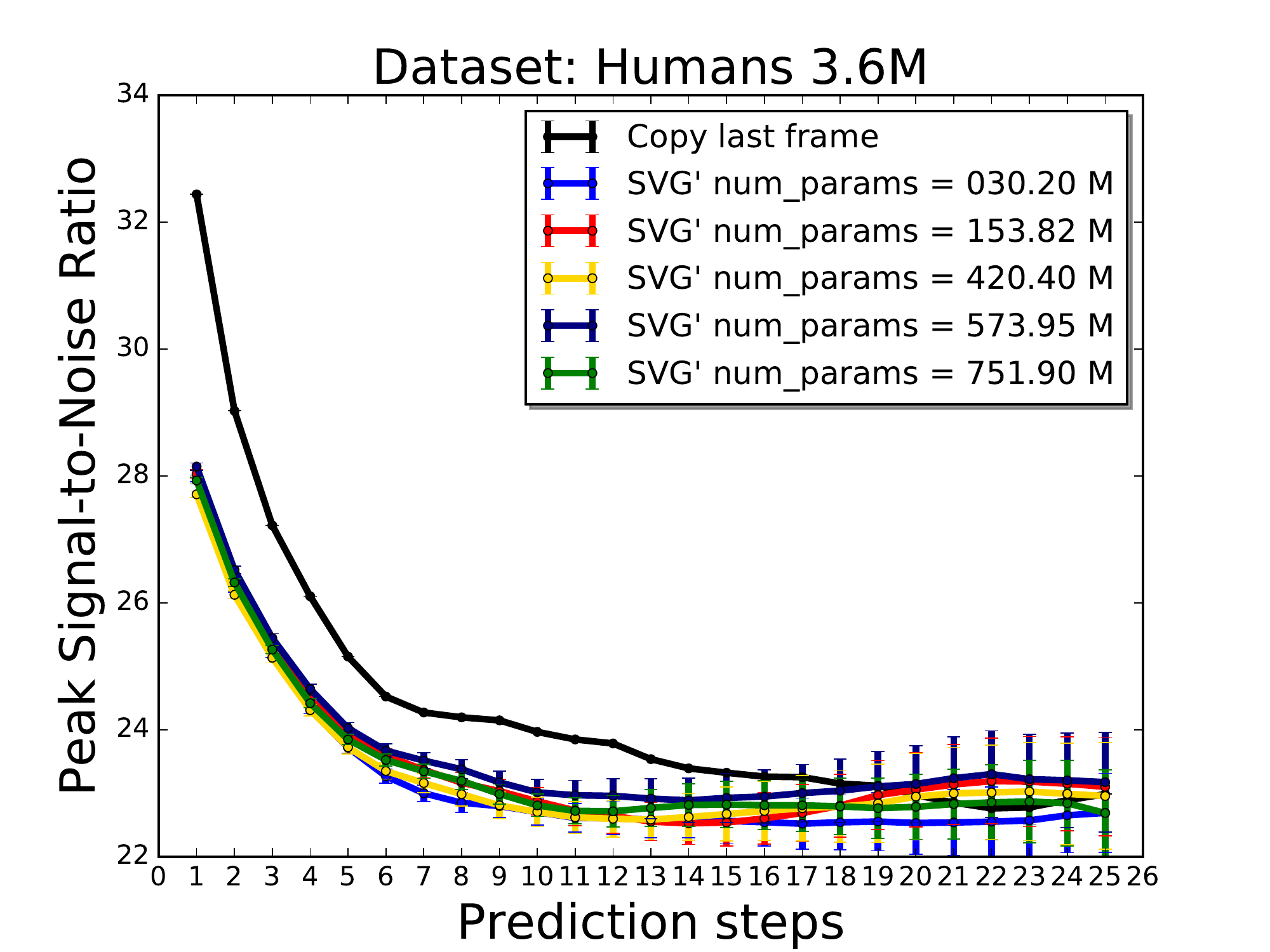} \hspace{-14pt}
	\includegraphics[width=.35\linewidth]{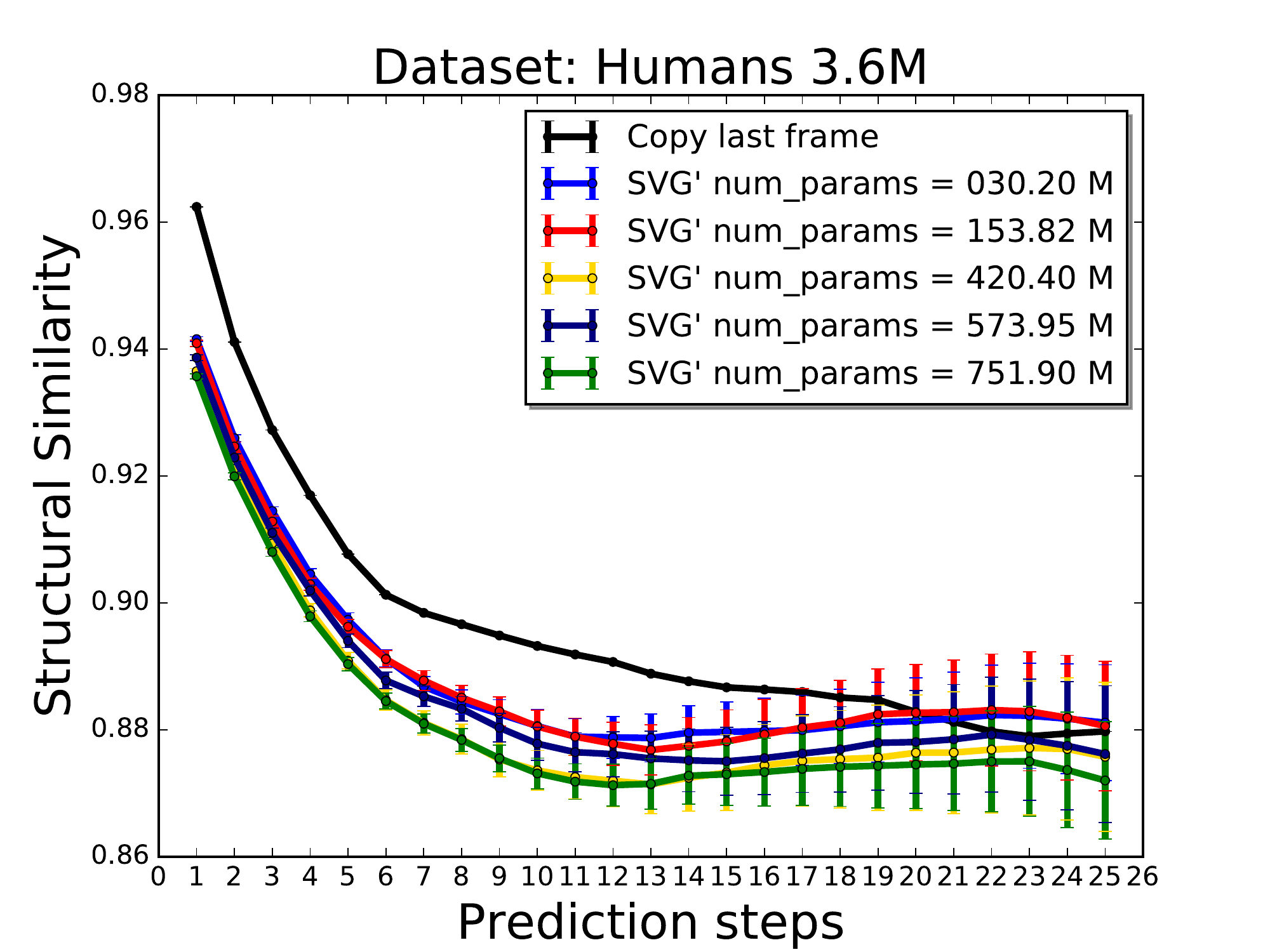}
	\caption{Human 3.6M per-frame evaluation (higher is better). In this dataset, there is a large amount of non-moving background that causes a per-frame evaluation to become not reliable. This is shown by the baseline based on simply copying the last observed frame through time which significantly outperforms all methods.}
\label{fig:frame_humans_svg}
\end{figure}

\clearpage
\subsubsection{Car driving.} \label{supp:driving}
In this dataset, as observed by the FVD measure in the main text, we see that the CNN model fails to make improvement in the per-frame evaluation metrics. However, the LSTM and SVG' models performance improves as the capacity of the models increases. The metric in which this is the most obvious is the VGG Cosine Similarity. This may be due to the partial observability of the dataset which makes it very difficult to predict exact pixels into the future, and so, PSNR and SSIM do not result in a large gap between the larger and baseline models. However, VGG Cosine Similarity compares high-level features of the predicted frames. Therefore, even if the predicted pixels are not exact, the predicted structures in the frames may be similar to those the ground-truth future. For this dataset, we do not present a copy last frame baseline because most pixels move (in contrast to the robot arm and Human 3.6M dataset, where many pixels stay fixed).

\begin{figure}[htp!]
    \centering
    \hspace{-8pt}
	\includegraphics[width=.35\linewidth]{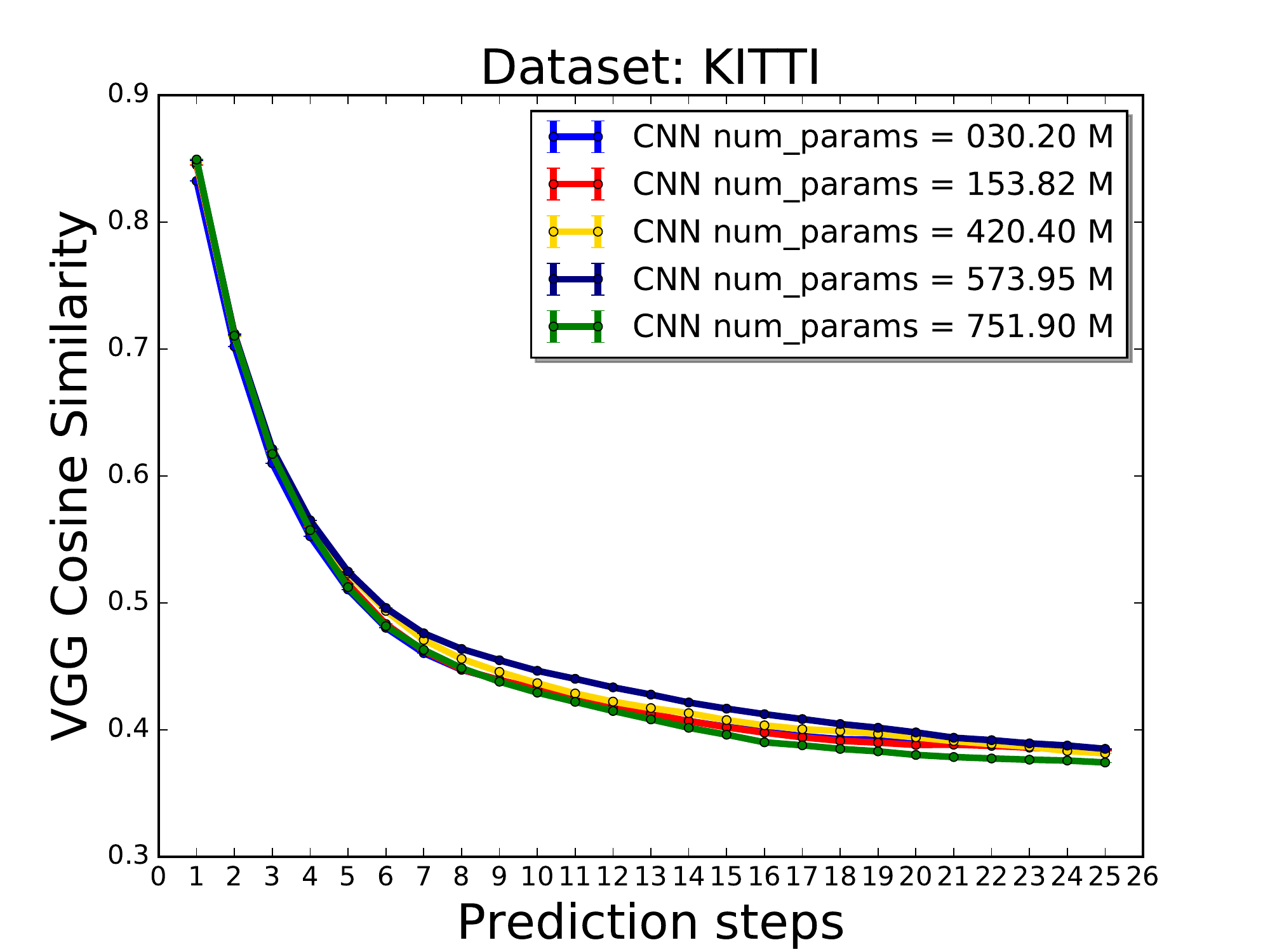} \hspace{-14pt}
	\includegraphics[width=.35\linewidth]{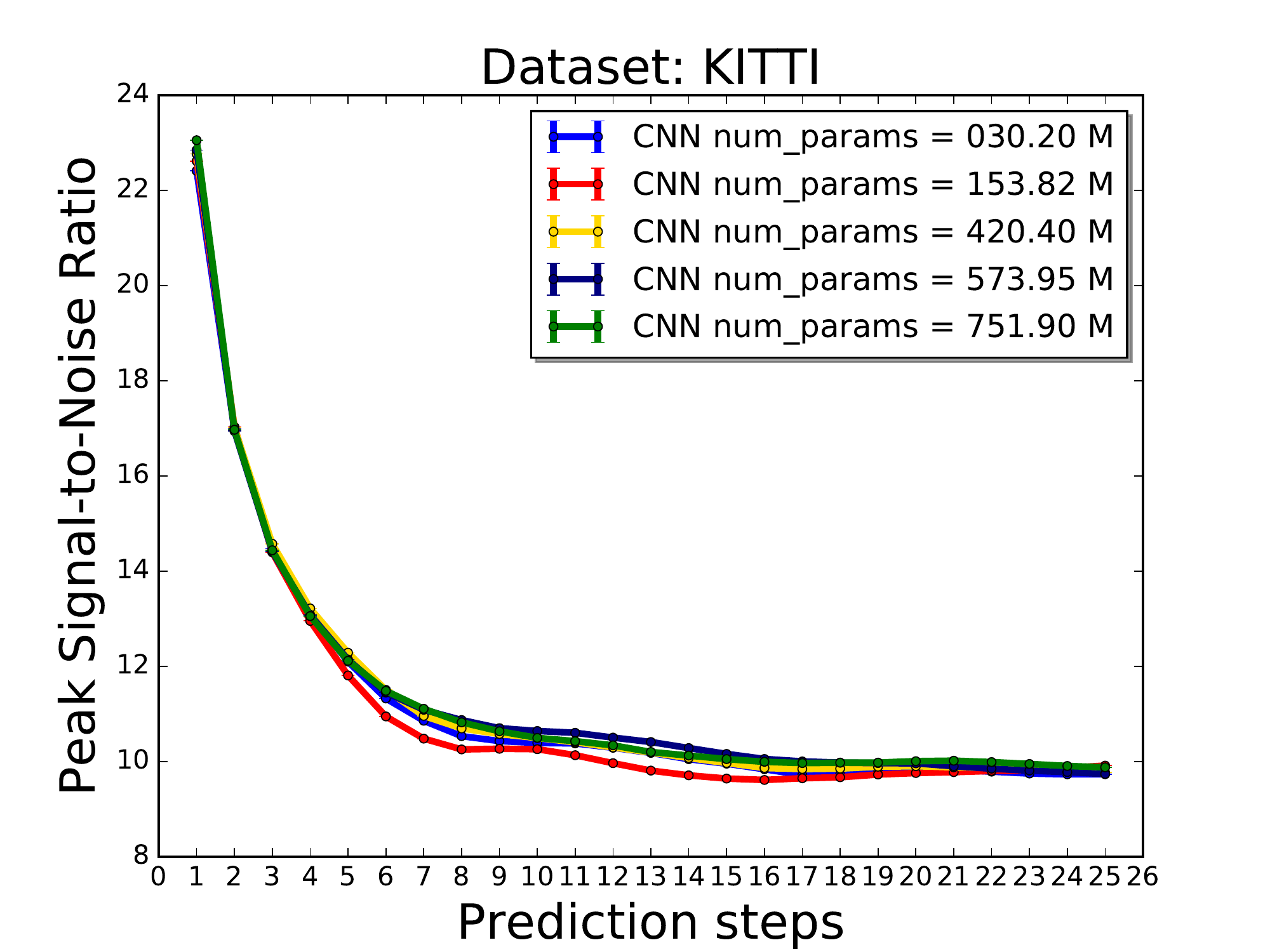} \hspace{-14pt}
	\includegraphics[width=.35\linewidth]{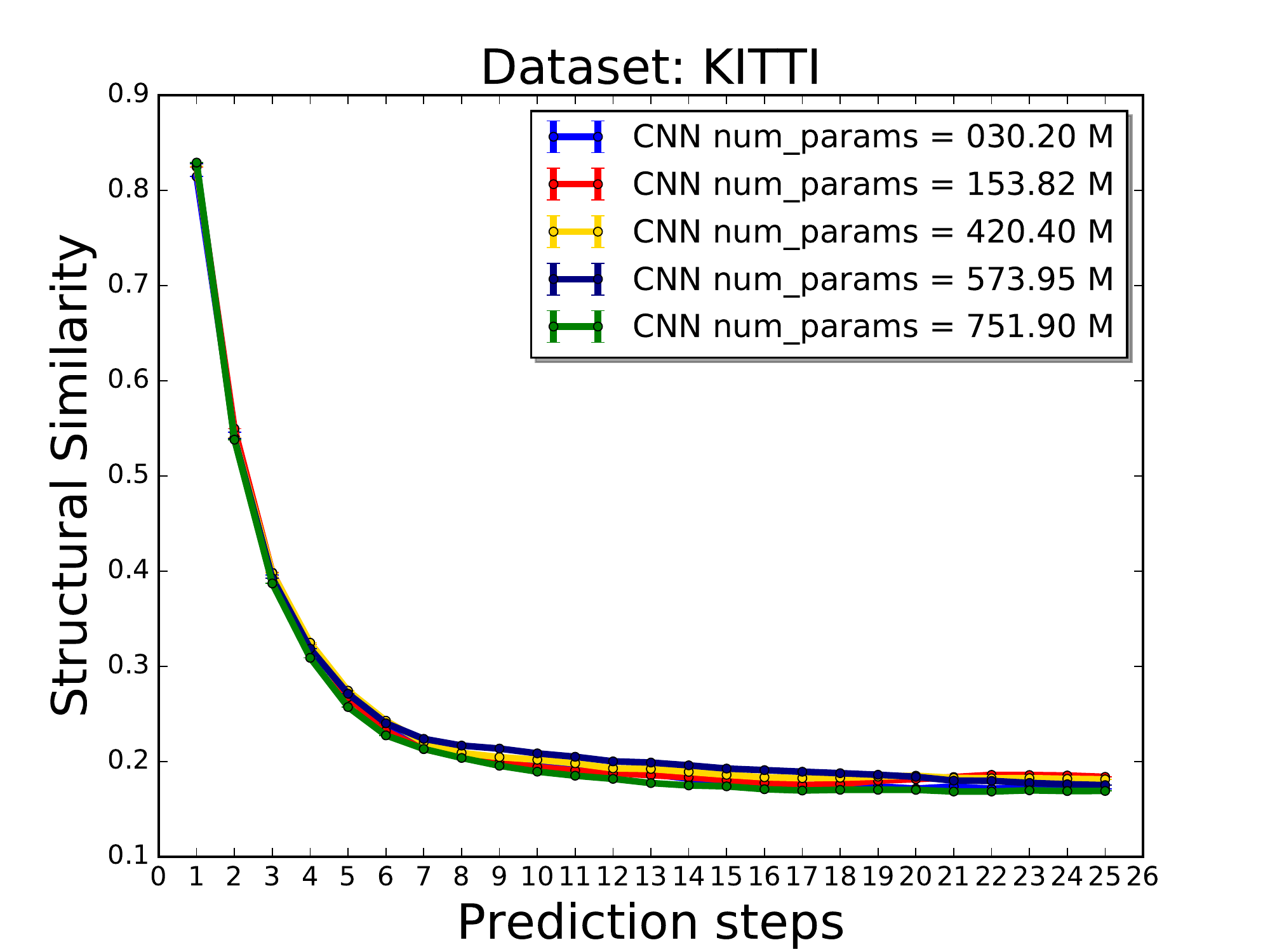}
\label{fig:frame_kitti_cnn}
\end{figure}

\begin{figure}[htp!]
    \centering
    \hspace{-8pt}
	\includegraphics[width=.35\linewidth]{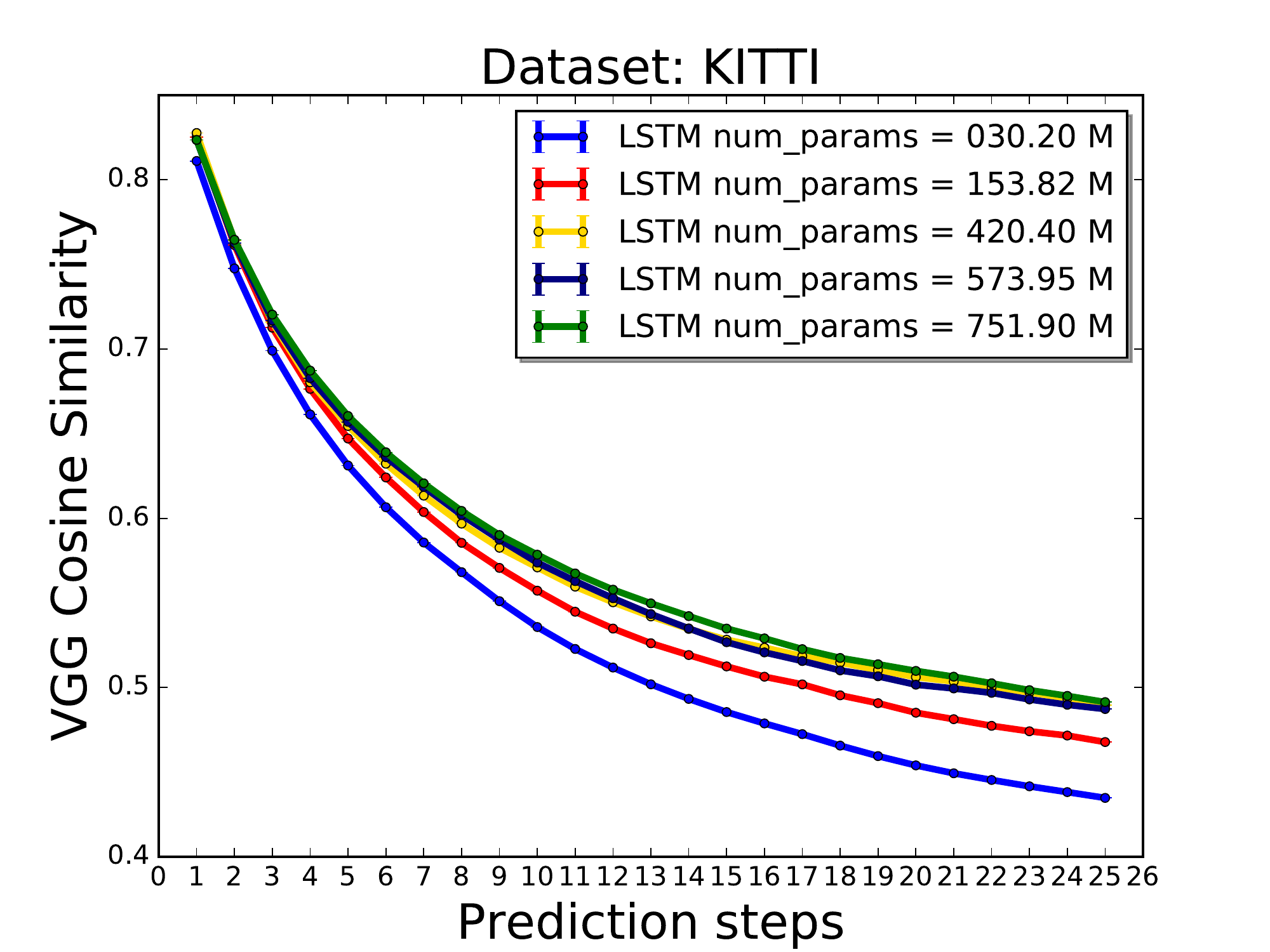} \hspace{-14pt}
	\includegraphics[width=.35\linewidth]{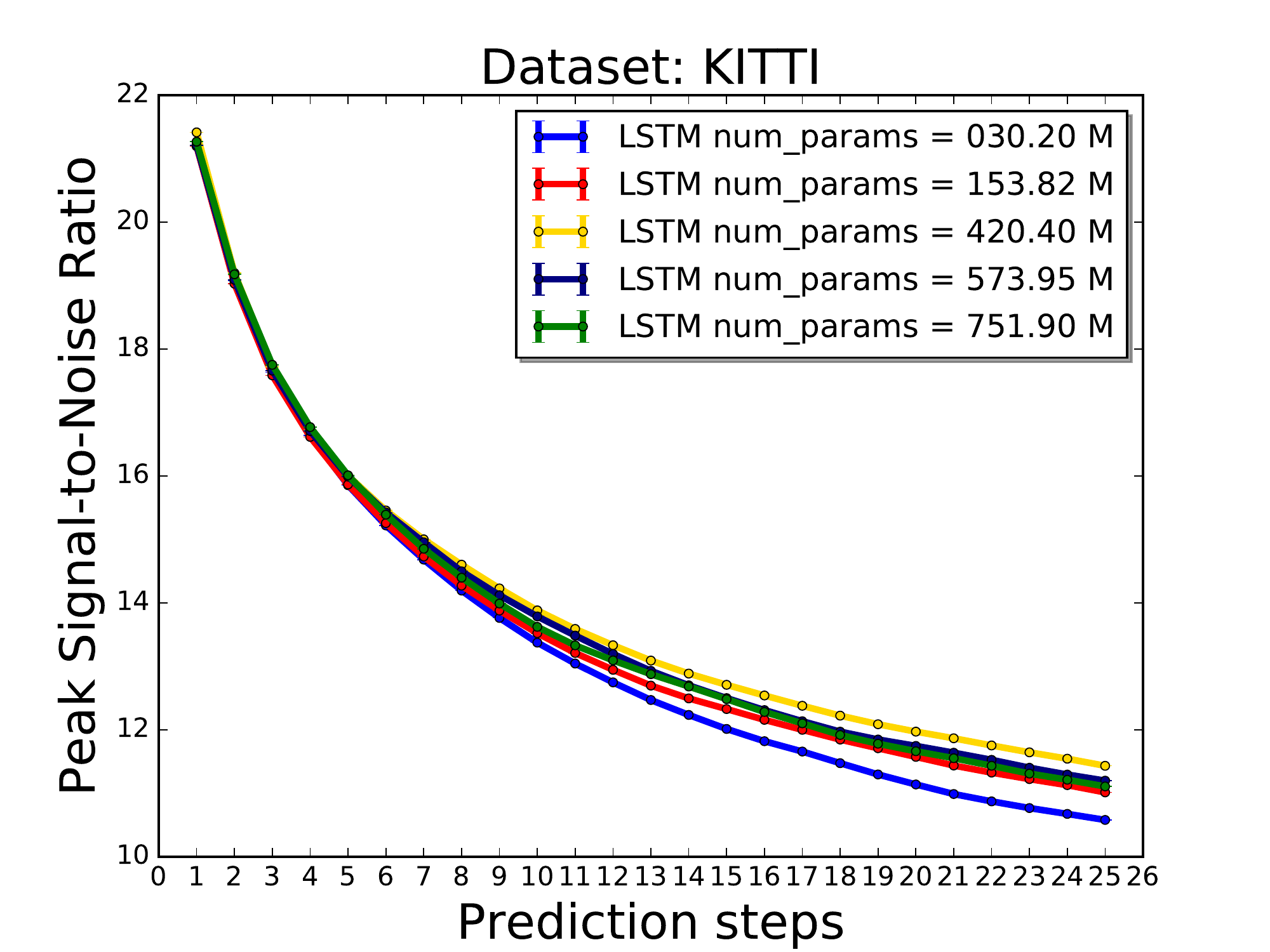} \hspace{-14pt}
	\includegraphics[width=.35\linewidth]{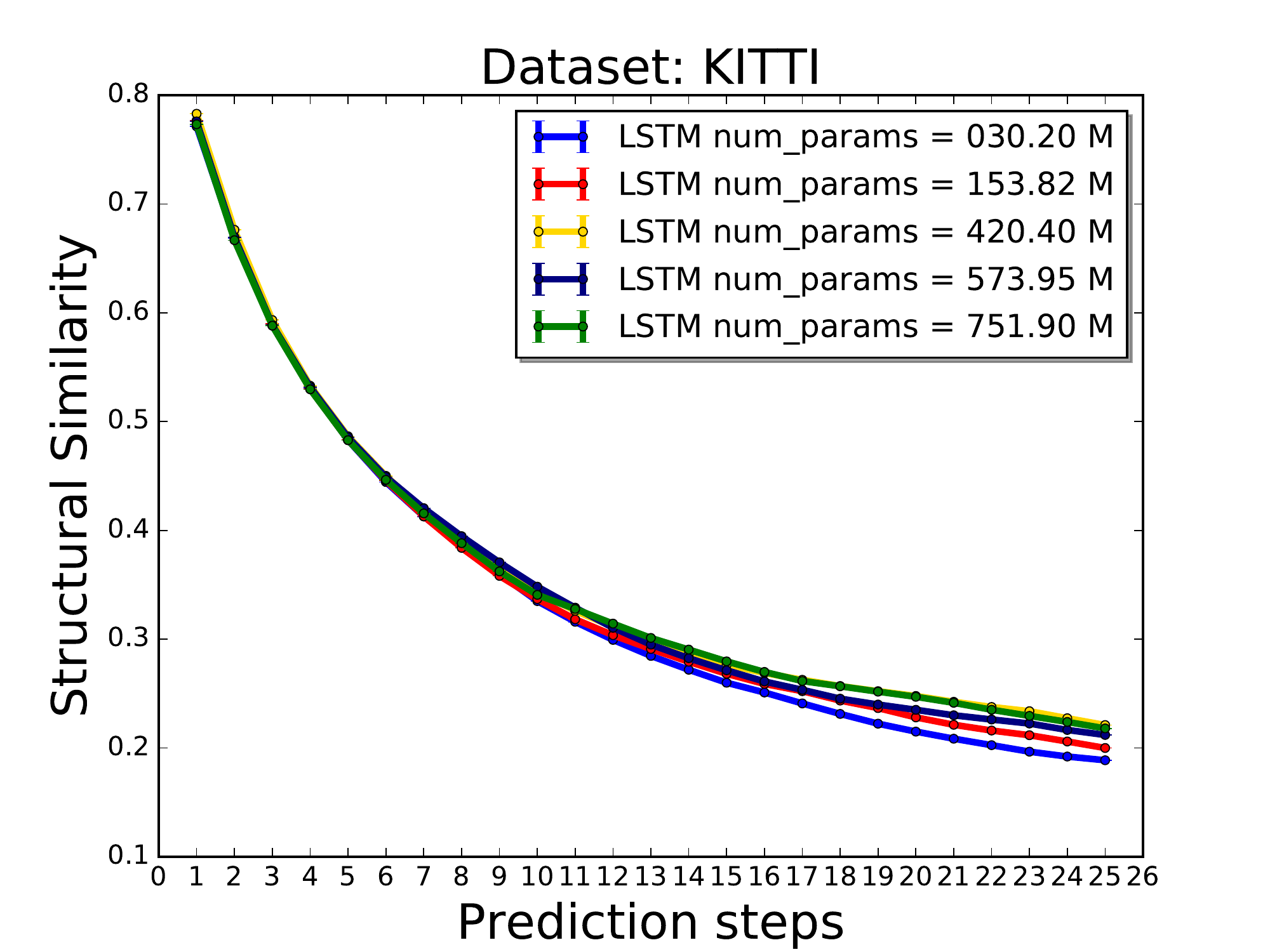}
\label{fig:frame_kitti_lstm}
\end{figure}

\begin{figure}[htp!]
    \centering
    \hspace{-8pt}
	\includegraphics[width=.35\linewidth]{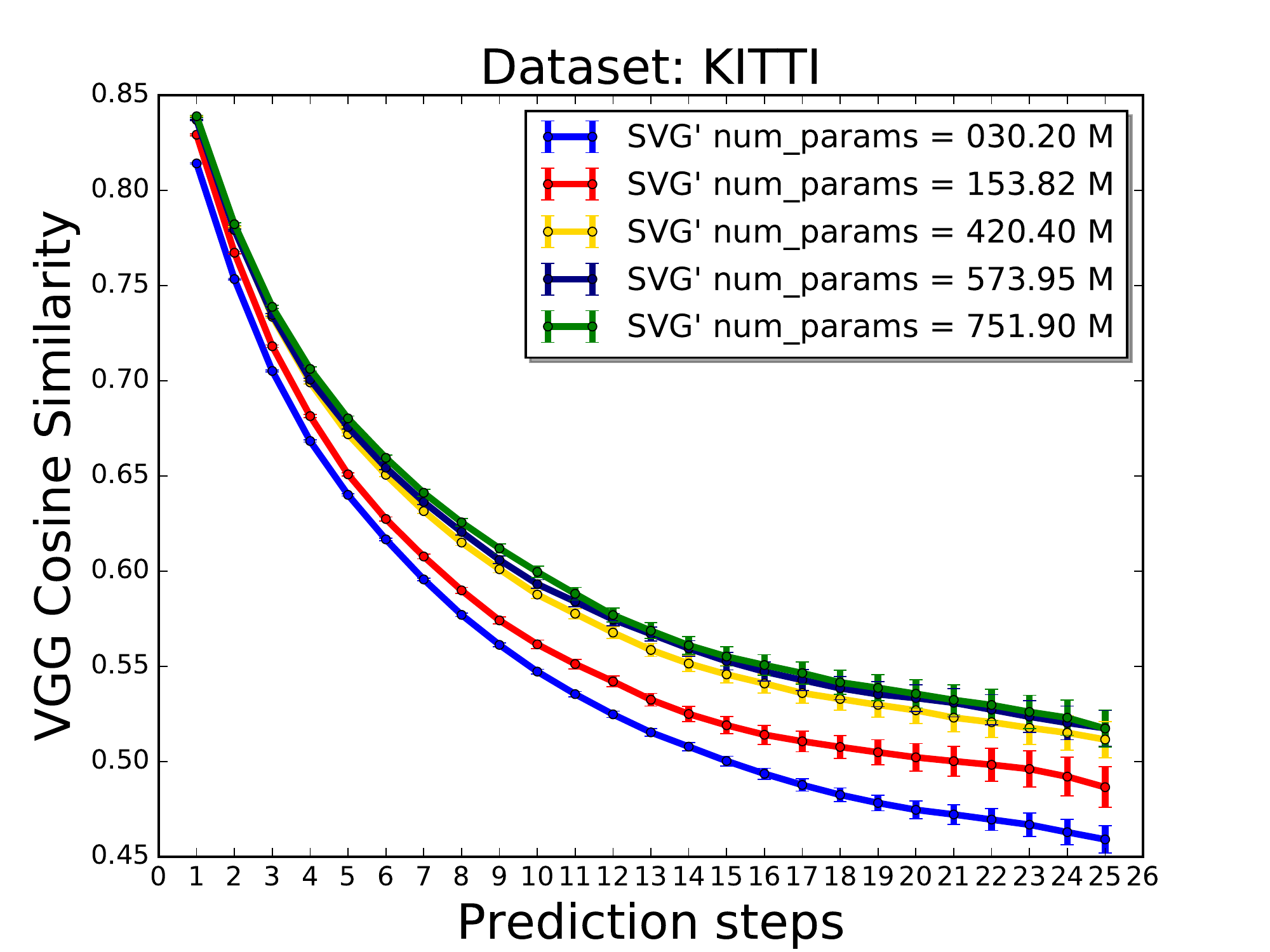} \hspace{-14pt}
	\includegraphics[width=.35\linewidth]{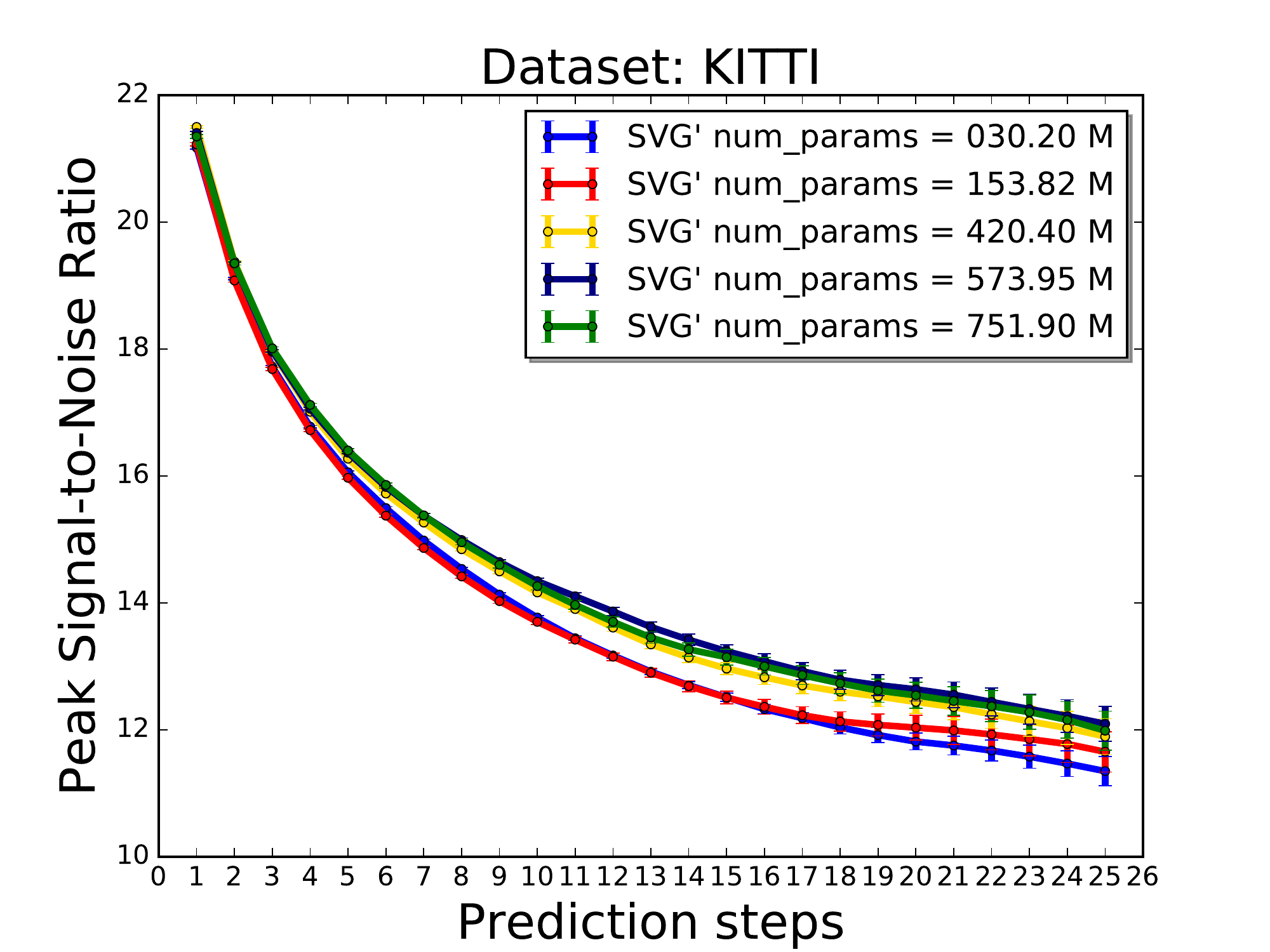} \hspace{-14pt}
	\includegraphics[width=.35\linewidth]{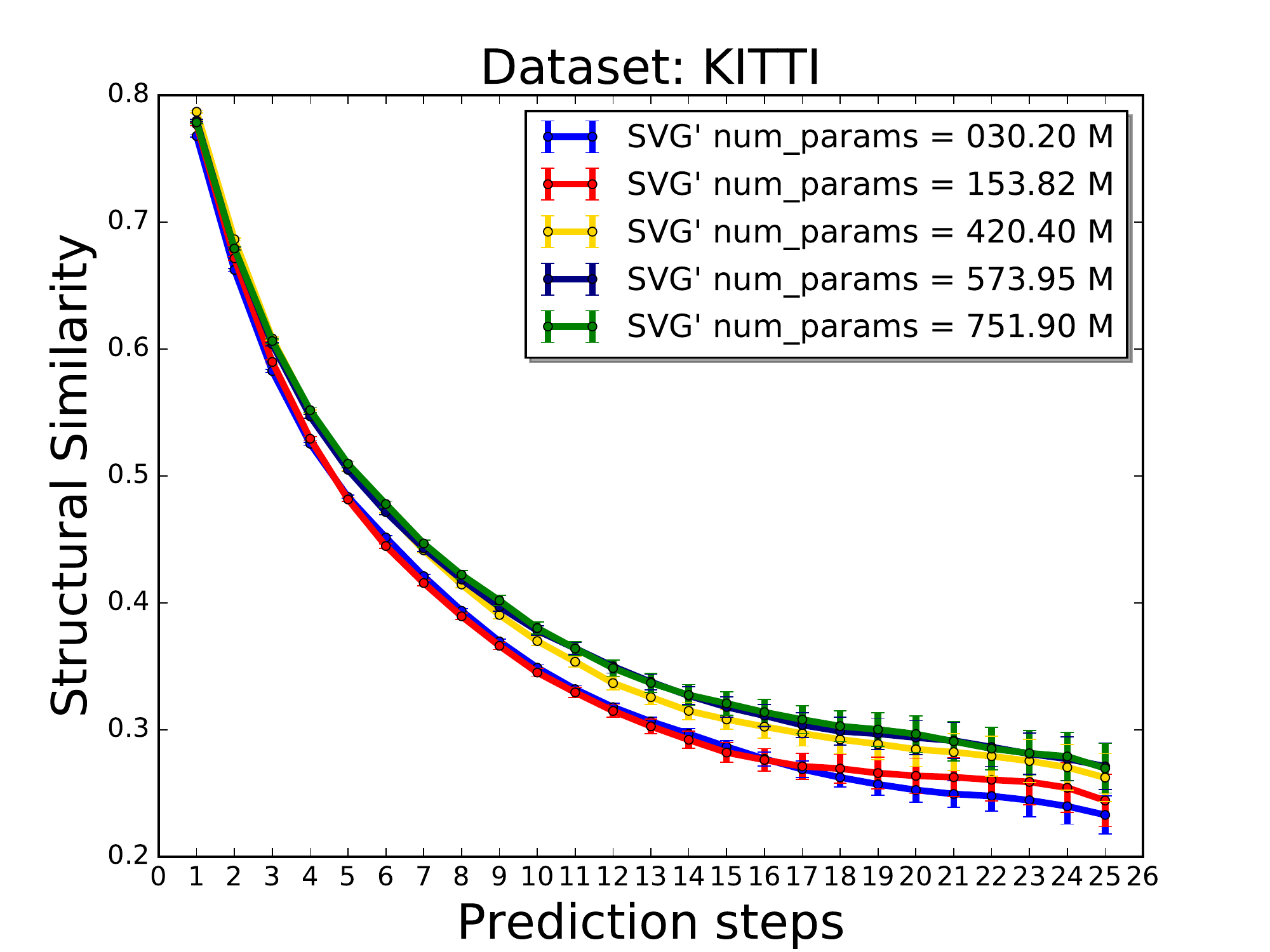}
	\caption{KITTI driving per-frame evaluation (higher is better). As capacity increases, the per frame evaluation metrics become better. The increase is due to better modeling the driving dynamics and partial observability. Due to the difficulty of predicting the exact not-observed parts of the image, the performance converges toward the largest models.}
\label{fig:frame_kitti_svg}
\end{figure}

\clearpage

\subsection{Effects of using skip connections in video prediction} \label{supp:skipconnections}
In this section, we present a study on the effects of using skip connections from encoder to decoder.
Similar to \citet{emily}, the method presented in the main text has skip connections going from the encoder of the last observed frame directly to the decoder for all frame predictions.
This allows the video prediction method to choose to transfer pixels that did not move from the input frame directly into the output frame, and generate the pixels that move.
Below, we show the performance for each of the datasets presented in this work.

\subsubsection{Robot Arm.}
\begin{figure}[htp!]
    \centering
    \hspace{-8pt}
	\includegraphics[width=.4\linewidth]{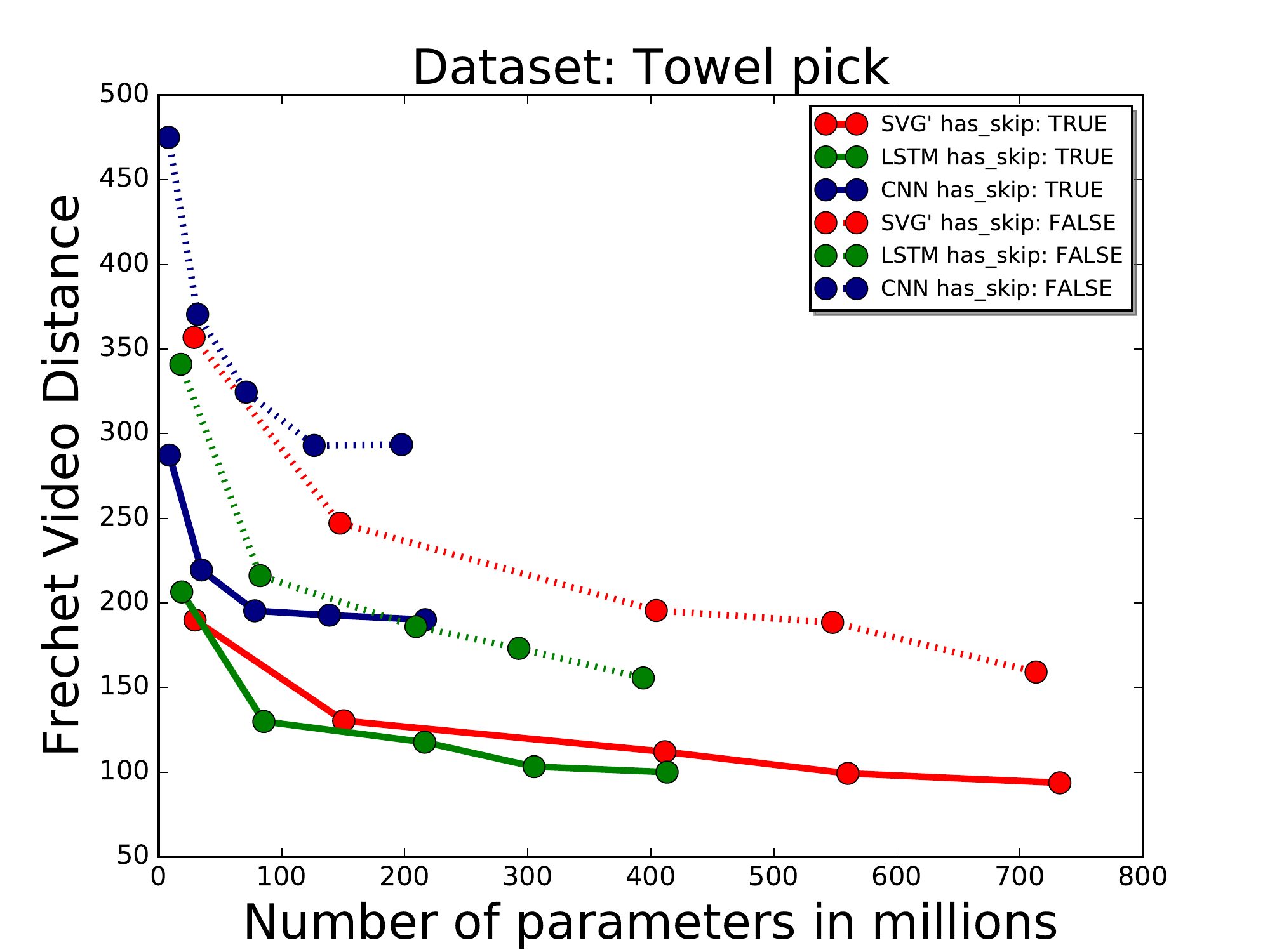}
	\caption{Towel pick video dynamics evaluation (lower is better). Solid lines define method with skip connections and dotted lines without skip connections.}
    \label{fig:dyn_towel_extra}
\end{figure}

In Figure \ref{fig:dyn_towel_extra}, we can see that skip connections do play an important role in terms of FVD evaluation for the robot arm action conditioned experiments. This implies that having skip connections eases the difficulty of video prediction in that it is only required to model the dynamics of the moving parts and everything else can simply be transferred to the output frames.

\begin{figure}[htp!]
    \centering
    \hspace{-8pt}
	\includegraphics[width=.35\linewidth]{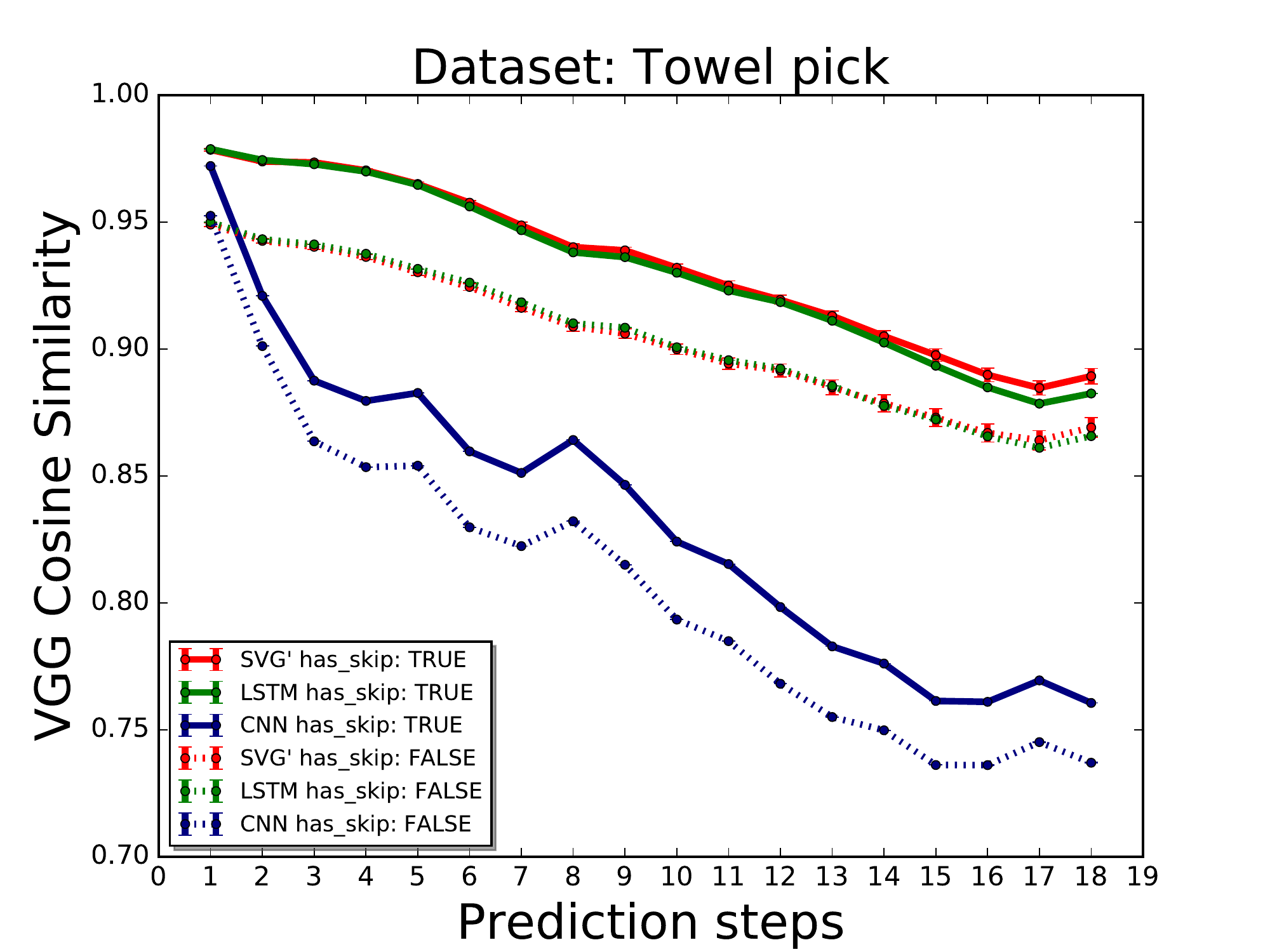} \hspace{-14pt}
	\includegraphics[width=.35\linewidth]{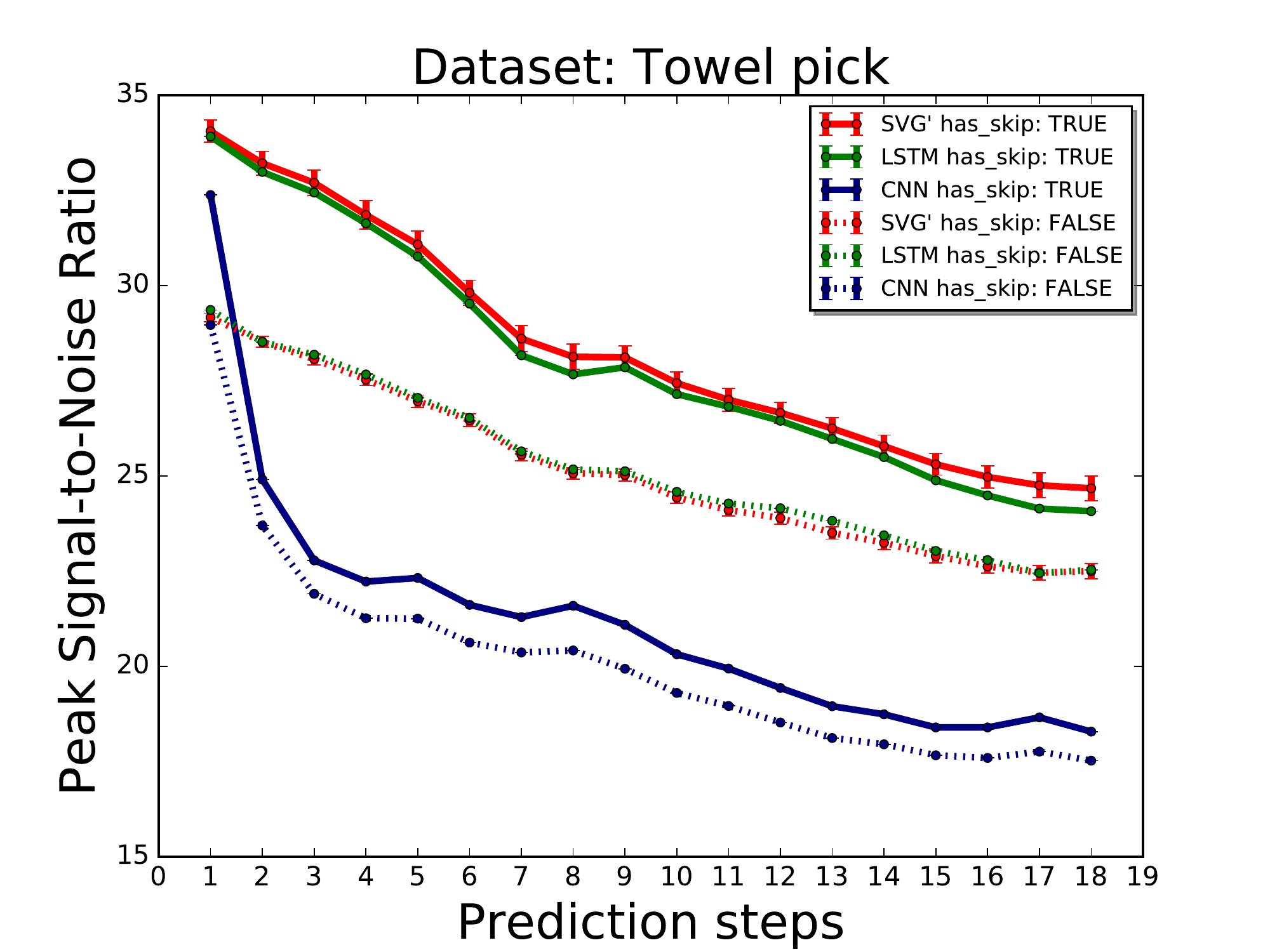} \hspace{-14pt}
	\includegraphics[width=.35\linewidth]{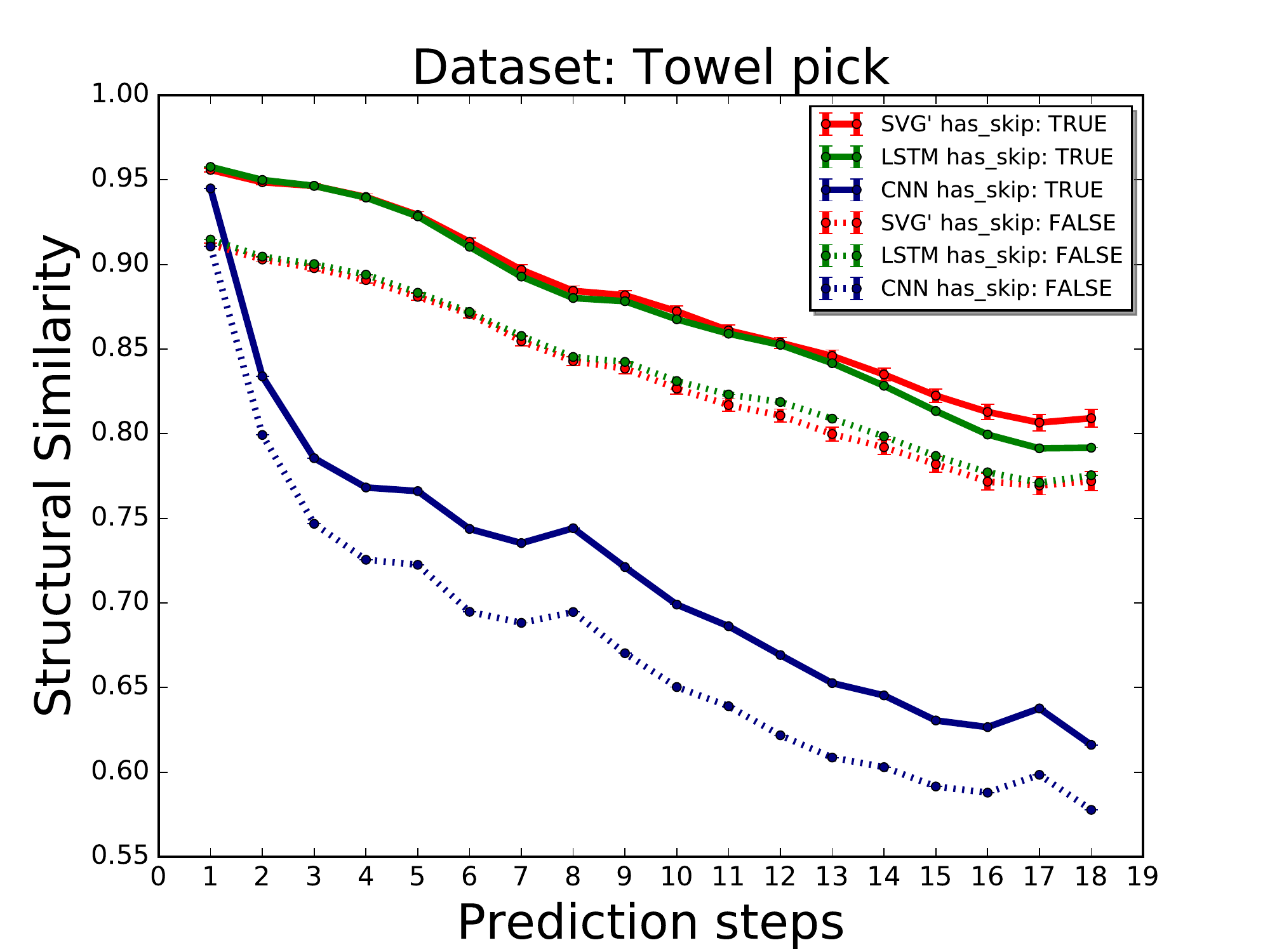}
	\caption{Towel pick per-frame evaluation (higher is better). Solid lines define method with skip connections and dotted lines without skip connections.}
\label{fig:frame_towel_extra}
\end{figure}

In addition, having skip connections also help to make more accurate frame-wise predictions. In Figure \ref{fig:frame_towel_extra}, the advantage of having skip connections is clear in all prediction steps. This indicates that skip connections are not just essential for predicting dynamics that look like the ground-truth videos, but also, the accuracy of the predicted pixels becomes better.

\subsubsection{Human activities.}
\begin{figure}[htp!]
    \centering
    \hspace{-8pt}
	\includegraphics[width=.4\linewidth]{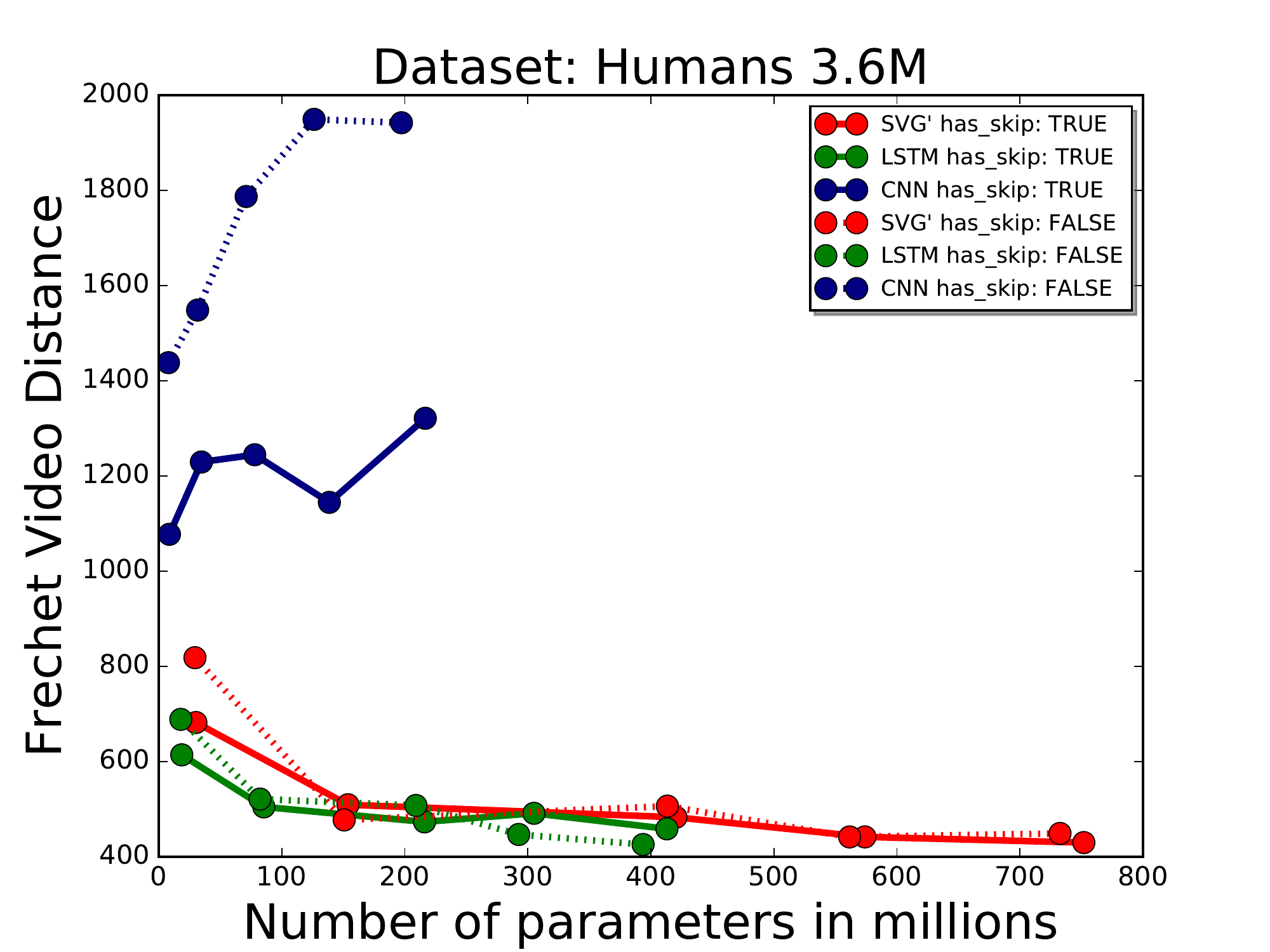}
	\caption{Human 3.6M video dynamics evaluation (lower is better). Solid lines define method with skip connections and dotted lines without skip connections.}
    \label{fig:dyn_humans_extra}
\end{figure}

In Figure \ref{fig:dyn_humans_extra}, having skip connections results in a large performance improvement in FVD for the CNN based video prediction architecture. However, for the LSTM and SVG' based architectures, we can that there is not clear improvement as the model size increases. We hypothesize that, since there are no interactions, the background is static, and the background between training and testing data is similar, the dataset dynamics become easier to model. Therefore, there is no need for the model to separate moving and non-moving parts to achieve good predictions.

\begin{figure}[htp!]
    \centering
    \hspace{-8pt}
	\includegraphics[width=.35\linewidth]{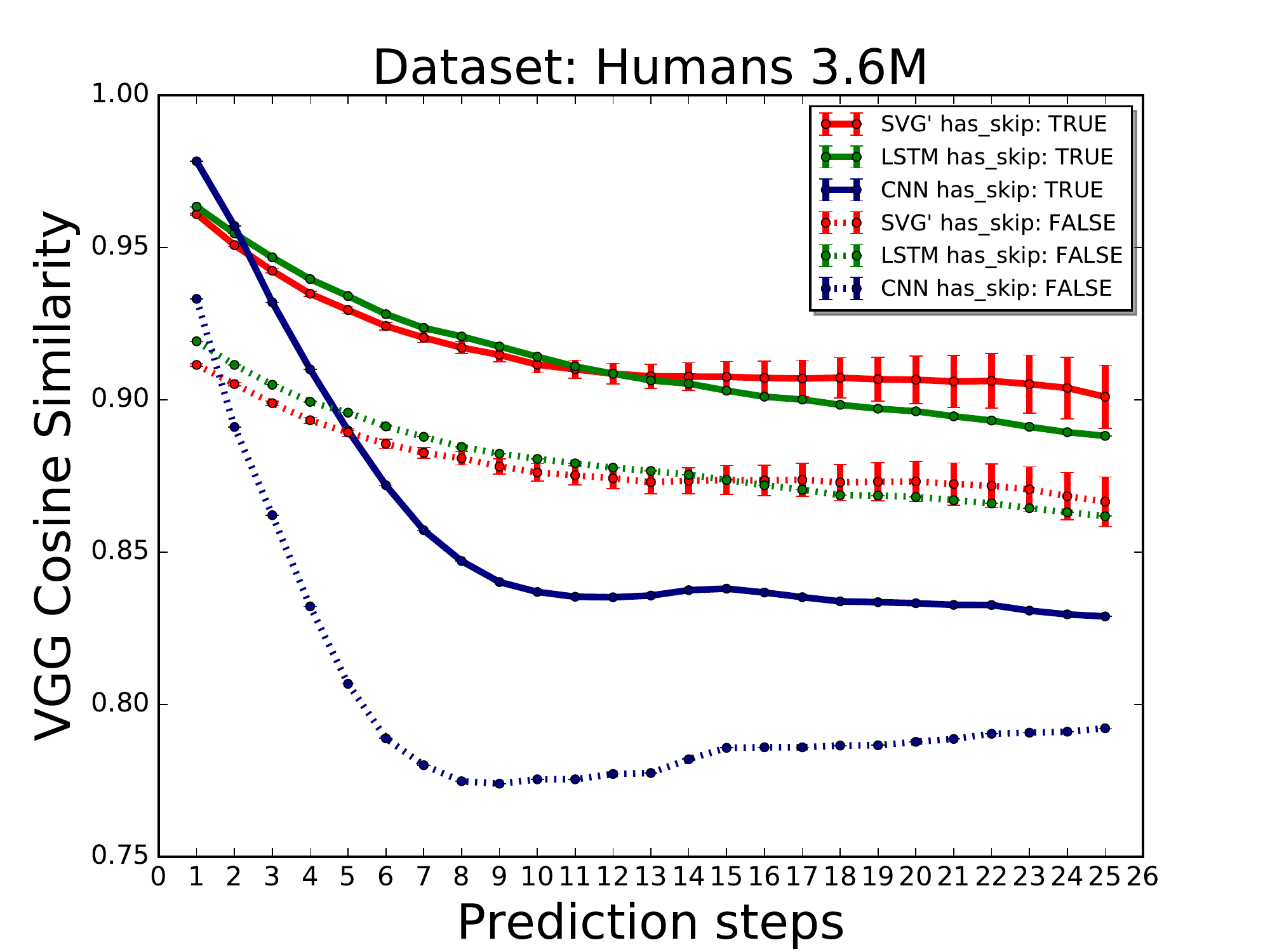} \hspace{-14pt}
	\includegraphics[width=.35\linewidth]{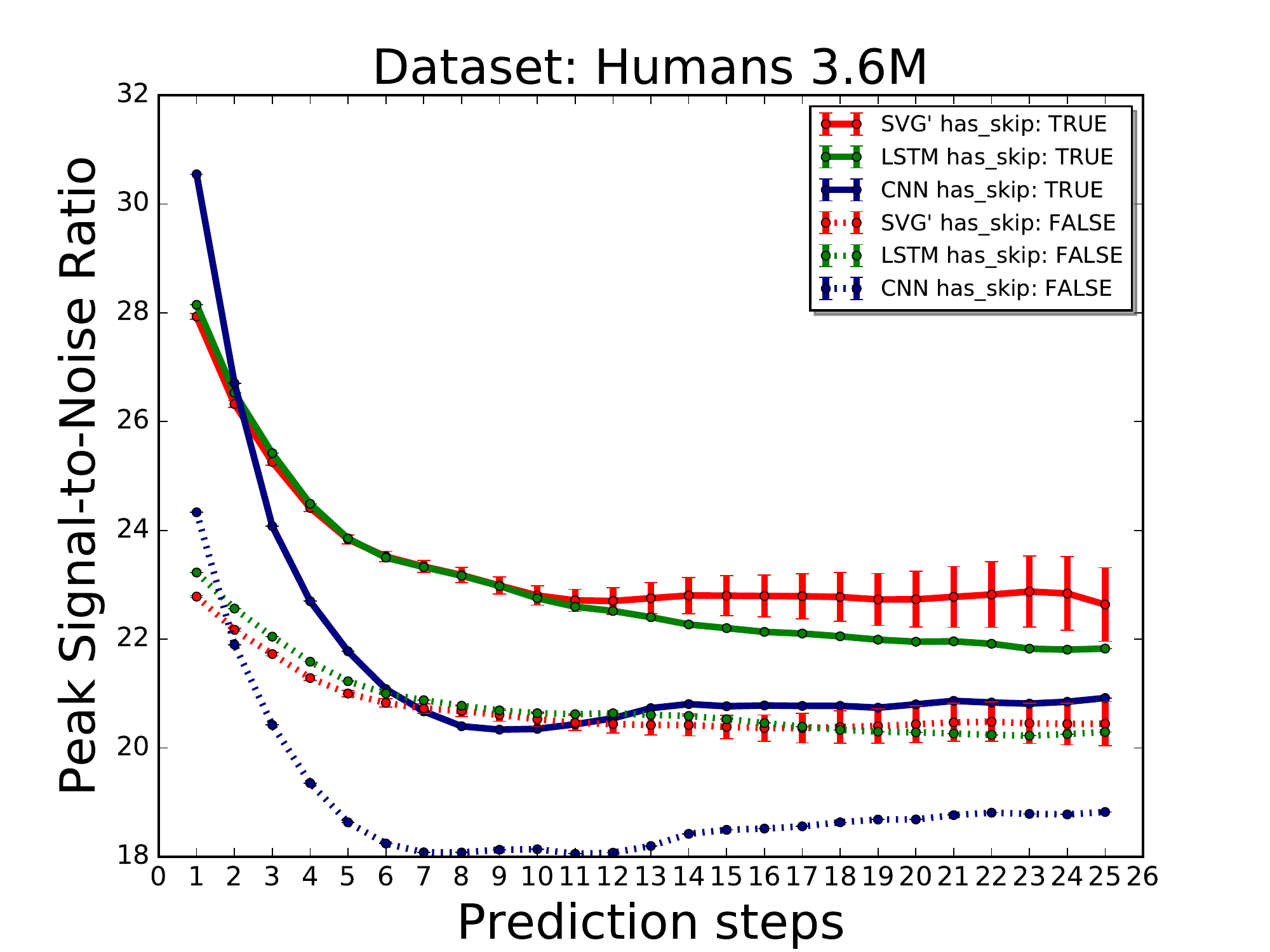} \hspace{-14pt}
	\includegraphics[width=.35\linewidth]{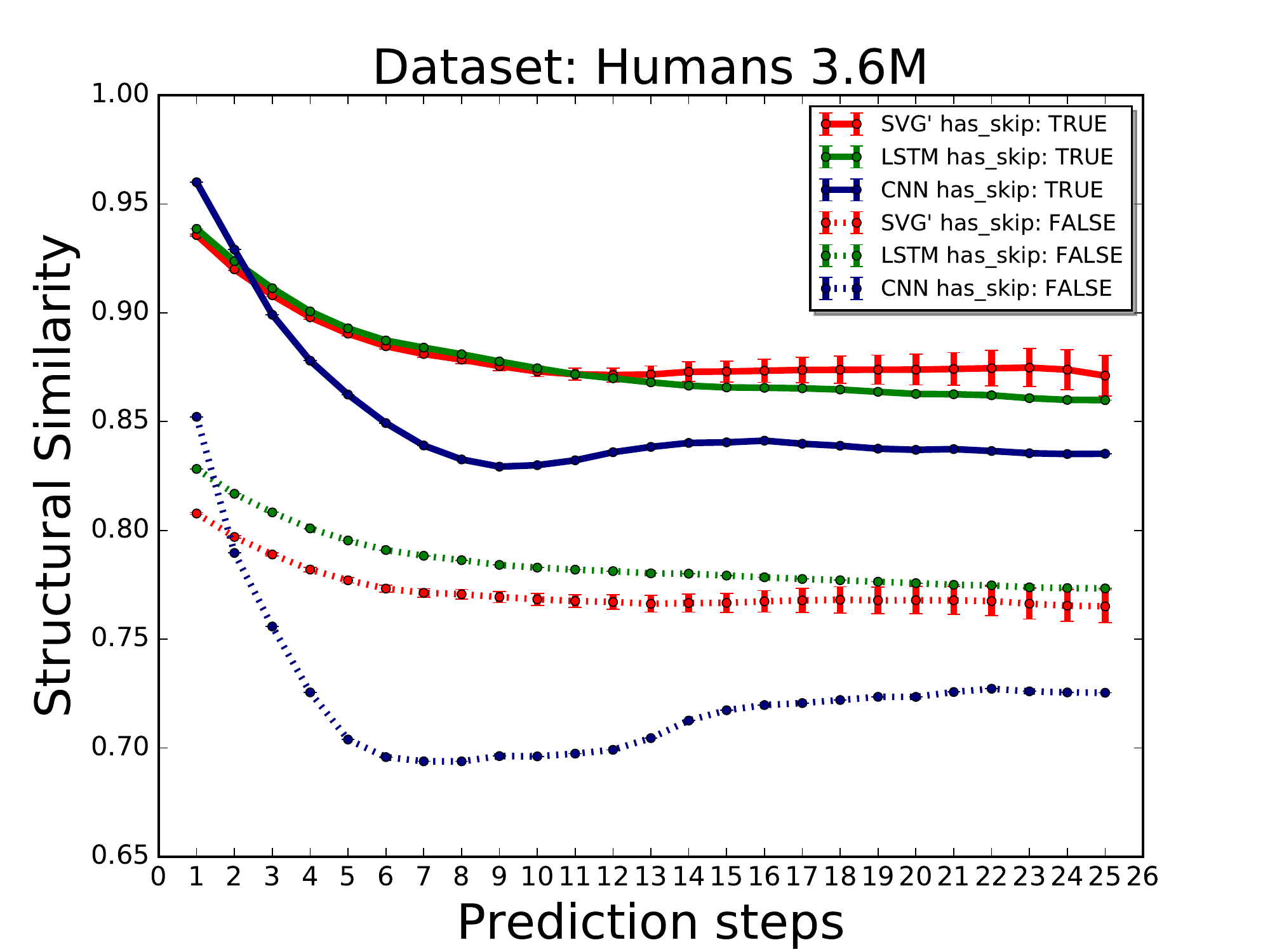}
	\caption{Human 3.6M per-frame evaluation (higher is better). Solid lines define method with skip connections and dotted lines without skip connections.}
\label{fig:frame_humans_extra}
\end{figure}

In contrast to FVD evaluation, having skip connections greatly improves the performance in the per-frame evaluation metrics for all models (Figure \ref{fig:frame_humans_extra}). This is mainly due to the fact that the moving humans take up a very small portion of the image. Thus, having a way to transfer non-moving pixels directly into the output frames results in more accurate per-frame performance.

\subsubsection{KITTI driving.}
\begin{figure}[htp!]
    \centering
    \hspace{-8pt}
	\includegraphics[width=.4\linewidth]{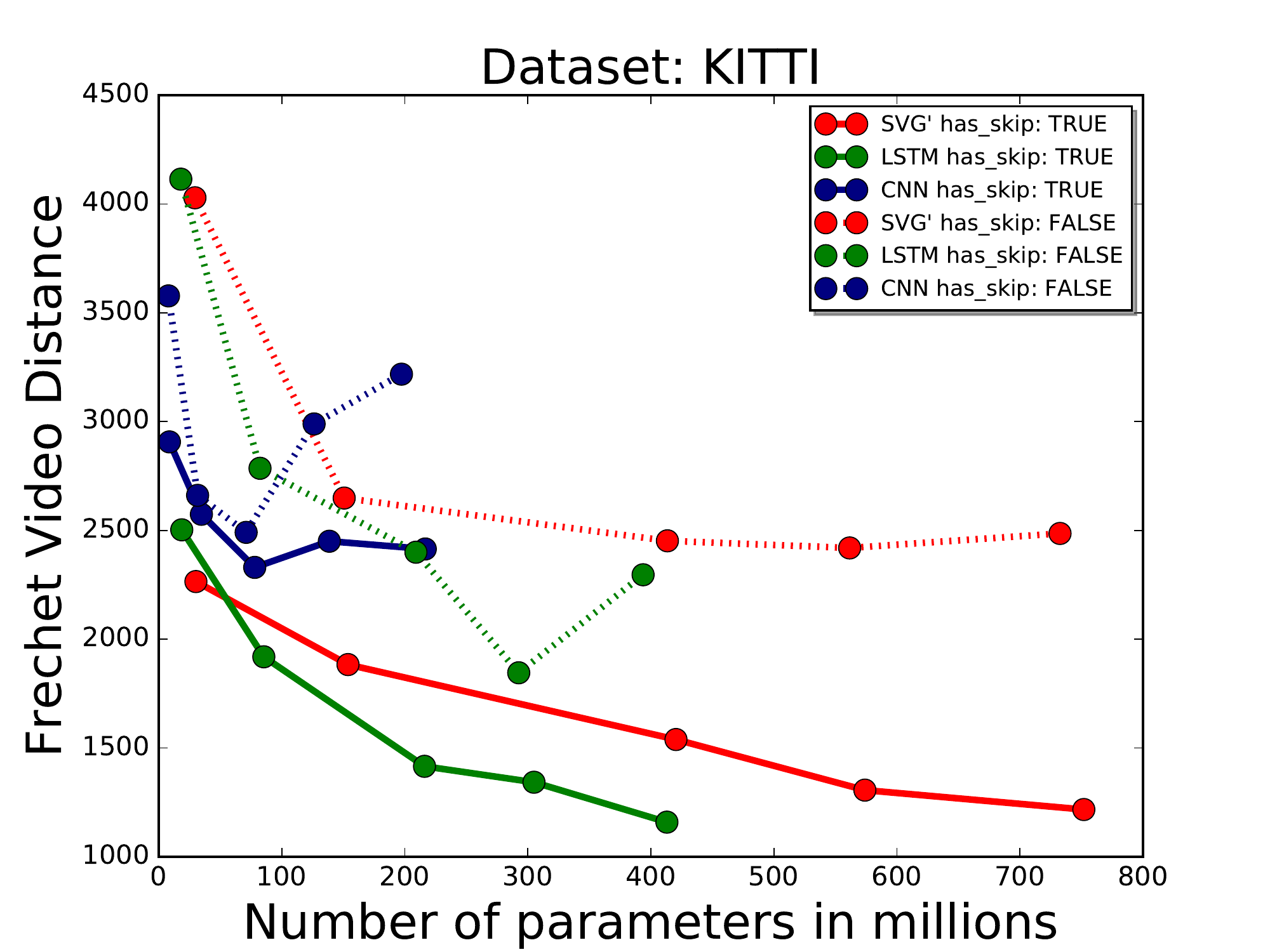}
	\caption{KITTI driving video dynamics evaluation (lower is better). Solid lines define method with skip connections and dotted lines without skip connections.}
    \label{fig:dyn_driving_extra}
\end{figure}

In Figure \ref{fig:dyn_driving_extra}, we can see that for the recurrent models (LSTM and SVG') having skip connections results in improved FVD performance. However, when using a CNN based architecture, is clear for most models, but not all of them as the two curves become close to each other when $M$ and $K$ are make the model twice and three times bigger than the original model (second and third parameter value in the x-axis). We hypothesize that this happens because almost all pixels move in these videos, and so, simple skip connections without recurrent steps to remember what pixels are moving throughout the prediction makes skip connections not as critical for the intermediate size models.

\begin{figure}[htp!]
    \centering
    \hspace{-8pt}
	\includegraphics[width=.35\linewidth]{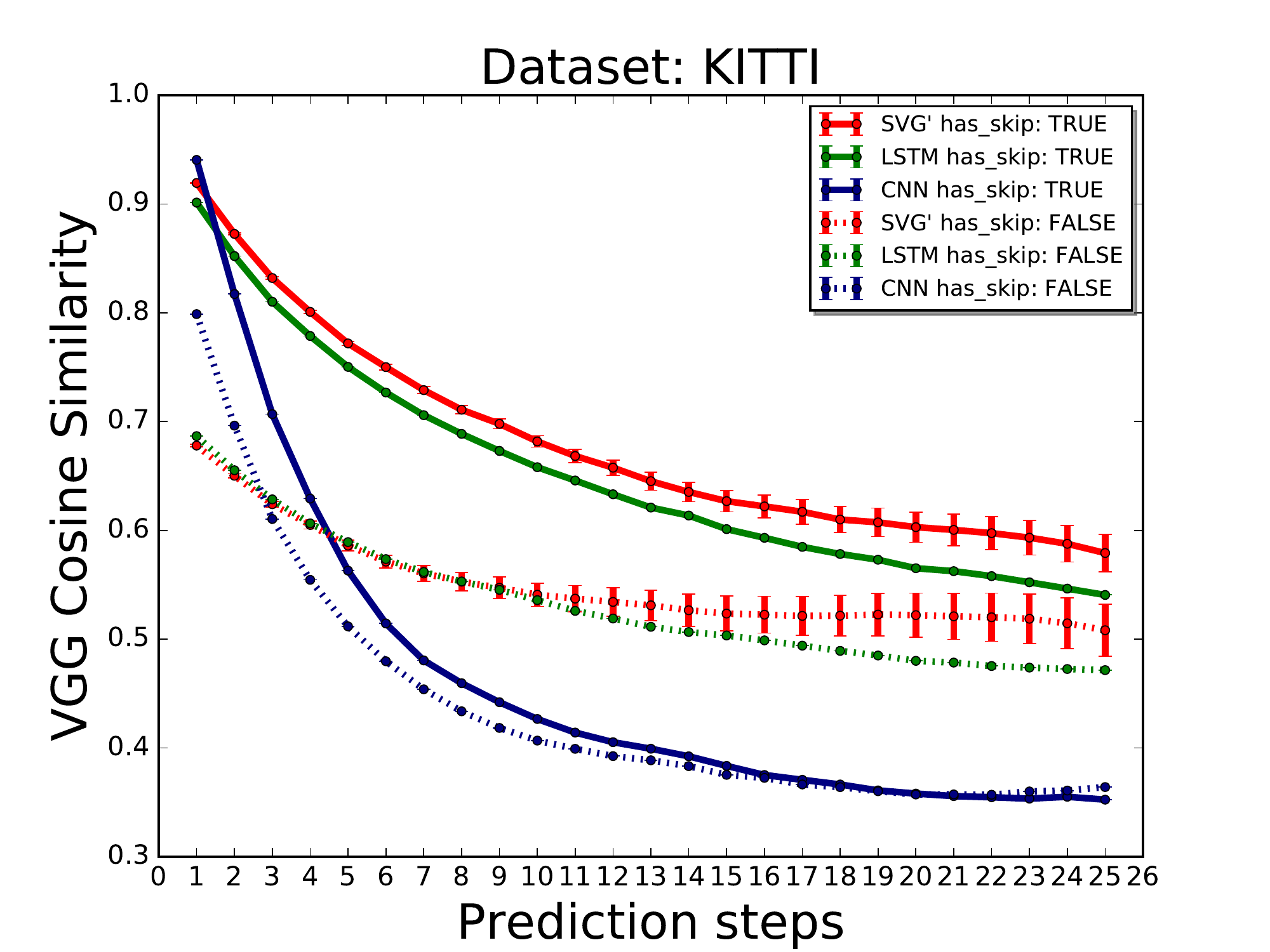} \hspace{-14pt}
	\includegraphics[width=.35\linewidth]{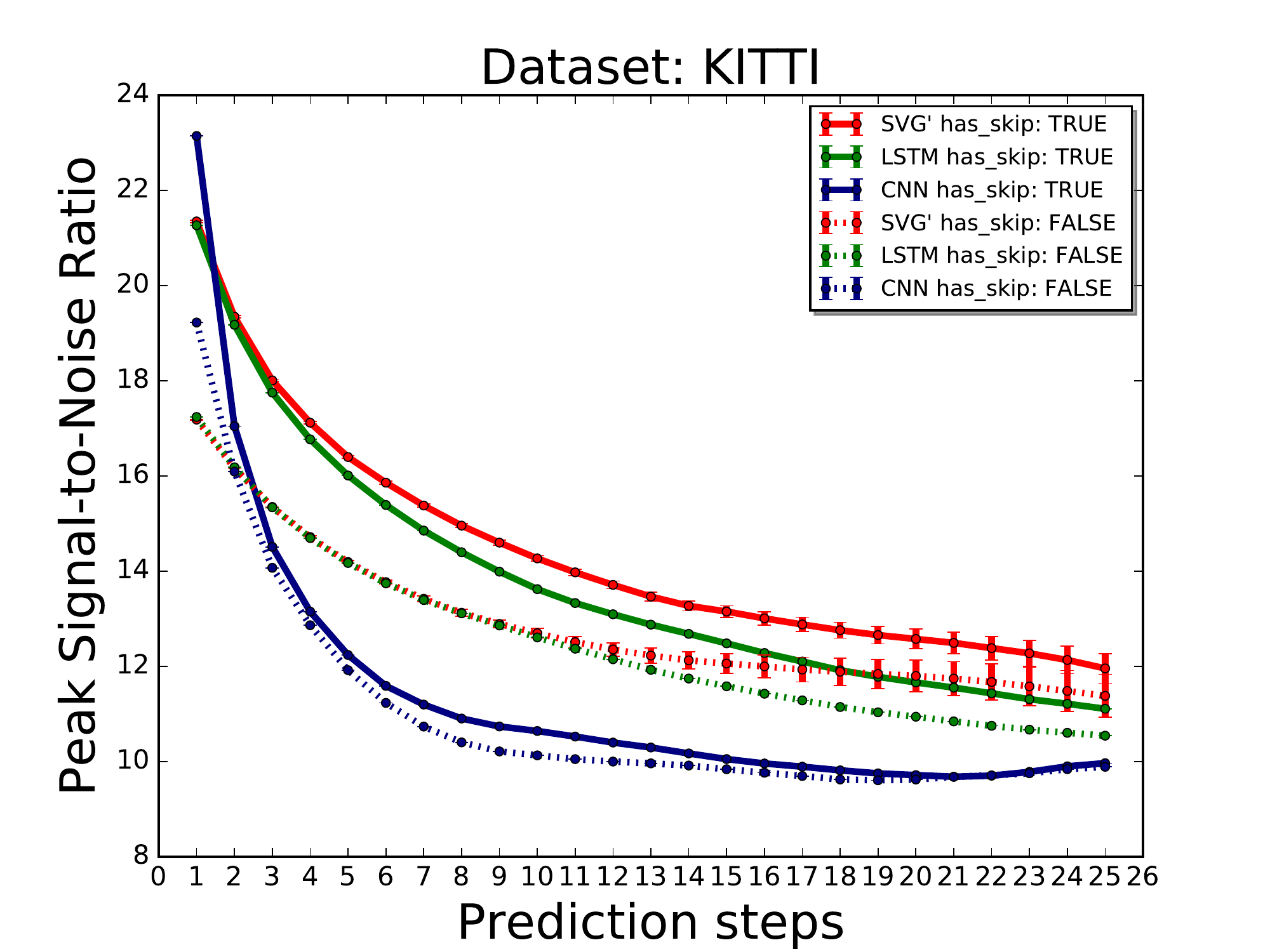} \hspace{-14pt}
	\includegraphics[width=.35\linewidth]{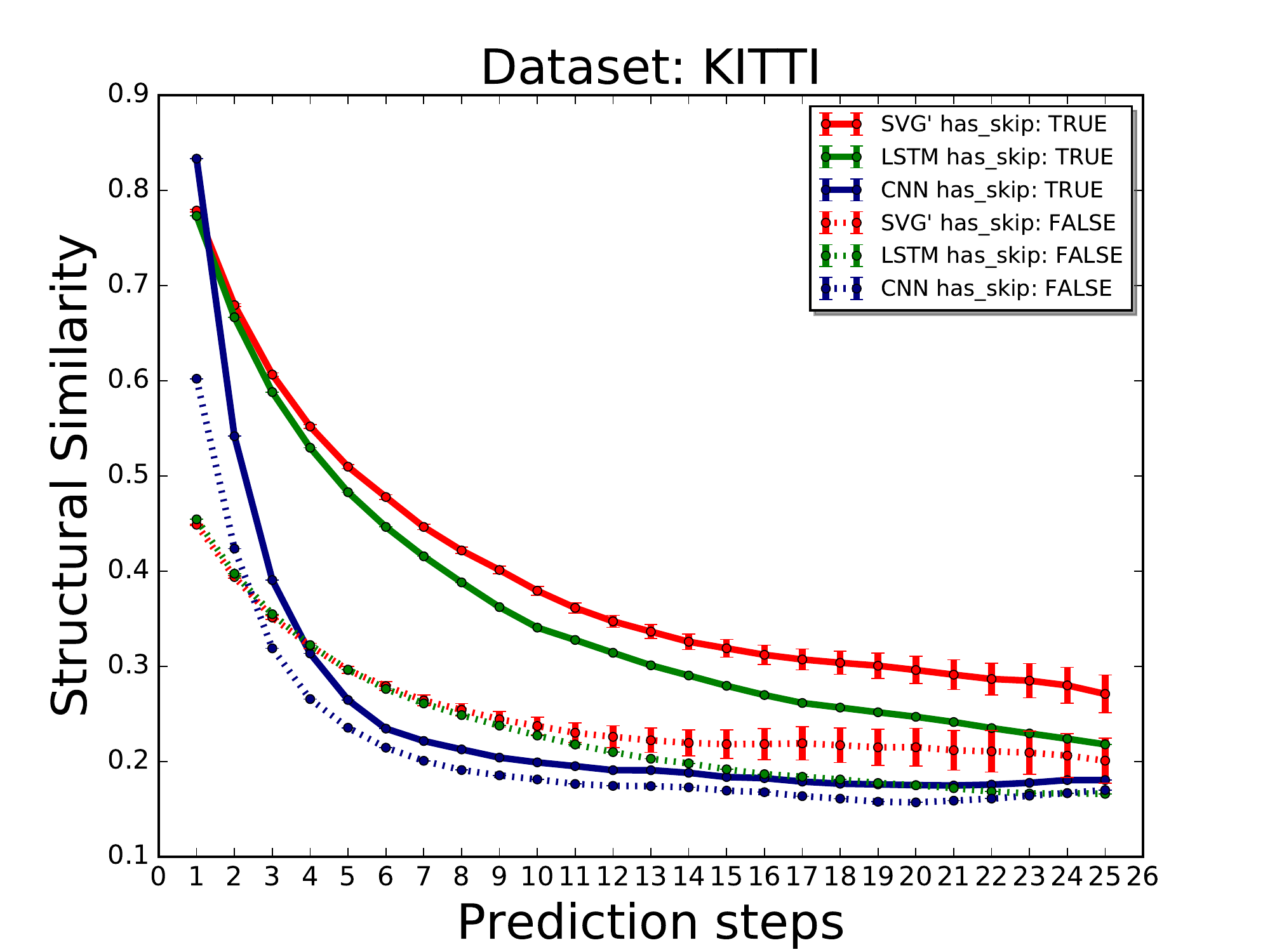}
	\caption{KITTI driving per-frame evaluation (higher is better). Solid lines define method with skip connections and dotted lines without skip connections.}
\label{fig:frame_driving_extra}
\end{figure}

In terms of per-frame evaluation, we see an interesting behavior as prediction move forward in time (Figure \ref{fig:frame_driving_extra}). The predicted frames become less accurate as time moves forward; effectively reducing the performance gap between the architectures with and without skip connections. This happens because predicting videos in this dataset requires predicting unseen pixels moving into view (e.g., partial observability). Therefore, having skip connections can only help for predicting nearby frames and eventually requires generating fully unseen objects in the frames. The probability that the exact pixels are generated reduces as time moves forward, even if the overall predicted dynamics are within what is realistic in the dataset.

\newpage
\subsection{Effects of the number of context frames} \label{supp:contextframes}
In this section, we present a study over number of context frames given to each of the considered networks. We consider models that observe 2, 5 and 10 frames to predict 20 frames into the future for our action-free experiments (Human 3.6M and KITTI), and models observe 2, 4 and 8 frames to predict 12 frames into the future for our action-conditioned experiments (Towel pick). We test on a slightly different test set from the one in the main paper to make sure the future frames during evaluation are all the same for all the models in this section. We present the per-frame metrics used in the main paper but averaged over time, and also, the Fr\'echet Video Distance (FVD) dynamics evaluation metric.

\subsubsection{Per-frame evaluation}
Firstly, we perform per-frame evaluation of the predicted frames. We want to observe how context affects the accuracy of the predicted future with respect to the ground-truth future.

In Table \ref{table:context_actionfree_perframe} (action-free evaluation), we can see that increasing the number of context frames improves the performance in most of the recurrent models or converges at context of 5 frames. In contrast, we cannot conclude the same for the CNN models. In fact, most of our experiments perform better with less number of context frames. We hypothesize that this may be due to the lack of recurrence in the CNN model which has to infer dynamics from all context frames in one shot at every prediction step while not keeping a history. The recurrent models have the advantage of keeping a history while deciding what information to keep or discard.

\begin{table}[h!] 
\centering
\setlength{\tabcolsep}{3pt}
\vspace{10pt}
\begin{tabular}{l||c||c||ccc}
\Xhline{4\arrayrulewidth}
\Xhline{4\arrayrulewidth}
Dataset & \makecell{Metric} & \makecell{Network} & \makecell{Context = 2} & \makecell{Context = 5} & \makecell{Context = 10} \\
\Xhline{4\arrayrulewidth}
\Xhline{4\arrayrulewidth}
Human 3.6M & \makecell{\quad \\ Cosine Sim. \\ \quad \\ \quad \\ \quad \\ PSNR \\ \quad \\ \quad \\ \quad \\ SSIM \\ \quad } & \makecell{SVG' \\ LSTM \\ CNN \\ \quad \\ SVG' \\ LSTM \\ CNN \\ \quad \\ SVG' \\ LSTM \\ CNN} & \makecell{0.916 \\ 0.912 \\ \textbf{0.890} \\ \quad \\ 23.420 \\ 22.956 \\ \textbf{22.804} \\ \quad \\ 0.880 \\ 0.876 \\ 0.862} & \makecell{\textbf{0.925} \\ 0.918 \\ 0.869 \\ \quad \\ 23.778 \\ 23.391 \\ 21.740 \\ \quad \\ 0.889 \\ 0.883 \\ 0.857} & \makecell{\textbf{0.925} \\ \textbf{0.921} \\ 0.875 \\ \quad \\ \textbf{23.948} \\ \textbf{23.734} \\ 21.845 \\ \quad \\ \textbf{0.892}\\ \textbf{0.886} \\ \textbf{0.863}}\\
\Xhline{4\arrayrulewidth}
KITTI & \makecell{\quad \\ Cosine Sim. \\ \quad \\ \quad \\ \quad \\ PSNR \\ \quad \\ \quad \\ \quad \\ SSIM \\ \quad } & \makecell{SVG' \\ LSTM \\ CNN \\ \quad \\ SVG' \\ LSTM \\ CNN \\ \quad \\ SVG' \\ LSTM \\ CNN} & \makecell{0.689 \\ 0.639 \\ \textbf{0.594} \\ \quad \\ 14.549 \\ 13.623 \\ \textbf{13.522} \\ \quad \\ 0.403 \\ 0.349 \\ \textbf{0.316}} & \makecell{\textbf{0.702} \\ 0.672 \\ 0.493 \\ \quad \\ 14.953 \\ 14.476 \\ 11.883 \\ \quad \\ \textbf{0.419} \\ 0.387 \\ 0.264} & \makecell{0.700 \\ \textbf{0.688} \\ 0.508 \\ \quad \\ \textbf{14.960} \\ \textbf{14.694} \\ 11.989 \\ \quad \\ 0.417 \\ \textbf{0.403} \\ 0.275}\\
\Xhline{4\arrayrulewidth}
\Xhline{4\arrayrulewidth}
\end{tabular}
\vspace{12pt}
\caption{Average per-frame evaluation of the effects of the number of context frames in the action-free datasets (Human 3.6M and KITTI). We compare models with different number of context frames and prediction of 20 frames.}
\vspace{-.1in}
\label{table:context_actionfree_perframe}
\end{table}

In Table \ref{table:context_actioncondition_perframe} (action-conditioned evaluation), we see a similar pattern as in Table \ref{table:context_actionfree_perframe} for the recurrent models. Having more context frames enables recurrent models to make more accurate predictions of the future with respect to the ground-truth future. In addition, the CNN based architecture performance does not degrade as more context frames are given as input. Having actions as input makes the prediction easier, and the CNN does not have to infer all future frame dynamics from pixels alone.

\begin{table}[h!] 
\centering
\setlength{\tabcolsep}{3pt}
\vspace{10pt}
\begin{tabular}{l||c||c||ccc}
\Xhline{4\arrayrulewidth}
\Xhline{4\arrayrulewidth}
Dataset & \makecell{Metric} & \makecell{Network} & \makecell{Context = 2}& \makecell{Context = 4} & \makecell{Context = 8} \\
\Xhline{4\arrayrulewidth}
\Xhline{4\arrayrulewidth}
Towel pick & \makecell{\quad \\ Cosine Sim. \\ \quad \\ \quad \\ \quad \\ PSNR \\ \quad \\ \quad \\ \quad \\ SSIM \\ \quad } & \makecell{SVG' \\ LSTM \\ CNN \\ \quad \\ SVG' \\ LSTM \\ CNN \\ \quad \\ SVG' \\ LSTM \\ CNN} & \makecell{0.906 \\ 0.904 \\ 0.835 \\ \quad \\ 26.125 \\ 25.328 \\ 21.425 \\ \quad \\ 0.834 \\ 0.827 \\ 0.725} & \makecell{0.926 \\ 0.922 \\ 0.819 \\ \quad \\ 27.814 \\ 27.304 \\ 20.913 \\ \quad \\ 0.868 \\ 0.862 \\ 0.708} & \makecell{\textbf{0.932} \\ \textbf{0.931} \\ \textbf{0.837} \\ \quad \\ \textbf{28.703} \\ \textbf{28.706} \\ \textbf{21.767} \\ \quad \\ \textbf{0.875}\\ \textbf{0.878} \\ \textbf{0.729}}\\
\Xhline{4\arrayrulewidth}
\Xhline{4\arrayrulewidth}
\end{tabular}
\vspace{12pt}
\caption{Average per-frame evaluation of the effects of the number of context frames in the action-conditioned datasets (Towel pick). We compare models with different number of context frames and prediction of 12 frames.}
\vspace{-.1in}
\label{table:context_actioncondition_perframe}
\end{table}

\subsubsection{Fr\'echet Video Distance evaluation}
In this section, we evaluate the dynamics of the generated videos using the Fr\'echet Video Distance (FVD). In Table \ref{table:context_actionfree_fvd}, we see a similar pattern in the Human 3.6M and KITTI driving experiments. For the SVG' architecture, 5 context frames are the most optimal number of frames in terms to predict the best full video dynamics. In the LSTM architecture, 10 context frames are the most optimal. Finally, for the CNN architecture, 2 context frames are the most optimal. From these results, we see that for both datasets the SVG' model the improvement stops at 5 context frames. This could be due to the more conditioning frames impacting the predictions in terms of the distribution of future dynamics. However, we need to investigate further to determine why this is happening. For the LSTM model, more context frames keep improving the predicted dynamics quality. Finally, for the CNN architecture, we see a similar behavior as in the per-frame evaluations where less context frames are better for inferring future dynamics.

\begin{table}[h!] 
\centering
\setlength{\tabcolsep}{3pt}
\vspace{10pt}
\begin{tabular}{l||c||c||ccc}
\Xhline{4\arrayrulewidth}
\Xhline{4\arrayrulewidth}
Dataset & \makecell{Metric} & \makecell{Network} & \makecell{Context = 2} & \makecell{Context = 5} & \makecell{Context = 10} \\
\Xhline{4\arrayrulewidth}
\Xhline{4\arrayrulewidth}
Human 3.6M & \makecell{\quad \\ FVD \\ \quad} & \makecell{SVG' \\ LSTM \\ CNN} & \makecell{440.511 \\ 484.011 \\ \textbf{470.751}} & \makecell{\textbf{428.792} \\ 490.375 \\ 1006.216} & \makecell{434.743 \\ \textbf{463.984} \\ 908.939} \\
\Xhline{4\arrayrulewidth}
KITTI & \makecell{\quad \\ FVD \\ \quad } & \makecell{SVG' \\ LSTM \\ CNN} & \makecell{1183.945 \\ 1309.101 \\ \textbf{1408.143}} & \makecell{\textbf{1125.285} \\ 1228.919 \\ 2673.012} & \makecell{1391.642 \\ \textbf{1224.859} \\ 2494.317} \\
\Xhline{4\arrayrulewidth}
\Xhline{4\arrayrulewidth}
\end{tabular}
\vspace{12pt}
\caption{Fr\'echet Video Distance (FVD) evaluation of the effects of the number of context frames in the action-free datasets (Human 3.6M and KITTI). We compare models with different number of context frames and prediction of 20 frames.}
\vspace{-.1in}
\label{table:context_actionfree_fvd}
\end{table}

In Table \ref{table:context_actionconditioned_fvd}, we see a slightly different result in comparison to Table \ref{table:context_actionfree_fvd}. For both SVG' and LSTM architectures, 8 context frames (the most we tried) are the most optimal number of frames in terms to predict the best video dynamics. The difference in these experiments is that we have action inputs that determine the robot arm motion (albeit the objects with which the arm interacts still have a stochastic behavior). For the CNN architecture, 2 context frames are the most optimal. This is the same finding we have in table \ref{table:context_actionfree_fvd} for both action-free datasets regarding the predicted video dynamics.

\begin{table}[h!] 
\centering
\setlength{\tabcolsep}{3pt}
\vspace{10pt}
\begin{tabular}{l||c||c||ccc}
\Xhline{4\arrayrulewidth}
\Xhline{4\arrayrulewidth}
Dataset & \makecell{Metric} & \makecell{Network} & \makecell{Context = 2} & \makecell{Context = 4} & \makecell{Context = 8} \\
\Xhline{4\arrayrulewidth}
\Xhline{4\arrayrulewidth}
Towel Pick & \makecell{\quad \\ FVD \\ \quad} & \makecell{SVG' \\ LSTM \\ CNN} & \makecell{93.977 \\ 96.138 \\ \textbf{127.281}} & \makecell{71.415 \\ 73.494 \\ 143.394} & \makecell{\textbf{69.038} \\ \textbf{67.015} \\ 131.376} \\
\Xhline{4\arrayrulewidth}
\Xhline{4\arrayrulewidth}
\end{tabular}
\vspace{12pt}
\caption{Fr\'echet Video Distance (FVD) evaluation of the effects of the number of context frames in the action-conditioned dataset (Towel Pick). We compare models with different number of context frames and prediction of 12 frames.}
\vspace{-.1in}
\label{table:context_actionconditioned_fvd}
\end{table}

\newpage
\subsection{All-vs-all Amazon Mechanical Turk comparison} \label{supp:all_compare}
In this section, we compare the largest models we trained for the different inductive bias considered in our study. Similar to the experiments presented in the may text, we use 10 unique workers per video and choose the selection with the most votes as the final answer.
The videos used in the comparison are determined by the highest VGG Cosine Similarity score amongst all samples for the stochastic model, and we use the single trajectory produced by LSTM and CNN.

\begin{table}[h!] 
\centering
\setlength{\tabcolsep}{3pt}
\begin{tabular}{l||cc||ccc}
\Xhline{4\arrayrulewidth}
\Xhline{4\arrayrulewidth}
Dataset & \makecell{Method 1} & \makecell{Method 2} & \makecell{Method 1} & \makecell{Method 2}& \makecell{About the same} \\
\Xhline{4\arrayrulewidth}
\Xhline{4\arrayrulewidth}
Towel Pick & \makecell{SVG' \\ SVG' \\ CNN} & \makecell{LSTM \\ CNN \\ LSTM} & \makecell{43.8\% \\ 38.7\% \\ 32.7\%} & \makecell{53.5\% \\58.2 \% \\ 66.0\%} & \makecell{2.7\% \\ 3.1\% \\ 2.0\%} \\
\Xhline{4\arrayrulewidth}
Human 3.6M & \makecell{SVG' \\ SVG' \\ CNN} & \makecell{LSTM \\ CNN \\ LSTM} & \makecell{34.5\% \\ 96.6\% \\ 2.5\%} & \makecell{63.0\% \\ 2.9\% \\ 97.5\%} & \makecell{2.5\% \\ 0.4\% \\ 0.0\%} \\
\Xhline{4\arrayrulewidth}
KITTI & \makecell{SVG' \\ SVG' \\ CNN} & \makecell{LSTM \\ CNN \\ LSTM} & \makecell{55.4\% \\ 97.3\% \\ 0.7\%} & \makecell{44.6\% \\ 2.7\% \\ 99.3\%} & \makecell{0.0\% \\ 0.0\% \\ 0.0\%} \\
\Xhline{4\arrayrulewidth}
\Xhline{4\arrayrulewidth}
\end{tabular}
\vspace{7pt}
\caption{\textbf{Amazon Mechanical Turk human worker preference}. We compared the biggest and baseline models from LSTM and SVG'. The bigger models are more frequently preferred by humans.}
\vspace{-0.2in}
\label{table:all_vs_all}
\end{table}


\newpage
\subsection{Device and network details}
To scale up the capacity of the model, we use 32 Google TPUv3 Pods \citep{tpu} for each experiment and a batch size of 32. We distribute the training batch such that there is a single batch element in each 16GB TPU. This way we can use each device to the maximum capacity. We first increase $K$ and $M$ together while keeping $K$ to be equals to $M$. By simply doubling the number of neurons in each layer, we see an improvement. We then continue to increase $K$ and $M$ up to three times the number of neurons in each layer. At this, point we are not able to increase $M$ anymore without running out of memory, and so, we only continue increasing $K$.

\subsection{Architecture and hyper-parameters}
For the encoder network we use VGG-net \citep{vggnet} up to layer conv3\_3 after pooling and a single convolutional layer with output of 128 channels. A mirrored architecture of the encoder is used for the decoder network. For the Convolutional LSTMs used throughout we use a single layer network with 512 units for $\text{LSTM}_{\psi}$ and $\text{LSTM}_{\phi}$, and a two layer network with 512 units for $\text{LSTM}_{\theta}$. Other than that, we follow a similar architecture as \cite{emily} including the skip connections from encoder to decoder. We use $\beta=0.0001$ for all of our experiments. The number of hidden units in $z$ are 64 for the robot arm dataset and 128 for all other datasets.

\end{document}